\documentclass[10pt,journal,compsoc]{IEEEtran}
\usepackage{bm}
\usepackage{color}
\usepackage{array}
\usepackage{amssymb}
\usepackage{amsmath}
\usepackage{subfiles}
\usepackage{ragged2e}
\usepackage{multirow}
\usepackage{longtable}
\usepackage{graphicx}
\usepackage{booktabs}
\usepackage{subfigure}
\usepackage{stfloats}
\usepackage[switch]{lineno}
\usepackage[square, comma, sort&compress, numbers]{natbib}
\usepackage[colorlinks, linkcolor=red, anchorcolor=red, citecolor=green,urlcolor=blue]{hyperref}

\definecolor{mark}{rgb}{0,0,0}
\newcommand{\tabincell}[2]{\begin{tabular}{@{}#1@{}}#2\end{tabular}}
\begin{document}

\title{Self-supervised Learning on Graphs: Contrastive, Generative,or Predictive}

\author{Lirong~Wu,~Haitao~Lin,~
        Cheng~Tan,\\~Zhangyang~Gao,
        and~Stan.Z.Li,~\IEEEmembership{Fellow,~IEEE}}
        


\IEEEtitleabstractindextext{
\justifying
\begin{abstract}
Deep learning on graphs has recently achieved remarkable success on a variety of tasks, while such success relies heavily on the massive and carefully labeled data. However, precise annotations are generally very expensive and time-consuming. To address this problem, self-supervised learning (SSL) is emerging as a new paradigm for extracting informative knowledge through well-designed pretext tasks without relying on manual labels. In this survey, we extend the concept of SSL, which first emerged in the fields of computer vision and natural language processing, to present a timely and comprehensive review of existing SSL techniques for graph data. Specifically, we divide existing graph SSL methods into three categories: contrastive, generative, and predictive. More importantly, unlike other surveys that only provide a high-level description of published research, we present an additional mathematical summary of existing works in a unified framework. Furthermore, to facilitate methodological development and empirical comparisons, we also summarize the commonly used datasets, evaluation metrics, downstream tasks, open-source implementations, and experimental study of various algorithms. Finally, we discuss the technical challenges and potential future directions for improving graph self-supervised learning. Latest advances in graph SSL are summarized in a GitHub repository \url{https://github.com/LirongWu/
awesome-graph-self-supervised-learning}.
\end{abstract}

}

\maketitle

\section{Introduction}
\IEEEPARstart{I}{n} recent years, deep learning on graphs has emerged as a popular research topic, due to its ability to capture both graph structures and node/edge features. However, most of the works have focused on supervised or semi-supervised learning settings, where the model is trained by specific downstream tasks with abundant labeled data, which are often limited, expensive, and inaccessible. Due to the heavy reliance on the number and quality of labels, these methods are hardly applicable to real-world scenarios, especially those requiring expert knowledge for annotation, such as medicine, meteorology, transportation, etc. More importantly, these methods are prone to suffer from problems of over-fitting, poor generalization, and weak robustness \cite{liu2020self}.

Recent advances in SSL \cite{he2020momentum,chen2020simple,grill2020bootstrap,devlin2018bert,radford2019language,lan2019albert} have provided novel insights into reducing the dependency on annotated labels and enable the training on massive unlabeled data. The primary goal of SSL is to learn transferable knowledge from abundant unlabeled data with well-designed pretext tasks and then generalize the learned knowledge to downstream tasks with specific supervision signals. Recently, SSL has shown its promising capability in the field of computer vision (CV) \cite{he2020momentum,chen2020simple,grill2020bootstrap} and natural language processing (NLP) \cite{devlin2018bert,radford2019language,lan2019albert}. For example, Moco \cite{he2020momentum} and SimCLR \cite{chen2020simple} augment image data by various means and then train Convolutional Neural Networks (CNNs) to capture dependencies between different augmentations. Besides, BERT \cite{devlin2018bert} pre-trains the model with Masked LM and Next Sentence Prediction as pretext tasks, achieving state-of-the-art performance on up to 11 tasks compared to those conventional methods. Inspired by the success of SSL in CV and NLP domains, it is a promising direction to apply SSL to the graph domain to fully exploit graph structure information and rich unlabeled data.

Compared with image and language sequence data, applying SSL to graph data is very important and has promising research prospects. Firstly, Firstly, along with node features, graph data also contains structure, where extensive pretext tasks can be designed to capture the intrinsic relations of nodes. Secondly, real-world graphs are usually formed by specific rules, e.g., links between atoms in molecular graphs are bounded by valence bond theory. Thus, extensive expert knowledge can be incorporated as a priori into the design of pretext tasks. Finally, graph data generally supports transductive learning \cite{hamilton2017inductive}, such as node classification, where the features of Train/Val/Test samples are all available during the training process, which makes it possible to design more feature-related pretext tasks.

\begin{figure}[ht]
	\begin{center}
		\subfigure[Contrastive Method]{\includegraphics[width=1\linewidth]{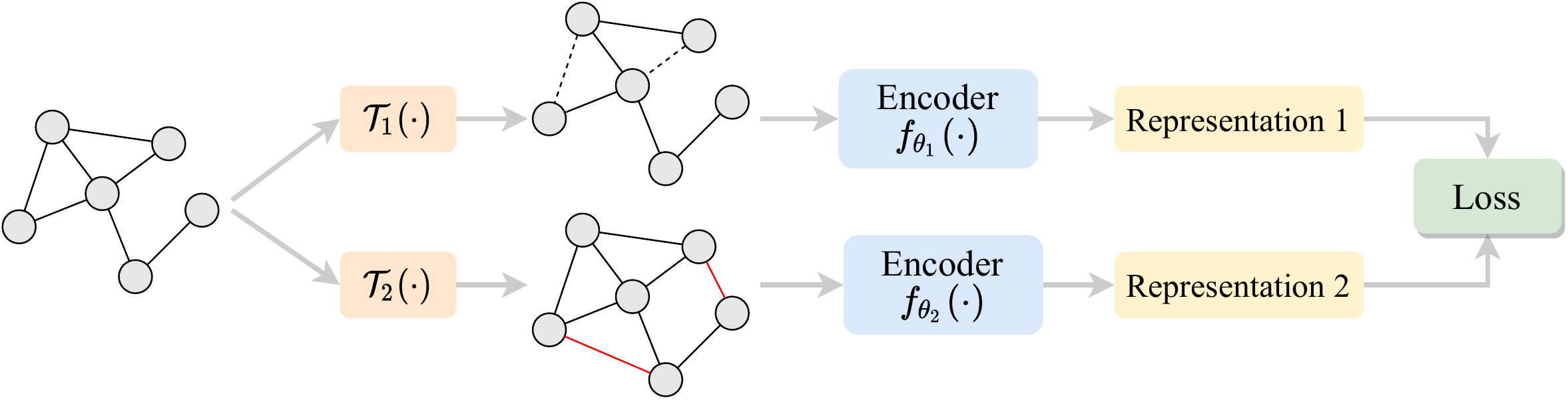}}
		\subfigure[Generative Method]{\includegraphics[width=1\linewidth]{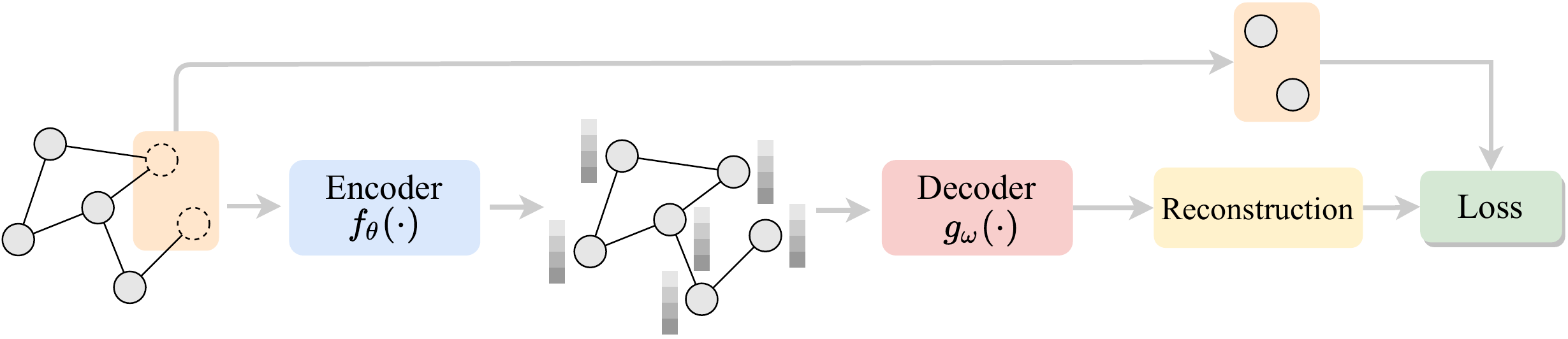}}
		\subfigure[Predictive Method]{\includegraphics[width=1\linewidth]{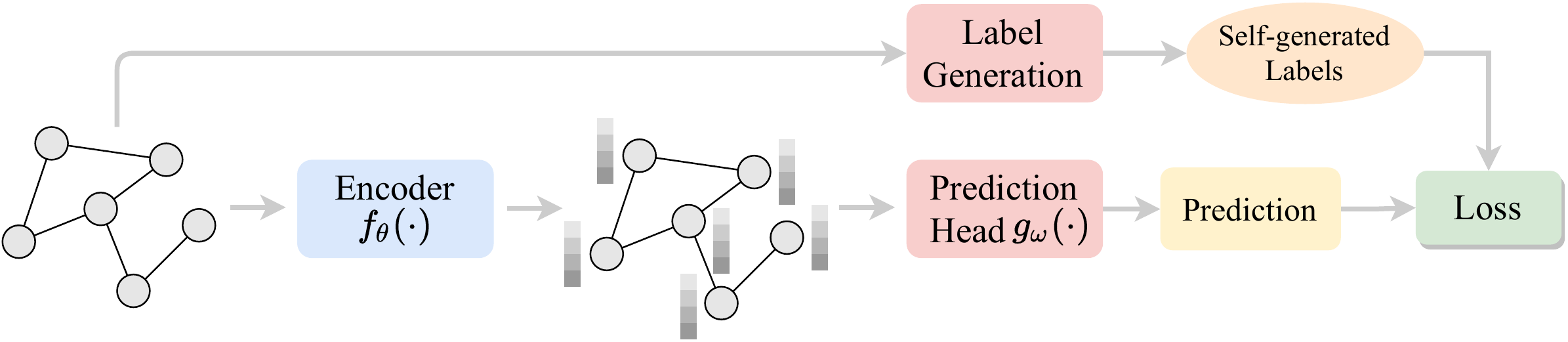}}
	\end{center}
	\vspace{-1em}
	\caption{A comparison among the contrastive, generative, and predictive method. \textbf{(a)}: the contrastive method contrasts the views generated from different augmentation $\mathcal{T}_1(\cdot)$ and $\mathcal{T}_2(\cdot)$. The information about the differences and sameness between (inter-data) data-data pairs are used as self-supervision signals. \textbf{(b)}: the generative method focuses on the (intra-data) information embedded in the graph, generally based on pretext tasks such as reconstruction, which exploit the attributes and structures of graph data as self-supervision signals. \textbf{(c)}: the predictive method generally self-generates labels by some simple statistical analysis or expert knowledge, and designs prediction-based pretext tasks based on self-generated labels to handle the data-label relationship.}
	\vspace{-1em}
	\label{fig:1}
\end{figure}

Though some works have been proposed recently to apply SSL to graph data and have achieved remarkable success \cite{velickovic2019deep,jin2020self,hu2019strategies,you2020does,manessi2020graph,zhu2020self,zhang2020graph,peng2020graph,hu2019pre,hassani2020contrastive,jiao2020sub}, it is still very challenging to deal with the inherent differences between grid-like and structured-like data. Firstly, the topology of the image is a fixed grid, and the text is a simple sequence, while graphs are not restricted to these rigid structures. Secondly, unlike the assumption of independent and identical sample distribution for image and text data, nodes in the graph are correlated with each other rather than completely independent. This requires us to design pretext tasks by considering both node attributes and graph structures. Finally, there exists a gap between self-supervised pretext tasks and downstream tasks due to the divergence of their optimization objectives. Inevitably, such divergence will significantly hurt the generalization of learned models. Therefore, it is crucial to reconsider the objectives of pretext tasks to better match that of downstream tasks and make them consistent with each other.

In this survey, we extend the concept of SSL, which first emerged in the fields of computer vision and natural language processing, to present a timely and comprehensive review of the existing SSL techniques for graph data. Specifically, we divide existing graph SSL methods into three categories: contrastive, generative, and predictive, as shown in Fig.~\ref{fig:1}. The core contributions of this survey are as follow:

\begin{itemize}
\item Present comprehensive and up-to-date reviews on existing graph SSL methods and divide them into three categories: contrastive, generative, and predictive, to improve their clarity and accessibility.
\item Summarize the core mathematical ideas of recent research in graph SSL within a unified framework. 
\item Summarize the commonly used datasets, evaluation metrics, downstream tasks, open-source codes, and experimental study of surveyed methods, setting the stage for developments of future works.
\item Point out the technical limitations of current research and discuss promising directions on graph SSL.
\end{itemize}

Compared to the existing surveys on SSL \cite{liu2020self}, we purely focus on SSL for graph data and present a more mathematical review on the recent progress from the year 2019 to 2021. Though there have been two surveys on graph SSL, we argue that they are immature work with various flaws and shortcomings. For example, \cite{xie2021self} clumsily \emph{describes each method in 1-2 sentences}, lacking deep insight into the mathematical ideas and implementation details behind. Moreover, the number of reviewed methods in \cite{liu2021graph} are even fewer than half of ours, as it spends too much description on those \emph{less important} contents, but ignores the core of graph SSL, i.e., the design of the pretext task. Compared with \cite{xie2021self,liu2021graph}, our advantages are as follow: (1) more mathematical details, striving to summarize each method with one equation; (2) more implementation details, including 41 datasets statistics (\textit{vs} 20 datasets in \cite{xie2021self}), evaluation metrics, and open-source code; (3) more thorough experimental study for fair comparison; (4) more fine-grained, clarified and rational taxonomy; (5) more surveyed works, 71 methods (\textit{vs} 47 methods in \cite{liu2021graph} \textit{vs} 18 methods in \cite{xie2021self}); (5) more up-to-data review, with almost all surveyed works published after 2019.
\section{Problem Statement}
\subsection{Notions and Definitions}

Unless particularly specified, the notations used in this survey are illustrated in Table.~\ref{tab1:notations}.

\textit{Definition 1 (Graph):}
We use $g=(\mathcal{V},\mathcal{E})$ to denote a graph where $\mathcal{V}$ is the set of $N$ nodes and $\mathcal{E}$ is the set of $M$ edges. Let $v_i\in\mathcal{V}$ denote a node and $e_{i,j}$ denote an edge between node $v_i$ and $v_j$. The $l$-hop neighborhood of a node $v_i$ is denoted as $\mathcal{N}_i^{(l)}=\{v_j\in \mathcal{V} | d(v_i,v_j) \leq l\}$ where $d(v_i,v_j)$ is the shortest path length between node $v_i$ and $v_j$. In particular, the $1$-hop neighborhood of a node $v_i$ is denoted as $\mathcal{N}_i=\mathcal{N}_i^{(1)}=\{v_j\in \mathcal{V} | e_{i,j} \in \mathcal{E}\}$. The graph structure can also be represented by an adjacency matrix $\mathbf{A} \in[0,1]^{N \times N}$ with $\mathbf{A}_{i,j}=1$ if $e_{i,j}\in\mathcal{E}$ and $\mathbf{A}_{i,j}=0$ if $e_{i,j} \notin \mathcal{E}$. 

\textit{Definition 2 (Attribute Graph):}
Attributed graph, an opposite concept to the unattributed one, refers to a graph where nodes or edges are associated with their own features (a.k.a, attributes). For example, each node $v_i$ in graph $g$ may be associated with a feature vector $\mathbf{x}_{i}\in\mathbb{R}^{d_0}$, such a graph is referred to an attributed graph $g=(\mathcal{V},\mathcal{E},\mathbf{X})$ or $g=(\mathbf{A},\mathbf{X})$, where $\mathbf{X}=\left[\mathbf{x}_{1}, \mathbf{x}_{2}, \ldots, \mathbf{x}_{N}\right]$ is the node feature matrix. Meanwhile, an attributed graph $g=(\mathcal{V},\mathcal{E},\mathbf{X}^e)$ may have edge attributes $\mathbf{X}^e$, where $\mathbf{X}^e \in \mathbb{R}^{M \times b_0}$ is an edge feature matrix with $\mathbf{x}^e_{i,j}\in\mathbb{R}^{b_0}$ being the feature vector of edge $e_{i,j}$.

\textit{Definition 3 (Dynamic Graph):}
A dynamic graph is a special attributed graph where the node set, graph structure and node attributes may change dynamically over time. The dynamic graph can be formalized as $g=(\mathcal{V}^{(t)},\mathcal{E}^{(t)},\mathbf{X}^{(t)})$ or $g=(\mathbf{A}^{(t)},\mathbf{X}^{(t)})$, where
$\mathcal{E}^{(t)}$ represents the edge set at the time step $t$ and  $\mathbf{A}^{(t)}_{i,j}=1$ denotes an interaction between node $v_i$ and $v_j$ at the time step $t$ ($1\leq t \leq T$).

\textit{Definition 4 (Heterogeneous Graph):}
Consider a graph $g=(\mathcal{V},\mathcal{E})$ with a node type mapping function $f_{v}: \mathcal{V} \rightarrow \mathcal{Y}^{v}$ and an edge type mapping function $f_{e}: \mathcal{E} \rightarrow \mathcal{Y}^{e}$, where $\mathcal{Y}^{v}$ is the set of node types and $\mathcal{Y}^{e}$ is the set of edge types. For a graph with more than one type of node or edge, e.g., $\left|\mathcal{Y}^{v}\right|>1$ or $\left|\mathcal{Y}^{e}\right|>1$, we define it as a heterogeneous graph, otherwise, it is a homogeneous graph. There are some special types of heterogeneous graphs: a bipartite graph with $\left|\mathcal{Y}^{v}\right|=2$ and $\left|\mathcal{Y}^{e}\right|=1$, and a multiplex graph with $\left|\mathcal{Y}^{v}\right|=1$ and $\left|\mathcal{Y}^{e}\right|>1$.

\textit{Definition 5 (Spatial-Temporal Graph):}
A spatial-temporal graph is a special dynamic graph, but noly the node attributes change over time with the node set and graph structure unchanged. The spatial-temporal graph is defined as $g=(\mathcal{V},\mathcal{E},\mathbf{X}^{(t)})$ or $g=(\mathbf{A},\mathbf{X}^{(t)})$, where $\mathbf{X}^{(t)}\in\mathbb{R}^{N \times d_0}$ is the node feature matrix at the time step $t$ ($1\leq t \leq T$).

\subsection{Downstream Tasks}
The downstream tasks for graph data are divided into three categories: node-level, link-level, and graph-level tasks. A node-level graph encoder $f_\theta(\cdot)$ is often used to generate node embeddings for each node, and a graph-level graph encoder $f_\gamma(\cdot)$ is used to generate graph-level embeddings. Finally, the learned embeddings are fed into an optional prediction head $g_\omega(\cdot)$ to perform specific downstream tasks.

\textbf{Node-level tasks.}
Node-level tasks focus on the properties of nodes, and  node classification is a typical node-level task where only a subset of node $\mathcal{V}_L$ with corresponding labels $\mathcal{Y}_L$ are known, and we denote the labeled data as $\mathcal{D}_L=(\mathcal{V}_L,\mathcal{Y}_L)$ and unlabeled data as $\mathcal{D}_U=(\mathcal{V}_U,\mathcal{Y}_U)$. Let $f_{\theta}: \mathcal{V} \rightarrow \mathcal{Y}$ be a graph encoder trained on labeled data $\mathcal{D}_L$ so that it can be used to infer the labels $\mathcal{Y}_U$ of unlabeled data. Thus, the objective function for node classification can be defined as minimizing loss $\mathcal{L}_{node}$, as follows
\begin{equation}
\begin{aligned}
\min _{\theta,\omega} \mathcal{L}_{node}&\left(\mathbf{A},\mathbf{X}, \theta, \omega\right)= \sum_{(v_{i}, y_{i}) \in \mathcal{D}_{L}} \ell\big(g_\omega(\mathbf{h}_i), y_{i}\big)
\end{aligned}
\end{equation}

\noindent where $\mathbf{H}=f_{\theta}(\mathbf{A},\mathbf{X})$ and $\mathbf{h}_i$ is the embedding of node $v_i$ in the embedding matrix $\mathbf{H}$. $\ell(\cdot, \cdot)$ denotes the cross entropy.

\textbf{Link-level tasks.}
Link-level tasks focus on the representation of node paris or properties of edges. Taking link prediction as an example, given two nodes, the goal is to discriminate if there is an edge between them. Thus, the objective function for link prediction can be defined as,
\begin{equation}
\begin{split}
\min _{\theta,\omega} \mathcal{L}_{link}\left(\mathbf{A},\mathbf{X}, \theta, \omega\right)=\sum_{v_{i}, v_{j} \in \mathcal{V}, i \neq j} \ell\big(g_\omega(\mathbf{h}_i,\mathbf{h}_j), \mathbf{A}_{i,j}\big) 
\end{split}
\end{equation}

\noindent where $\mathbf{H}=f_{\theta}(\mathbf{A},\mathbf{X})$ and $\mathbf{h}_i$ is the embedding of node $v_i$ in the embedding matrix $\mathbf{H}$. $g_\omega(\cdot)$ linearly maps the input to a 1-dimension value, and $\ell(\cdot, \cdot)$ is the cross entropy.

\begin{table}
\caption{Notations used in this paper.}
\label{tab1:notations}
\centering
\vspace{-1.5em}
\begin{tabular}{l|l}

\hline
\textbf{Notations} & \textbf{Descriptions}\\
\hline

$\mathbb{R}^m$ & $m$-dimensional Euclidean space\\ \hline
$a, \mathbf{a}, \mathbf{A}$ & Scalar, vector, matrix\\ \hline
$\mathcal{G}$ & The set of graphs, $\mathcal{G}=\{g_1,g_2,\cdots,g_{|\mathcal{G}|}\}$ \\ \hline
$g$ & A graph $g=(\mathcal{V}, \mathcal{E})$\\ \hline
$\mathcal{V}$ & The set of nodes in graph $g$\\ \hline
$v_i$ & A node $v_i \in \mathcal{V}$\\ \hline
$\mathcal{E}$ & The set of edges in graph $g$\\ \hline
$e_{i,j}$ & An edge $e_{i,j} \in \mathcal{E}$ between node $v_i$ and node $v_j$\\ \hline
$N$ & Number of nodes, $N=|\mathcal{V}|$\\ \hline
$M$ & Number of edges, $M=|\mathcal{E}|$\\ \hline

$\mathbf{A}\in\mathbb{R}^{N \times N}$ & A graph adjacency matrix\\ \hline
$\mathbf{D}$ & The degree matrix of $\mathbf{A}$, $\mathbf{D}_{i i}=\sum_{j=1}^{n} \mathbf{A}_{i j}$\\ \hline
$\mathbf{I}_N$ & Identity matrix of dimension $N$ \\ \hline
$\mathcal{N}_i^{(l)}$ & $l$-hop Neighborhood set of node $v_i$\\ \hline
$\mathcal{N}_i$ & $1$-hop Neighborhood set of node $v_i$\\ \hline

$L$ & The layer number\\ \hline
$l$ & The layer index, $1\leq l \leq L$\\ \hline
$T$ & The time step/iteration number\\ \hline
$t$ & The time step/iteration index, $1\leq t \leq T$\\ \hline
$d_0$ & Dimension of node feature vectors\\ \hline
$d_l$ & Dimension of node embeddings in the $l$-th layer\\ \hline
$b_0$ & Dimension of edge feature vectors\\ \hline
$\mathbf{x}_i\in\mathbb{R}^{d_0}$ & Feature vector of node $v_i$\\ \hline
$\mathbf{X}\in\mathbb{R}^{N \times d_0}$ & Node feature matrix, $\mathbf{X}=\left[\mathbf{x}_{1}, \mathbf{x}_{2}, \ldots, \mathbf{x}_{N}\right]$\\ \hline
$\mathbf{X}^{(t)}\in\mathbb{R}^{N \times d_0}$ & Node feature matrix at the time step $t$\\ \hline
$\mathbf{x}^e_{i,j}\in\mathbb{R}^{b_0}$ & Feature vector of edge $e_{i,j}$\\ \hline
$\mathbf{X}^e\in\mathbb{R}^{M \times b_0}$ & Edge feature matrix\\ \hline
$\mathbf{h}_i^{(l)}\in\mathbb{R}^{d_l}$ & Node embedding of node $v_i$ in the $l$-th layer\\ \hline
$\mathbf{H}^{(l)}\in\mathbb{R}^{N \times d_l}$ & Embedding matrix in the $l$-th layer\\ \hline
$\mathbf{h}_i\in\mathbb{R}^{d_L}$ & Node embedding in the $L$-th layer, $\mathbf{h}_i=\mathbf{h}_i^{(L)}$\\ \hline
$\mathbf{H}\in\mathbb{R}^{N \times d_L}$ & Embedding matrix in the $L$-th layer, $\mathbf{H}=\mathbf{H}^{(L)}$\\ \hline
$\mathbf{h}_g\in\mathbb{R}^{d_L}$ & Graph-level representation of graph $g$\\ \hline

$|\cdot|$ & The length of a set\\ \hline
$\odot$ & Element-wise multiplication operation\\ \hline
$\|$ & Vector concatenation\\ \hline

$\sigma(\cdot)$ & The logistic sigmoid activation function\\ \hline
$\tanh(\cdot)$ & The hyperbolic tangent activation function\\ \hline
LeakyReLU$(\cdot)$ & The LeakyReLU activation function\\ \hline

$\operatorname{READOUT}(\cdot)$ & The readout function\\ \hline
$f_\theta,f_{\theta_1},f_{\theta_2},\cdots$ & Node-level encoder to output $\mathbf{H}=f_\theta(\mathbf{A},\mathbf{X})$\\ \hline
$f_\gamma,f_{\gamma_1},f_{\gamma_2},\cdots$ & Graph-level encoder to output $\mathbf{h}_g=f_\gamma(\mathbf{A},\mathbf{X})$\\ \hline
$g_\omega,g_{\omega_1},g_{\omega_2},\cdots$ & The prediction head\\ \hline
$\mathcal{T},\mathcal{T}_1,\mathcal{T}_2,\cdots$ & The data augmentation transformation \\ \hline
$\mathbf{W}, \Theta, \theta, \gamma, \omega$ & Learnable model parameters\\ \hline

\end{tabular}
\vspace{-1em}
\end{table}

\textbf{Graph-level tasks.}
Graph-level tasks learn from multiple graphs in a dataset
and predict the property of a single graph. For example, graph regression is a typical graph-level task where only a subset of graphs $\mathcal{G}_L$ with corresponding properties $\mathcal{P}_L$ are known, and we denote it as $\mathcal{D}_L=(\mathcal{G}_L,\mathcal{P}_L)$. Let $f_{\gamma}: \mathcal{G} \rightarrow \mathcal{P}$ be a graph encoder trained on labeled data $\mathcal{D}_L$ and then used to infer the properties $\mathcal{P}_U$ of unlabeled graphs $\mathcal{G}_U$. Thus, the objective function for graph regression can be defined as minimizing loss $\mathcal{L}_{graph}$,
\begin{equation}
\min _{\gamma,\omega} \mathcal{L}_{graph}\left(\mathbf{A}_{i},\mathbf{X}_{i}, \gamma, \omega\right)=\sum_{\left(g_{i}, p_{i}\right) \in \mathcal{D}_{L}} \ell\big(g_\omega(\mathbf{h}_{g_i}), p_{i}\big)
\end{equation}

\noindent where $\mathbf{h}_{g_i}=f_{\gamma}(\mathbf{A},\mathbf{X})$ is the graph-level representation of graph $g_i$. $g_\omega(\cdot)$ linearly maps the input to a 1-dimension value, and $\ell(\cdot, \cdot)$ is the mean absolute error.

\subsection{Graph Neural Networks}
Graph neural networks (GNN) \cite{kipf2016semi,velivckovic2017graph,hamilton2017inductive} are a family of neural networks that have been widely used as the backbone encoder in most of the reviewed works. A general GNN framework involves two key computations for each node $v_i$ at every layer: (1) $\operatorname{AGGREGATE}$ operation: aggregating messages from neighborhood $\mathcal{N}_i$; (2) $\operatorname{UPDATE}$ operation: updating node representation from its representation in the previous layer and the aggregated messages. Considering a $L$-layer GNN, the formulation of the $l$-th layer is as follows
\begin{equation}
\begin{small}
\begin{aligned}
\mathbf{a}_{i}^{(l)} & = \operatorname{AGGREGATE}^{(l)}\left(\left\{\mathbf{h}_{j}^{(l-1)}: v_{j} \in \mathcal{N}_i\right\}\right) \\
\mathbf{h}_{i}^{(l)} & = \operatorname{UPDATE}^{(l)}\left(\mathbf{h}_{i}^{(l-1)}, \mathbf{a}_{i}^{(l)}\right)
\end{aligned}
\end{small}
\end{equation}

\noindent where $1\leq l \leq L$ and $\mathbf{h}_{i}^{(l)}$ is the embedding of node $v_i$ in the $l$-th layer with $\mathbf{h}_{i}^{(0)}=\mathbf{x}_{i}$. For node-level or edge-level tasks, the node representation $\mathbf{h}_{i}^{(L)}$ can sometimes be used for downstream tasks directly. However, for graph-level tasks, an extra $\operatorname{READOUT}$ function is required to aggregate node features to obtain a graph-level representation $\mathbf{h}_{g}$, as follow
\begin{equation}
\begin{small}
\begin{aligned}
\mathbf{h}_{g}=\operatorname{READOUT}\left(\left\{\mathbf{h}_{i}^{(L)} \mid v_{i} \in \mathcal{V}\right\}\right)
\end{aligned}
\end{small}
\end{equation}

The design of these component functions is crucial, but it is beyond the scope of this paper. For a thorough review, we refer readers to the recent survey \cite{wu2020comprehensive}.

\begin{figure}[ht]
	\begin{center}
		\subfigure[Pre-train\&Fine-tune (P\&F)]{\includegraphics[width=1\linewidth]{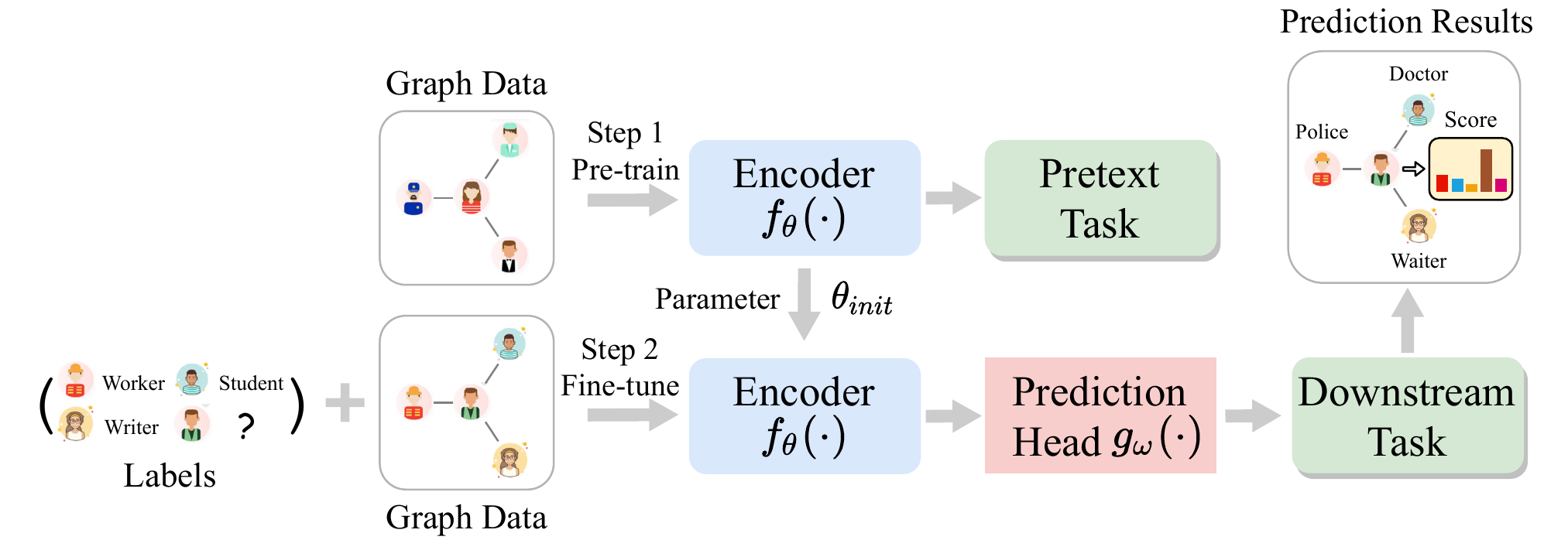}}
		\vspace{-1em}
		\subfigure[Joint Learning (JL)]{\includegraphics[width=1\linewidth]{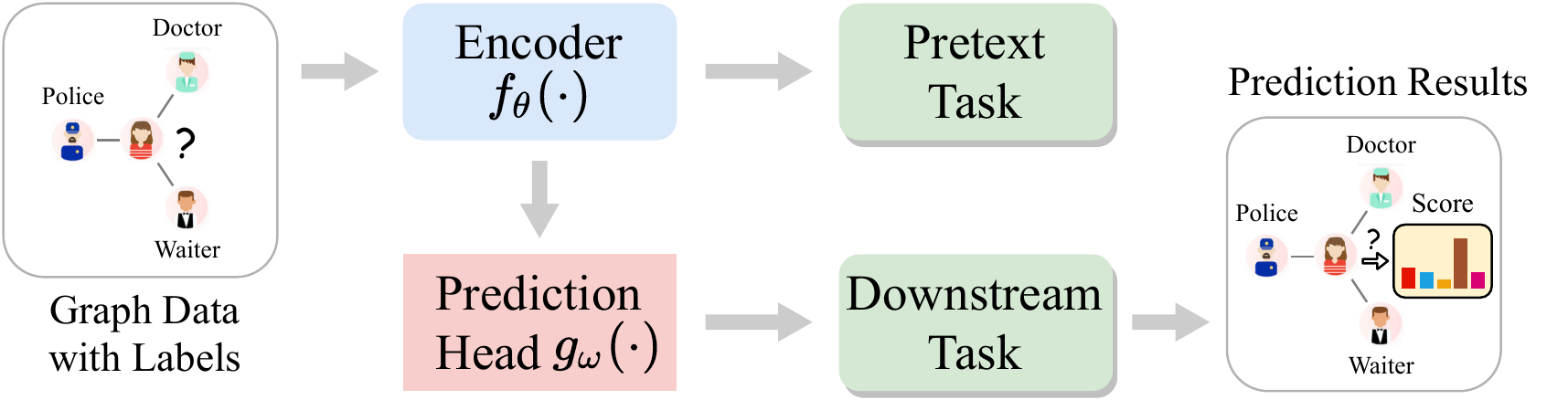}}
		\subfigure[Unsupervised Representation Learning (URL)]{\includegraphics[width=1\linewidth]{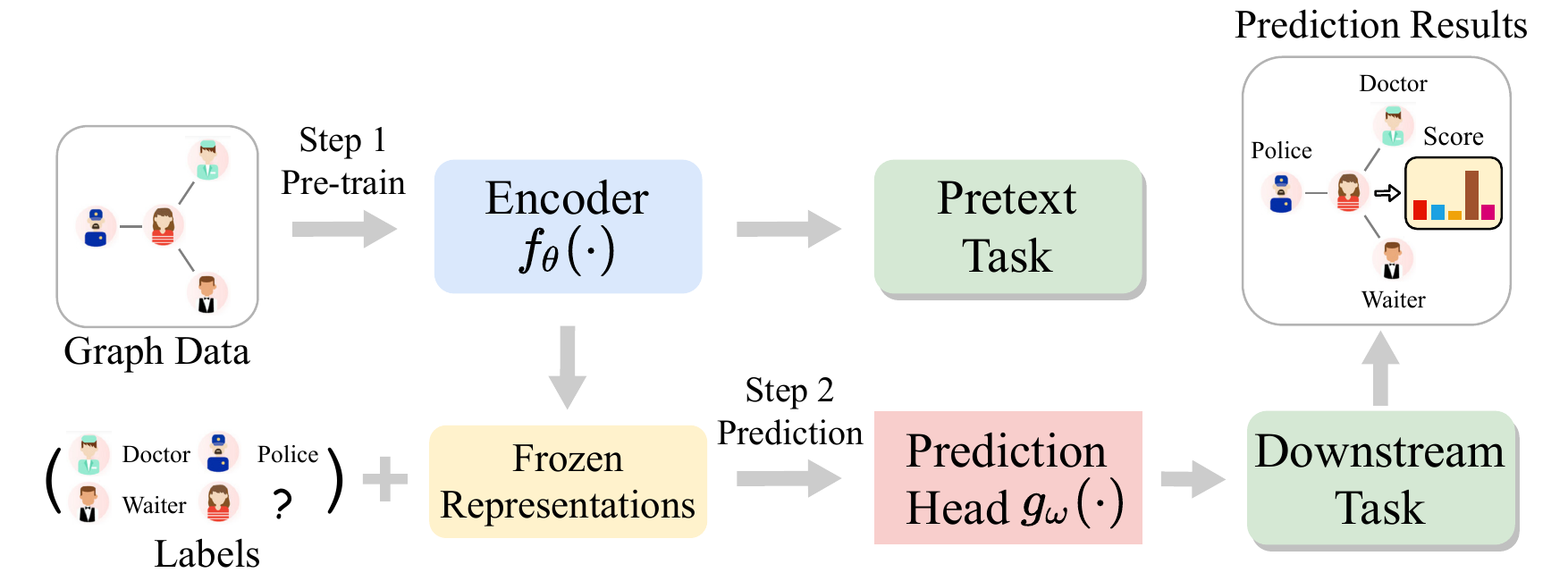}}
	\end{center}
	\vspace{-1em}
	\caption{An overview of training strategies for graph SSL. The training strategies can be divided into three categories. \textbf{(a)}: for the Pre-train\&Fine-tune strategy, it first pre-trains the encoder $f_\theta(\cdot)$ with unlabeled nodes by the self-supervised pretext tasks. The pre-trained encoder’s parameters $\theta_{init}$ are then used as the initialization of the encoder for supervised fine-tuning on downstream tasks. \textbf{(b)}: for the Joint Learning strategy, an auxiliary pretext task is included to help learn the supervised downstream task. The encoder is trained through both the pretext task and the downstream task simultaneously. \textbf{(c)}: for the Unsupervised Representation Learning strategy, it first pre-trains the encoder $f_\theta(\cdot)$ with unlabeled nodes by the self-supervised pretext tasks. The pre-trained encoder’s parameters $\theta_{init}$ are then frozen and used in the supervised downstream task with additional labels.}
	\vspace{-1em}
	\label{fig:2}
\end{figure}

\subsection{Training Strategy}
The training strategies can be divided into three categories: Pre-training and Fine-tuning (P\&F), Joint Learning (JL), and Unsupervised Representation Learning (URL), with their detailed workflow shown in Fig.~\ref{fig:2}.

\textbf{Pre-training and Fine-tuning (P\&F).}
In this strategy, the model is trained in a two-stage paradigm \cite{jin2020self}. The encoder $f_\theta(\cdot)$ is pre-trained with the pretext tasks, then the pre-trained parameters $\theta_{init}$ are used as the initialization of the encoder $f_{\theta_{init}}(\cdot)$. At the fine-tuning stage, the pre-trained encoder $f_{\theta_{init}}(\cdot)$ is fine-tuned with a prediction head $g_\omega(\cdot)$ under the supervision of specific downstream tasks. The learning objective is formulated as
\begin{equation}
\theta^{*}, \omega^{*}=\arg \min _{(\theta, \omega)} \mathcal{L}_{task}(f_{\theta_{init}}, g_\omega)
\end{equation}

\noindent with initialization $\theta_{init}=\arg \min _{\theta} \mathcal{L}_{ssl}(f_\theta)$, where $\mathcal{L}_{task}$ and $\mathcal{L}_{ssl}$ is the loss function of downstream tasks and self-supervised pretext tasks, respectively.

\textbf{Joint Learning.}
In this scheme, the encoder $f_\theta(\cdot)$ is jointly trained with a prediction head $g_\omega(\cdot)$ under the supervision of pretext tasks and downstream tasks. The joint learning strategy can also be considered as a kind of multi-task learning or the pretext tasks are served as a regularization. The learning objective is formulated as
\begin{equation}
\theta^{*}, \omega^{*}=\arg \min _{(\theta, \omega)} \mathcal{L}_{task}(f_\theta, g_\omega)+\alpha \arg \min _{\theta} \mathcal{L}_{ssl}(f_\theta)
\end{equation}

\noindent where $\alpha$ is a trade-off hyperparameter.

\textbf{Unsupervised Representation Learning.}
This strategy can also be considered as a two-stage paradigm, with the first stage similar to Pre-training. However, at the second stage, the pre-trained parameters $\theta_{init}$ are \emph{frozen} and the model is trained on the \emph{frozen representations} with downstream tasks only. The learning objective is formulated as
\begin{equation}
\omega^{*}=\arg \min _{\omega} \mathcal{L}_{task}(f_{\theta_{init}}, g_\omega)
\end{equation}

\noindent with initialization $\theta_{init}=\arg \min _{\theta}\mathcal{L}_{ssl}(f_\theta)$. \textcolor{mark}{Compared to other schemes, unsupervised representation learning is more challenging since the learning of the encoder $f_\theta(\cdot)$ depends only on the pretext task and is frozen in the second stage. In contrast, in the P\&F strategy, the encoder $f_\theta(\cdot)$ can be further optimized under the supervision of the downstream task during the fine-tuning stage.}
\section{Contrastive Learning}
\subsection{A Unified Perspective}
Inspired by the recent advances of contrastive learning in CV and NLP domains, some works have been proposed to apply contrastive learning for graph data. However, most works simply present motivations or implementations from different perspectives, but \emph{adopt very similar (or even the same) architectures and designs in practice}, which leads to the emergence of \emph{duplicative efforts} and hinders the healthy development of the community. In this survey, we therefore review existing work from a unified perspective and unify them into a general framework, and present various designs for the three main modules for contrastive learning, e.g., data augmentation, pretext tasks, and contrastive objectives. In turn, the contributions of existing work can be essentially summarized as innovations in these three modules.

In practice, we usually generate multiple views for each instance through various data augmentations. Two views generated from the same instance are usually considered as a positive pair, while two views generated from different instances are considered as a negative pair. The primary goal of contrastive learning is to maximize the agreement of two jointly sampled positive pairs against the agreement of two independently sampled negative pairs. The agreement between views is usually measured through Mutual Information (MI) estimation. Given a graph $
g=(\mathbf{A},\mathbf{X})$, $K$ different transformations $\mathcal{T}_1,\mathcal{T}_1,\cdots,\mathcal{T}_K$ can be applied to obtain multiple views $\{(\mathbf{A}_k,\mathbf{X}_k)\}_{k=1}^K$, defined as 
\begin{equation}
\mathbf{A}_k,\mathbf{X}_k = \mathcal{T}_k(\mathbf{A},\mathbf{X});k=1,2,\cdots,K
\end{equation}

\noindent Secondly, a set of graph encoders $\{f_{\theta_k}\}_{k=1}^K$ (may be \emph{identical} or \emph{share weights}) can be used to generate different representations $\mathbf{h}_1,\mathbf{h}_2,\cdots,\mathbf{h}_K$ for each view, given by
\begin{equation}
\mathbf{h}_k = f_{\theta_k}(\mathbf{A}_i,\mathbf{X}_i);k=1,2,\cdots,K
\end{equation}

The contrastive learning aims to maximize the mutual information of two views from the same instance as 
\begin{equation}
\max_{\theta_1,\theta_2,\cdots,\theta_K} \sum_{i}\sum_{j \neq i} \alpha_{i,j}\mathcal{MI}(\mathbf{h}_i,\mathbf{h}_j)
\end{equation}

\noindent where $i,j\in\{1,2,\cdots,K\}$, $\{\mathbf{h}_i\}_{i=1}^K$ are representations generated from $g=(\mathbf{A},\mathbf{X})$, which are taken as positive samples. $\mathcal{MI}(\mathbf{h}_i,\mathbf{h}_j)$ are the mutual information between two representations $\mathbf{h}_i$ and $\mathbf{h}_j$. Note that \emph{depending on different pretext tasks}, $\{\mathbf{h}_k\}_{k=1}^K$ may not be at the same scale, either being a node-level, subgraph-level, or graph-level representation. The negative samples to contrast with $\{\mathbf{h}_i\}_{i=1}^K$ can be taken as representations $\{\widetilde{\mathbf{h}}_i\}_{i=1}^K$ that are generated from another graph $\widetilde{g}=(\widetilde{\mathbf{A}},\widetilde{\mathbf{X}})$. Besides, we have $\alpha_{i,j} \in \{0,1\}$, and their concrete values vary in different 
schemes. 

The design of the contrastive learning for graph data can be summarized as three main modules: (1) data augmentation strategy, (2) pretext task, and (3) contrastive objective. The design of graph encoder is not the focus of graph self-supervised learning and beyond the scope of this survey; for more details, please refer to the related survey \cite{wu2020comprehensive}.

\subsection{Data Augmentation}
\begin{figure*}[ht]
	\begin{center}
		\subfigure[Feature-based]{\includegraphics[width=0.35\linewidth]{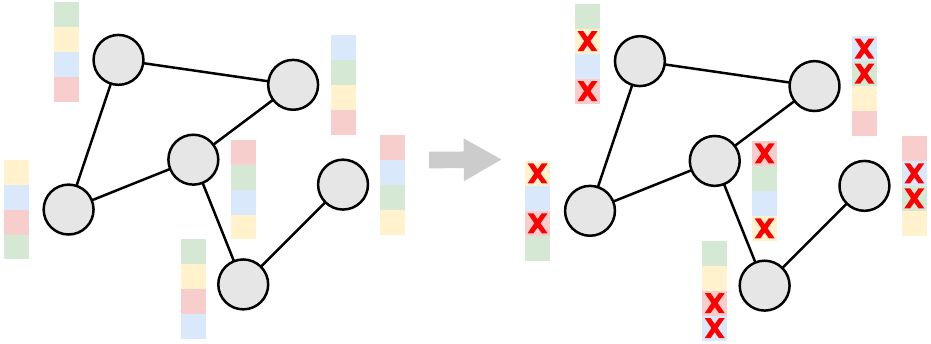}}
		\subfigure[Structue-based]{\includegraphics[width=0.35\linewidth]{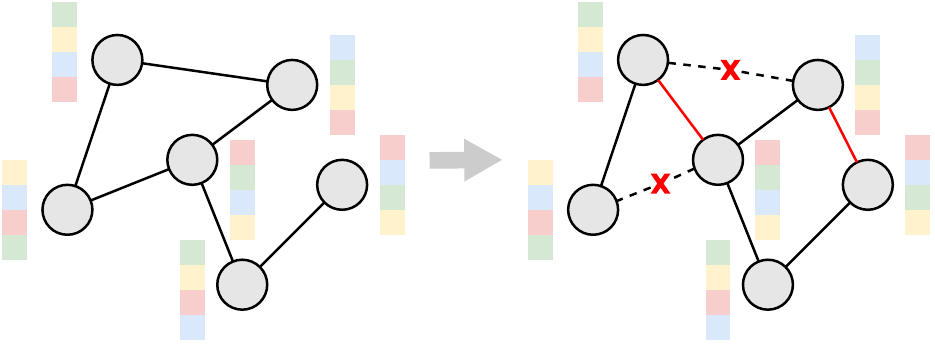}}
		\subfigure[Sampling-based]{\includegraphics[width=0.32\linewidth]{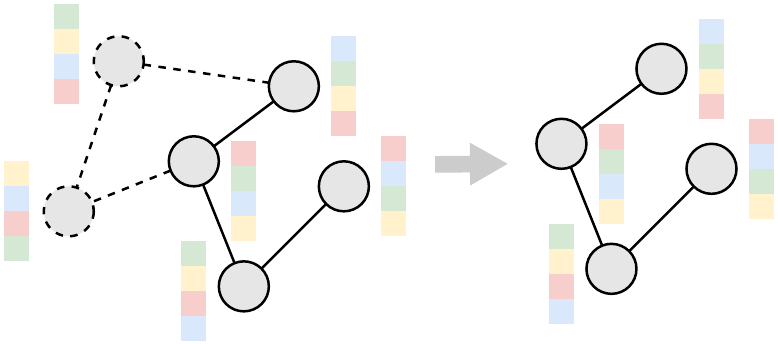}}
		\subfigure[Adaptive]{\includegraphics[width=0.36\linewidth]{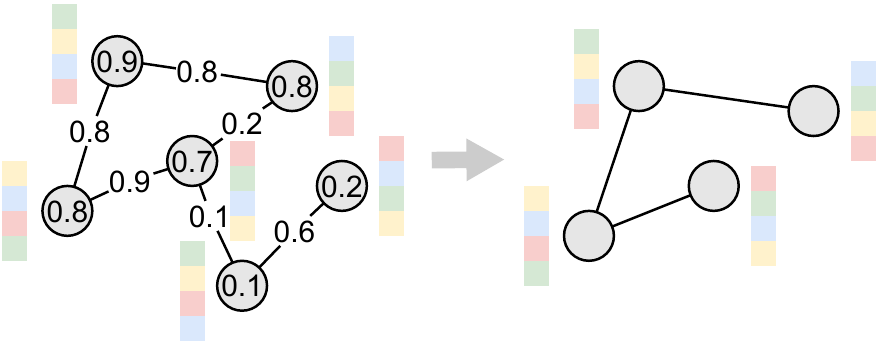}}
	\end{center}
	\vspace{-1em}
	\caption{A comparison of the feature-based, structue-based, sampling-based, and adaptive augmentation. The feature-based augmentation generally randomly (or manually) masks a small portion of node or edge attributes with constants or random values. The structue-based augmentation randomly (or manually) adds or removes small portions of edges from the graph, which includes methods like edge perturbation, node insertion, and edge diffusion. The sampling-based augmentation samples nodes and their connected edges from the graph under specific rules, which include Uniform Sampling, Ego-net Sampling, Random Walk Sampling, Importance Sampling, Knowledge Sampling, etc. The adaptive sampling adopts attention or gradient-based schemes to perform adaptive sampling based on the learned attention score or gradient magnitude. The numbers in the Fig.~3(d) are the importance scores of the nodes and edges, and we sample the most important 4 nodes and 3 edges as an example.}
	\vspace{-1em}
	\label{fig:3}
\end{figure*}

The recent works in the CV domain show that the success of contrastive learning relies heavily on well-designed data augmentation strategies, and in particular, certain kinds of augmentations play a very important role in improving performance. However, due to the inherent non-Euclidean properties of graph data, it is difficult to directly apply data augmentations designed for images to graphs. Here, we divide the data augmentation strategy for graph data into four categories: feature-based, structure-based, sampling-based, and adaptive augmentation. An overview of four types of augmentations is presented in Fig.~\ref{fig:3}.

\subsubsection{Feature-based Augmentation}
Given an input graph $(\mathbf{A},\mathbf{X})$, a feature-based augmentation only performs transformation on the node feature matrix $\mathbf{X}$ or edge feature matrix $\mathbf{X}^e$. Without loss of generality, we take $\mathbf{X}$ as an example, give by
\begin{equation}
    \widetilde{\mathbf{A}},\widetilde{\mathbf{X}}=\mathcal{T}(\mathbf{A},\mathbf{X})=\mathbf{A},\mathcal{T}_{\mathbf{X}}(\mathbf{X})
\end{equation}

\textbf{Attribute Masking.} The attribute masking \cite{you2020graph,jin2020self,hu2019strategies,zhu2020deep,thakoor2021bootstrapped} randomly masks a small portion of attributes. We specify $\mathcal{T}_{\mathbf{X}}(\mathbf{X})$ for the attribute masking as
\begin{equation}
    \mathcal{T}_{\mathbf{X}}(\mathbf{X})=\mathbf{X}\odot(1-\mathbf{L})+\mathbf{M}\odot\mathbf{L}
\end{equation}

\noindent where $\mathbf{L}$ is a masking location matrix where $\mathbf{L}_{i,j}=1$ if the $j$-th element of node $v_i$ is masked, otherwise $\mathbf{L}_{i,j}=0$. $\mathbf{M}$ denotes a masking value matrix. The matrix $\mathbf{L}$ is usually sampled by Bernoulli distribution or assigned manually. Besides, different schemes of $\mathbf{M}$ result in different augmentations. For example, $\mathbf{M}=\mathbf{0}$ denotes a constant masking, $\mathbf{M} \sim N(\mathbf{0}, \mathbf{\Sigma})$ replaces the original values by Gaussian noise and $\mathbf{M} \sim N(\mathbf{X}, \mathbf{\Sigma})$ adds Gaussian noise to the input.

\textbf{Attribute Shuffling.} The attribute shuffling \cite{velickovic2019deep,opolka2019spatio,ma2021improving,jing2021hdmi,ren2019heterogeneous} performs the row-wise shuffling on the attribute matrix $\mathbf{X}$. That is, the augmented graph consists of the same nodes as the original graph, but they are located in different places in the graph, and receive different contextual information. We specify $\mathcal{T}_{\mathbf{X}}(\mathbf{X})$ for the attribute shuffling as
\begin{equation}
    \mathcal{T}_{\mathbf{X}}(\mathbf{X})=\mathbf{X}[idx, :]
\end{equation}

\noindent where $idx$ is a list containing numbers from 1 to $N(N=|\mathcal{V}|)$, but with a random arrangement.

\subsubsection{Structue-based Augmentation}
Given a graph $(\mathbf{A},\mathbf{X})$, a structue-based augmentation only performs transformation on adjacent matrix $\mathbf{A}$, as follows
\begin{equation}
    \widetilde{\mathbf{A}},\widetilde{\mathbf{X}}=\mathcal{T}(\mathbf{A},\mathbf{X})=\mathcal{T}_{\mathbf{A}}(\mathbf{A}),\mathbf{X}
\end{equation}

\textbf{Edge Perturbation.} The edge perturbation \cite{you2020graph,zhu2020self,hu2020gpt,zhang2020iterative,zeng2020contrastive} perturbs structural connectivity through randomly adding or removing a certain ratio of edges. We specify $\mathcal{T}_{\mathbf{A}}(\mathbf{A})$ for the edge perturbation as
\begin{equation}
    \mathcal{T}_{\mathbf{A}}(\mathbf{A})=\mathbf{A}\odot(1-\mathbf{L})+(1-\mathbf{A})\odot\mathbf{L}
\end{equation}

\noindent where $\mathbf{L}$ is a perturbation location matrix where $\mathbf{L}_{i,j}=\mathbf{L}_{j,i}=1$ if the edge between node $v_i$ and $v_j$ will be perturbed, otherwise $\mathbf{L}_{i,j}=\mathbf{L}_{j,i}=0$. Different values in $\mathbf{L}$ result in different perturbation strategies, and more values set to 1 in $\mathbf{L}$, more server the perturbation is.

\textbf{Node Insertion.} The node insertion \cite{zeng2020contrastive} adds $K$ nodes $\mathcal{V}_{a}=\{v_{N+k}\}_{k=1}^K$ to node set $\mathcal{V}$ and add some edges between $\mathcal{V}_{a}$ and $\mathcal{V}$. For a structure transformation $\widetilde{\mathbf{A}}=\mathcal{T}_{\mathbf{A}}(\mathbf{A})$, we have $\widetilde{\mathbf{A}}_{:N,:N}=\mathbf{A}$. Given the connection ratio $r$, we have
\begin{equation}
p(\widetilde{\mathbf{A}}_{i,j}=\widetilde{\mathbf{A}}_{j,i}=1)=r, p(\widetilde{\mathbf{A}}_{i,j}=\widetilde{\mathbf{A}}_{j,i}=0)=1-r
\end{equation}

\noindent for $N+1 \leq i,j \leq N+K$.

\textcolor{mark}{\textbf{Edge Diffusion.} The edge diffusion \cite{kefato2021self,hassani2020contrastive} generates a different topological view of the original graph structure, with the general edge diffusion process defined as}
\begin{equation}
\mathcal{T}_{\mathbf{A}}(\mathbf{A})=\sum_{k=0}^{\infty} \Theta_{k} \mathbf{S}^{k}
\label{equ:diff}
\end{equation}

\noindent \textcolor{mark}{where $\mathbf{S} \in \mathbb{R}^{N \times N}$ is the generalized transition matrix and $\Theta$ is the weighting coefficient which satisfies $\sum_{k=0}^{\infty} \Theta_{k}=1, \Theta_{k} \in[0,1]$. Two instantiations of Equ.~\ref{equ:diff} are: (1) Personalized PageRank (PPR) with $\mathbf{S}=\mathbf{D}^{-1 / 2}\mathbf{A} \mathbf{D}^{-1 / 2}$ and $\Theta_{k}=\alpha(1-\alpha)^{k}$, and (2) Heat Kernel (HK) with $\mathbf{S}=\mathbf{A} \mathbf{D}^{-1}$ and $\Theta_{k}=e^{-t} t^{k} / k !$, where $\alpha$ denotes teleport probability in a random walk and $t$ is the diffusion time. The closed-form solutions of PPR and HK diffusion are formulated as}
\begin{equation}
\begin{small}
\begin{aligned}
\mathcal{T}_{\mathbf{A}}^{PPR}(\mathbf{A})=&\alpha\left(\mathbf{I}_{n}-(1-\alpha) \mathbf{D}^{-1 / 2} \mathbf{A} \mathbf{D}^{-1 / 2}\right)^{-1} & \\
\mathcal{T}_{\mathbf{A}}^{HK}(\mathbf{A})=&\exp\left(t\mathbf{A}\mathbf{D}^{-1}-t\right)
\end{aligned}
\end{small}
\end{equation}

\subsubsection{Sampling-based Augmentation}
Given an input graph $(\mathbf{A},\mathbf{X})$, a sampling-based augmentation performs transformation on both the adjacent matrix $\mathbf{A}$ and feature matrix $\mathbf{X}$, as follows
\begin{equation}
    \widetilde{\mathbf{A}},\widetilde{\mathbf{X}}=\mathcal{T}(\mathbf{A},\mathbf{X})=\mathbf{A}[\mathcal{S},\mathcal{S}],\mathbf{X}[\mathcal{S}, :]
\end{equation}

\noindent where $\mathcal{S}\in \mathcal{V}$ and existing methods usually apply five sampling strategies to obtain the node subset $\mathcal{S}$: uniform sampling, ego-nets sampling, random walk sampling, importance sampling, and knowledge-based sampling.

\textbf{Uniform Sampling.} The uniform sampling \cite{zeng2020contrastive} (a.k.a Node Dropping) uniformly samples a given number of nodes $\mathcal{S}$ from $\mathcal{V}$ and remove the remaining nodes directly.

\textbf{Ego-nets Sampling} \cite{hu2019strategies,zhu2020transfer,cao2021bipartite}. Given a typical graph encoder with $L$ layers, the computation of the node representation only depends on its $L$-hop neighborhood. In particular, for each node $v_i$, the transformation $\mathcal{T}(\cdot)$ samples the $L$-ego net surrounding node $v_i$, with $\mathcal{S}$ defined as
\begin{equation}
\mathcal{S}=\{v_j\mid d(v_i,v_j)\leq L\}
\end{equation}

\noindent where $d(v_i,v_j)$ is the shortest path length between node $v_i$ and $v_j$. The Ego-nets Sampling is essentially a special version of Breadth-First Search (BFS) sampling.

\textbf{Random Walk Sampling} \cite{qiu2020gcc,you2020graph,hassani2020contrastive}. It starts a random walk on graph $g$ from the ego node $v_i$. The walk iteratively travels to its neighborhood with the probability proportional to the edge weight. In addition, at each step, the walk returns back to the starting node $v_i$ with a positive probability $\alpha$. Finally, the visited nodes are collected into a node subset $\mathcal{S}$.

\textbf{Importance Sampling} \cite{jiao2020sub}.
Given a node $v_i$, we can sample a subgraph based on the importance of its neighboring nodes, with an importance score matrix $\mathbf{M}$ defined as
\begin{equation}
\mathbf{M}=\alpha \cdot(\mathbf{I}_n-(1-\alpha) \cdot \mathbf{A}\mathbf{D}^{-1})
\end{equation}

\noindent where $\alpha\in[0,1]$ is a hyperparameter. For a given node $v_i$, the subgraph sampler chooses top-$k$ important neighbors anchored by $v_i$ to constitute a subgraph with the index of chosen nodes denoted as $\mathcal{S}=\operatorname{top}_{-} \operatorname{rank}(\mathbf{M}(i,:), k)$. 

\textbf{Knowledge Sampling} \cite{zhang2020motif}.
The knowledge-based sampling incorporates domain knowledge into subgraph sampling. For example, the sampling process can be formalized as a \emph{library-based matching} problem by counting the frequently occurring and bioinformatics substructures in the molecular graph and building libraries (or tables) for them.

\subsubsection{Adaptive Augmentation}
The adaptive augmentation usually employs attention scores or gradients to guide the selection of nodes or edges.

\textbf{Attention-based.} The attention-based methods typically define importance scores for nodes or edges and then augment data based on their importance. For example, GCA \cite{zhu2020graph} proposes to keeps important structures and attributes unchanged, while perturbing possibly unimportant edges and features. Specifically, the probability of edge removal and feature masking should be closely related to their importance. Given a node centrality measure
$\varphi_{c}(\cdot): \mathcal{V} \rightarrow \mathbb{R}^{+}$, it defines edge centrality as the average of two adjacent nodes’ centrality scores, i.e., $s_{i,j}=\log\frac{\varphi_{c}(v_i)+\varphi_{c}(v_j)}{2}$. Then, the importance of the edge $e_{i,j}$ is defined as
\begin{equation}
p_{i,j}=\min \left(\frac{s_{\max }-s_{i,j}}{s_{\max }-\mu_{s}} \cdot p_{e}, p_{\tau}\right)
\end{equation}

\noindent where $p_{e}$ is a hyperparameter that controls the overall probability of removing edges, $s_{\max }$ and $\mu_{s}$ is the maximum and average of $\{s_{i,j}\}_{j=1}^N$ and $p_{\tau}<1$ is a cut-off probability, used to truncate the probabilities since extremely high removal probabilities will overly corrupt graph structures. The node centrality can be defined as degree centrality, Eigenvector centrality \cite{bonacich1987power}, or PageRank centrality \cite{page1999pagerank}, which results in three variants. The attribute masking based on node importance is the same as above and will not be repeated.

\textbf{Gradient-based.} Unlike the simple uniform edge removal and insertion as in GRACE \cite{zhu2020deep}, GROC \cite{jovanovic2021towards} adaptively performs gradient-based augmentation guided by edge gradient information. Specifically, it first applies two stochastic transformations $\mathcal{T}_1(\cdot)$ and $\mathcal{T}_2(\cdot)$ to graph $g=(\mathbf{A},\mathbf{X})$ to obtain two views, masking node attributes independently with probability $r_1$ and $r_2$ and then computing the contrastive loss $\mathcal{L}_{ssl}$ between these two views. For a given node $v_i$, an edge removal candidate set is defined as
\begin{equation}
\begin{small}
\begin{aligned}
\mathcal{S}^{-}=\left\{(v_i,v_k)\Big|v_k\in\mathcal{N}_i^{(l)}\right\}    
\end{aligned}
\end{small}
\end{equation}

\noindent, and an edge insertation candidate set is defined as 
\begin{equation}
\begin{small}
\begin{aligned}
\mathcal{S}^{+}=\left\{(v_i,v_k)\Big|v_k\in\left(\cup_{v_m \in \mathcal{B}} \mathcal{N}_m^{(l)}\backslash\mathcal{N}_i^{(l)}\right)\right\}  
\end{aligned}
\end{small}
\end{equation}

\noindent where $\mathcal{B} \subset \mathcal{V}$ is a node batch. $\mathcal{S}^{+}$ is restricted to the set of edges $(v_i,v_k)$ where $v_i$ is an anchor node, and $v_k$ is within the $l$-hop neighborhood of some other anchors $v_m \neq v_i$ but not within the $l$-hop neighborhood of node $v_i$. Finally, we backpropagate the loss $\mathcal{L}_{ssl}$ to obtain gradient intensity values for each edge in $\mathcal{S}^{-}$ and $\mathcal{S}^{+}$. A further gradient-based adaptive augmentations are applied on the views by removing a subset of edges with \emph{minimal} edge gradient magnitude values in $\mathcal{S}^{-}$ and inserting a subset of edges with the \emph{maximal} edge gradient magnitude values in $\mathcal{S}^{+}$.

\begin{figure*}[!htbp]
	\begin{center}
	\includegraphics[width=0.8\linewidth]{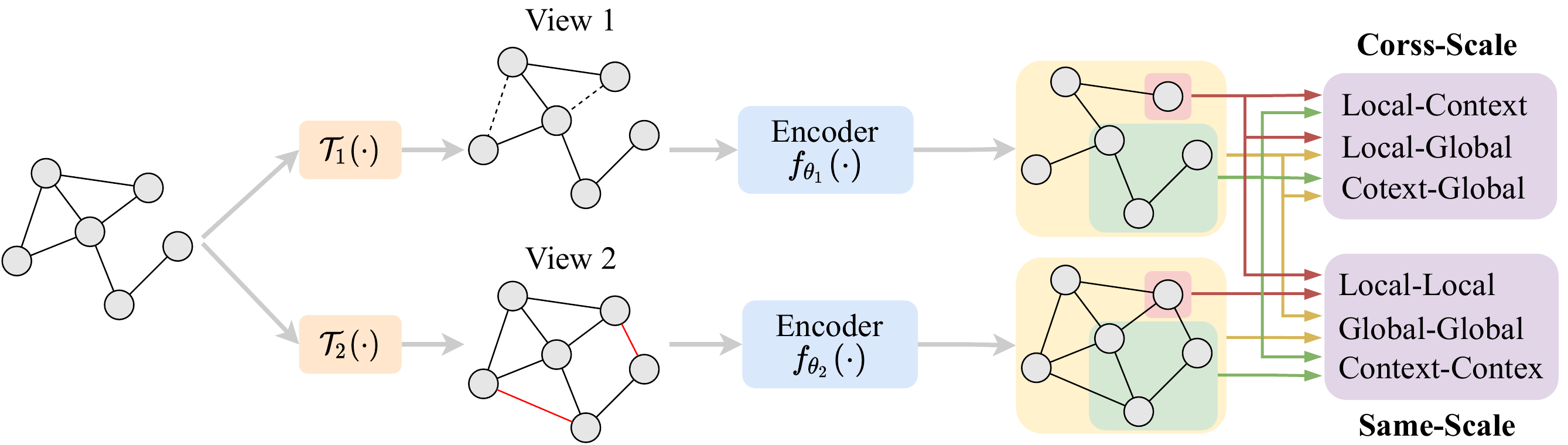}
	\end{center}
	\vspace{-1em}
	\caption{A general framework for contrastive learning methods with three main modules: data augmentation strategies, pretext tasks, and contrastive objectives. Different views can be generated by a single or a combination of augmentations $\mathcal{T}_1(\cdot)$ and $\mathcal{T}_2(\cdot)$. For graph encoder $f_{\theta_1}(\cdot)$ and $f_{\theta_2}(\cdot)$, the commonly used graph neural networks include GAE \cite{kipf2016variational}, VGAE \cite{kipf2016variational}, etc. However, the design of graph encoder is not the focus of graph SSL and thus beyond the scope of this survey. The two contrasting views may be local, contextual, or global, corresponding to node-level (marked in red), subgraph-level (marked in green), or graph-level (marked in yellow) information in the graph. The contrastive learning can thus contrast two views \emph {at the same or different scales}, which leads to two categories of algorithm: (1) same-scale contrasting, including local-local, context-context, and global-global contrasting; and (2) cross-scale contrasting, including local-context, local-global, and context-global contrasting.}
	\vspace{-1em}
	\label{fig:4}
\end{figure*}

\subsection{Pretext Task}
The contrastive learning aims to maximize the agreement of two jointly sampled positive pairs. Depending on the definition of a graph view, the scale of the view may be local, contextual, or global, corresponding to the node-level, subgraph-level, or graph-level information in the graph. Therefore, contrastive learning may contrast two graph views \emph{at the same or different scales}, which leads to two categories: (1) Contrasting with the same-scale and (2) Contrasting with the cross-scale. The two views in the same-scale contrasting, either positive or negative pairs, are at the same scale, such as node-node and graph-graph pairs, while the two views in the cross-scale contrasting have different scales, such as node-subgraph or node-graph contrasting. We categorize existing methods from these two perspectives and present them in a unified framework as shown Fig.~\ref{fig:4}. In this section, due to space limitations, we present only some representative contrastive methods and place those {relatively less important} works in \textbf{Appendix A}.

\subsubsection{Contrasting with the same-scale}
The same-scale contrastive learning is further refined into three categories based on the different scales of the views: local-local, context-context, and global-global contrasting.
\newline

\noindent \textbf{\textit{3.3.1.1 Global-Global Contrasting}}
\newline

\textbf{GraphCL} \cite{you2020graph}. Four types of graph augmentations $\{\mathcal{T}_k\}_{k=1}^4$ are applied to incorporate various priors: (1) Node Dropping $\mathcal{T}_1(\cdot)$; (2) Edge Perturbation $\mathcal{T}_2(\cdot)$; (3) Attribute Masking $\mathcal{T}_3(\cdot)$; (4) Subgraph Sampling $\mathcal{T}_4(\cdot)$. Given a graph $g_i=(\mathbf{A}_i,\mathbf{X}_i)\in\mathcal{G}$, it first applies a series of graph augmentations $\mathcal{T}(\cdot)$ randomly selected from $\{\mathcal{T}_k\}_{k=1}^4$ to generate an augmented graph $\widetilde{g}_i=(\widetilde{\mathbf{A}}_i,\widetilde{\mathbf{X}}_i)=\mathcal{T}(\mathbf{A}_i,\mathbf{X}_i)$, and then learns to predict whether two graphs originate from the same graph or not. Specifically, a shared graph-level encoder $f_{\gamma}(\cdot)$ is applied to obtain graph-level representations $\mathbf{h}_{g_i}=f_{\gamma}(\mathbf{A}_i,\mathbf{X}_i)$ and $\widetilde{\mathbf{h}}_{\widetilde{g}_i}=f_{\gamma}(\widetilde{\mathbf{A}}_i,\widetilde{\mathbf{X}}_i)$, respectively. Finally, the learning objective is defined as follows
\begin{equation}
\max _{\theta} \frac{1}{|\mathcal{G}|} \sum_{g_i\in\mathcal{G}}\mathcal{MI}\left(\mathbf{h}_{g_{i}}, \widetilde{\mathbf{h}}_{\widetilde{g}_i}\right)
\end{equation}

\textbf{Contrastive Self-supervised Learning (CSSL)} \cite{zeng2020contrastive} follows a very similar framework to GraphGL, differing only in the way the data is augmented. Along with node dropping, it also considers node insertion as an important augmentation strategy. Specifically, it randomly selects a strongly-connected subgraph $S$, removes all edges in $S$, adds a new node $v_i$, and adds an edge between $v_i$ and each node in $S$.

\textbf{Label Contrastive Coding (LCC)} \cite{ren2021label} is proposed to encourage intra-class compactness and inter-class separability. To power contrastive learning, LLC introduces a dynamic label memory bank and a momentum updated encoder. Specifically, the query graph $(g_q, y_q)$ and key graph $(g_k, y_k)$ are encoded by
two graph-level encoder $f_{\gamma_q}(\cdot)$ and $f_{\gamma_k}(\cdot)$ to obtain graph-level representations $\mathbf{h}_{g_q}$ and $\mathbf{h}_{g_k}$ respectively. If $\mathbf{h}_{g_q}$ and $\mathbf{h}_{g_k}$ have the same label, they are considered as the positive pair, otherwise, they are the negative pair. The label contrastive loss encourages the model to distinguish the positive pair from the negative pair. For the encoded query $(g_q, y_q)$, its label contrastive loss is calculated by
\begin{equation}
\max _{\gamma_q} \log \frac{\sum_{i=1}^{m} \mathbb{I}_{y_{i}=y_q} \cdot \exp \left(\mathbf{h}_{g_q} \cdot \mathbf{h}_{g_k}^{(i)} / \tau\right)}{\sum_{i=1}^{m} \exp \left(\mathbf{h}_{g_q} \cdot \mathbf{h}_{g_k}^{(i)} / \tau\right)}
\end{equation}

\noindent where $m$ is the size of memory bank, $\tau$ is the temperature hyperparameter, and $\mathbb{I}_{y_{i}=y_q}$ is an indicator function to determine whether the label of $i$-th key graph $g_k^{(i)}$ in the memory bank is the same as $y_q$. The parameter $\gamma_k$ of $f_{\gamma_k}(\cdot)$ follows a momentum-based update mechanism as Moco \cite{he2020momentum}, given by
\begin{equation}
\gamma_{k} \longleftarrow \alpha \gamma_{k}+(1-\alpha) \gamma_{q}
\label{equ:ema}
\end{equation}

\noindent where $\alpha\in[0,1)$ is the momentum weight to control the speed of $\gamma_k$ evolving.
\newline

\noindent \textbf{\textit{3.3.1.2 Context-Context Contrasting}}
\newline

\textbf{Graph Contrastive Coding (GCC)} \cite{qiu2020gcc} is a graph self-supervised pre-training framework, that captures the universal graph topological properties across multiple graphs. Specifically, it first samples multiple subgraphs for each graph $g\in\mathcal{G}$ by random walk and collect them in to a memory bank $\mathcal{S}$. Then the query subgraph $g_q\in\mathcal{S}$ and key subgraph $g_k\in\mathcal{S}$ are encoded by two graph-level encoders
$f_{\gamma_q(\cdot)}$ and $f_{\gamma_k}(\cdot)$ to obtain graph-level representations $\mathbf{h}_{g_q}$ and $\mathbf{h}_{g_k}$, respectively. If $g_q$ and $g_k$ are sampled from the same graph, they are considerd as the positive pair, otherwise they are the negative pair. For the encoded query $(g_q, y_q)$ where $y_q$ is the index of graph it sampled from, its graph contrastive loss is calculated by
\begin{equation}
\max _{\gamma_q} \log \frac{\sum_{i=1}^{|\mathcal{S}|} \mathbb{I}_{y_{i}=y_q} \cdot \exp \left(\mathbf{h}_{g_q} \cdot \mathbf{h}_{g_k}^{(i)} / \tau\right)}{\sum_{i=1}^{|\mathcal{S}|} \exp \left(\mathbf{h}_{g_q} \cdot \mathbf{h}_{g_k}^{(i)} / \tau\right)}
\end{equation}

\noindent where $\mathbb{I}_{y_{i}=y_q}$ is an indicator function to determine whether the $i$-th key graph $g_k^{(i)}$ in the memory bank and query graph $g_q$ are \emph{sampled from the same graph}. The parameter $\gamma_k$ of $f_{\gamma_k}(\cdot)$ follows a momentum-based updating as in Equ.~\ref{equ:ema}.
\newline

\noindent \textbf{\textit{3.3.1.3 Local-Local Contrasting}}
\newline

\textbf{GRACE} \cite{zhu2020deep}. Rather than contrasting global-global views as GraphCL \cite{you2020graph} and CSSL \cite{zeng2020contrastive}, GRACE focuses on contrasting views at the node level. Given a graph $g=(\mathbf{A},\mathbf{X})$, it first generates two augmentatd graphs $g^{(1)}=(\mathbf{A}^{(1)},\mathbf{X}^{(1)})=\mathcal{T}_1(\mathbf{A},\mathbf{X})$ and $g^{(2)}=(\mathbf{A}^{(1)},\mathbf{X}^{(2)})=\mathcal{T}_2(\mathbf{A},\mathbf{X})$. Then it applies a shared encoder $f_\theta(\cdot)$ to generate their node embedding matrices $\mathbf{H}^{(1)}=f_{\theta}(\mathbf{A}^{(1)},\mathbf{X}^{(1)})$ and $\mathbf{H}^{(2)}=f_{\theta}(\mathbf{A}^{(2)},\mathbf{X}^{(2)})$. Finally, the pairwise objective for each positive pair $(\mathbf{h}^{(1)}_i, \mathbf{h}^{(2)}_i)$ is defined as follows
\begin{equation}
\mathcal{L}(\mathbf{h}^{(1)}_i, \mathbf{h}^{(2)}_i) = \log \frac{e^{\mathcal{D}(\mathbf{h}^{(1)}_i, \mathbf{h}^{(2)}_i) / \tau}}{e^{ \mathcal{D}(\mathbf{h}^{(1)}_i, \mathbf{h}^{(2)}_i) / \tau}+Neg}
\end{equation}

\noindent where $Neg$ is defined as
\begin{equation}
Neg=\sum_{k=1}^{N} \mathbf{1}_{k\neq i} \left[e^{ \mathcal{D}(\mathbf{h}^{(1)}_i, \mathbf{h}^{(1)}_k) / \tau}+e^{\mathcal{D}(\mathbf{h}^{(1)}_i, \mathbf{h}^{(2)}_k) / \tau}\right]
\end{equation}

\noindent where $e^{ \mathcal{D}(\mathbf{h}^{(1)}_i, \mathbf{h}^{(1)}_k) / \tau}$ is the \emph{intra-view} negative pair and $e^{\mathcal{D}(\mathbf{h}^{(1)}_i, \mathbf{h}^{(2)}_k) / \tau}$ is the \emph{inter-view} negative pair. The overall objective to be maximized is then defined as,
\begin{equation}
\max_{\theta}\frac{1}{2N}\sum_{i=1}^{N}\left[\mathcal{L}(\mathbf{h}^{(1)}_i, \mathbf{h}^{(2)}_i)+\mathcal{L}(\mathbf{h}^{(2)}_i, \mathbf{h}^{(1)}_i)\right]
\end{equation}

\textbf{GCA} \cite{zhu2020graph} and \textbf{GROC} \cite{jovanovic2021towards} adopt the same framework and objective as GRACE but with more flexible and \emph{adaptive} data augmentation strategies. \textcolor{mark}{The framework proposed by SEPT \cite{yu2021socially} is similar to GRACE, but it is specifically designed for the specific downstream task (recommendation) by combining cross-view contrastive learning with semi-supervised tri-training. Technically, SEPT first augments the user data with the user social information, and then it builds three graph encoders upon the augmented views, with one for recommendation and the other two used to predict unlabeled users. Given a certain user, SEPT takes those nodes whose predicted labels are highly consistent with the target user as positive samples and then encourages the consistency between the target user and positive samples.}

\textbf{Cross-layer Contrasting (GMI)} \cite{peng2020graph}. Given a graph $g=(\mathbf{A},\mathbf{X})$, a graph encoder $f_\theta(\cdot)$ is applied to obtain the node embedding matrix $\mathbf{H}=f_\theta(\mathbf{A},\mathbf{X})$. Then the Cross-layer Node Contrasting can be defined as 
\begin{equation}
\max _{\theta} \frac{1}{N} \sum_{i=1}^{N}\mathcal{MI}\left(\mathbf{h}_i, \mathbf{x}_i\right)
\end{equation}

\noindent where the negative samples to contrast with $\mathbf{h}_i$ is $Neg(\mathbf{h}_i)=\{\mathbf{x}_j \mid v_j \in \mathcal{N}_i\}$. Similarly, the Cross-layer Edge Contrasting can be defined as 
\begin{equation}
\max _{\theta} \frac{1}{N} \sum_{i=1}^{N}\sum_{v_j \in \mathcal{N}_i}\mathcal{MI}\left(\mathbf{w}_{i,j}, \mathbf{A}_{i,j}\right)
\end{equation}

\noindent where $\mathbf{w}_{i,j}=\sigma(\mathbf{h}_{i}\mathbf{h}_{j}^T)$, and the negative samples to contrast with $\mathbf{w}_{i,j}$ are $Neg(\mathbf{w}_{i,j})=\{\mathbf{A}_{i,k} \mid v_k \in \mathcal{N}_i \quad \text{and} \quad k\neq j\}$.

\textbf{STDGI} \cite{opolka2019spatio} extents the idea of mutual information maximization to spatial-temporal graphs. Specifically, given two graphs $g_t=(\mathbf{A},\mathbf{X}^{(t)})$ and $g_{t+k}=(\mathbf{A},\mathbf{X}^{(t+k)})$ at the time $t$ and  $t+k$, a shared graph encoder $f_\theta(\cdot)$ is applied to obtain the node embedding matrix $\mathbf{H}^{(t)}=f_\theta(\mathbf{A},\mathbf{X}^{(t)})$. Besides, it generates an augmentatd graph $\widetilde{g}_{t+k}=(\mathbf{A},\widetilde{\mathbf{X}}^{(t+k)})=\mathcal{T}(\mathbf{A},\mathbf{X}^{(t+k)})$ by randomly permuting the node features. Finally, the learning objective is defined as follows
\begin{equation}
\max _{\theta} \frac{1}{N} \sum_{i=1}^{N}\mathcal{MI}\left(\mathbf{h}^{(t)}_i, \mathbf{x}^{(t+k)}_i\right)
\end{equation}

\noindent where the negative samples to contrast with $\mathbf{h}^{(t)}_i$ is $Neg(\mathbf{h}^{(t)}_i)=\widetilde{\mathbf{x}}^{(t+k)}_i$.

\textbf{BGRL} \cite{thakoor2021bootstrapped}. Inspired by BYOL, BGRL proposes to perform the self-supervised learning that \emph{does not require negative samples}, thus getting rid of the potentially quadratic bottleneck. Specifically, given a graph $g=(\mathbf{A},\mathbf{X})$, it first generates two augmentatd graph views $g^{(1)}=(\mathbf{A}^{(1)},\mathbf{X}^{(1)})=\mathcal{T}_1(\mathbf{A},\mathbf{X})$ and $g^{(2)}=(\mathbf{A}^{(1)},\mathbf{X}^{(2)})=\mathcal{T}_2(\mathbf{A},\mathbf{X})$. Then it applies two graph encoders $f_{\theta_1}(\cdot)$ and $f_{\theta_2}(\cdot)$ to generate their node embedding matrices $\mathbf{H}^{(1)}=f_{\theta_1}(\mathbf{A}^{(1)},\mathbf{X}^{(1)})$ and $\mathbf{H}^{(2)}=f_{\theta_2}(\mathbf{A}^{(2)},\mathbf{X}^{(2)})$. Moreover, a node-level prediction head $g_\omega(\cdot)$ is used to output $\mathbf{Z}^{(1)}=g_{\omega}(\mathbf{H}^{(1)})$. Finally, the learning objective is defined as follows
\begin{equation}
\max_{\theta_1,\omega}\frac{1}{N}\sum_{i=1}^{N} \frac{\mathbf{z}^{(1)}_i(\mathbf{h}^{(2)}_i)^T}{\|\mathbf{z}^{(1)}_i\|\|\mathbf{h}^{(2)}_i\|}
\label{equ:bgrl}
\end{equation}

\noindent where the parameter $\theta_2$ are updated as an exponential moving average (EMA) of parameters $\theta_1$, as done in Equ.~\ref{equ:ema}. 

\textbf{SelfGNN} \cite{kefato2021self} differs from BGRL only in the definition of the objective function. Unlike Equ.~\ref{equ:bgrl}, SelfGNN defines the implicit contrastive term directly in the form of MSE, 
\begin{equation}
\min_{\theta_1,\omega}\frac{1}{N}\sum_{i=1}^{N} \|\mathbf{z}^{(1)}_i-\mathbf{h}^{(2)}_i\|^2
\end{equation}

\textcolor{mark}{\textbf{HeCo} \cite{wang2021self}. Consider a meta-path $\Phi_k$ form the meta-path set $\{\Phi_k\}_{k=1}^K$, if there exist a meta-path $\Phi_k$ between node $v_i$ and node $v_j$, then $v_j$ can be considered as in the meta-path neighborhood $\mathcal{N}_i^{\Phi_k}$of node $v_i$, which yields a meta-path based adjacent matrix $\mathbf{A}^{\Phi_k}$. The HeCo first applies two graph encoder $f_{\theta_1}^{sc}(\cdot)$ and $f_{\theta_2}^{mp}(\cdot)$ to obtain node embedding matrices $\mathbf{H}^{sc}=f_{\theta_1}^{sc}(\mathbf{A},\mathbf{X})$ and $\mathbf{H}^{mp}=f_{\theta_2}^{ml}(\{\mathbf{A}^{\Phi_{k}}\}_{k=1}^K,\mathbf{X})$. To define positive and negative samples, HeCo first defines a function $\mathbb{C}_{i}(j)=\sum_{k=1}^{K} \mathbb{I}\left(j \in \mathcal{N}_{i}^{\Phi_{k}}\right)$ to count the number of meta-paths connecting nodes $v_i$ and $v_j$. Then it constructs a set $\mathcal{S}_i=\left\{j \mid j \in \mathcal{V} \text { and } \mathbb{C}_{i}(j) \neq 0\right\}$ and sort it in the descending order based on the value of $\mathbb{C}_{i}(j)$. Next it selects the top $T_{pos}$ nodes from $\mathcal{S}_i$ as positive samples $\mathbb{P}_i$ and treat the rest as negative samples $\mathbb{N}_i$ directly. Finally, the learning objective can be defined as follows}
\begin{equation}
\max _{\theta_1, \theta_2} \frac{1}{N} \sum_{i=1}^{N} \log \frac{\sum_{v_j \in \mathbb{P}_i} e^{\mathcal{D}(\mathbf{h}^{sc}_i, \mathbf{h}^{mp}_j) / \tau}}{\sum_{v_k \in \{\mathbb{P}_i \cup \mathbb{N}_i\}} e^{\mathcal{D}(\mathbf{h}^{sc}_i, \mathbf{h}^{mp}_k) / \tau} }
\end{equation}

\subsubsection{Contrasting with the cross-scale}
Based on different scales of two views, we further refined the scope of cross-scale contrastive into three categories: local-global, local-context, and context-global contrasting.
\newline

\noindent \textbf{\textit{3.3.2.1 Local-Global Contrasting}}
\newline

\textbf{Deep Graph Infomax (DGI)} \cite{velickovic2019deep} is proposed to contrast the patch representations and corresponding high-level summary of graphs. First, it applies an augmentation transformation $\mathcal{T}(\cdot)$ to obtain an augmented graph $\widetilde{g}=(\widetilde{\mathbf{A}},\widetilde{\mathbf{X}}) = \mathcal{T}(\mathbf{A},\mathbf{X})$. Then it passes these two graphs through two graph encoder $f_{\theta_1}(\cdot)$ and $f_{\theta_2}(\cdot)$ to obtain node embedding matrices $\widetilde{\mathbf{H}}=f_{\theta_1}(\widetilde{\mathbf{A}},\widetilde{\mathbf{X}})$ and $\mathbf{H}=f_{\theta_2}(\mathbf{A},\mathbf{X})$, respectively. Beside, a $\operatorname{READOUT}$ function is applied to obtain the graph-level representaion $\widetilde{\mathbf{h}}_{\widetilde{g}}=\operatorname{READOUT}(\widetilde{\mathbf{H}})$. Finally, the learning objective is defined as follows
\begin{equation}
\max _{\theta_1,\theta_2} \frac{1}{N} \sum_{v_i\in\mathcal{V}}\mathcal{MI}\left(\widetilde{\mathbf{h}}_{\widetilde{g}}, \mathbf{h}_i\right)
\end{equation}

\noindent where $\mathbf{h}_i$ is the node embedding of node $v_i$, and the negative samples to contrast with $\widetilde{\mathbf{h}}_{\widetilde{g}}$ is $Neg(\widetilde{\mathbf{h}}_{\widetilde{g}})=\{\mathbf{h}_{j}\}_{v_j\in\mathcal{V},j\neq i}$.

\textbf{MVGRL} \cite{hassani2020contrastive} maximize the the mutual information between the cross-view representations of nodes and graphs. Given a $g=(\mathbf{A},\mathbf{X})\in\mathcal{G}$, it first applies an augmentation to obtain $\widetilde{g}=(\widetilde{\mathbf{A}},\widetilde{\mathbf{X}})=\mathcal{T}(\mathbf{A},\mathbf{X})$ and then samples two subgraph $g^{(1)}=(\mathbf{A}^{(1)},\mathbf{X}^{(1)})=\mathcal{T}_1(\mathbf{A},\mathbf{X})$ and $g^{(2)}=(\mathbf{A}^{(2)},\mathbf{X}^{(2)})=\mathcal{T}_2(\mathbf{A},\mathbf{X})$ from it. Then two graph encoder $f_{\theta_1}(\cdot)$ and $f_{\theta_2}(\cdot)$ and a prejection head $g_{\omega_1}(\cdot)$ are applied to obtain node embedding matrices $\mathbf{H}^{(1)}=g_{\omega_1}(f_{\theta_1}(\mathbf{A}^{(1)},\mathbf{X}^{(1)}))$ and $\mathbf{H}^{(2)}=g_{\omega_1}(f_{\theta_2}(\mathbf{A}^{(2)},\mathbf{X}^{(2)}))$. In addition, a $\operatorname{READOUT}$ function and another prejection head $g_{\omega_2}(\cdot)$ are use to obtain graph-level representations $\mathbf{h}_g^{(1)}=f_{\omega_2}(\operatorname{READOUT}(\mathbf{H}^{(1)}))$ and $\mathbf{h}_g^{(2)}=f_{\omega_2}(\operatorname{READOUT}(\mathbf{H}^{(2)}))$. The learning objective is defined as follows:
\begin{equation}
\begin{small}
\begin{aligned}
\max _{\theta_1, \theta_2, \omega_1, \omega_2} \frac{1}{N} \sum_{v_i\in\mathcal{V}}\left[\mathcal{MI}(\mathbf{h}_{g}^{(1)}, \mathbf{h}_{i}^{(2)})+\mathcal{MI}(\mathbf{h}_{g}^{(2)}, \mathbf{h}_{i}^{(1)})\right]
\end{aligned}
\end{small}
\end{equation}

\noindent where the negative samples to contrast with $\mathbf{h}_{g}^{(1)}$ is $Neg(\mathbf{h}_{g}^{(1)})=\{\mathbf{h}_{j}^{(2)}\}_{v_j\in\mathcal{V},j\neq i}$ and the negative samples to contrast with $\mathbf{h}_{g}^{(2)}$ is $Neg(\mathbf{h}_{g}^{(2)})=\{\mathbf{h}_{j}^{(1)}\}_{v_j\in\mathcal{V},j\neq i}$.
\newline

\noindent \textbf{\textit{3.3.2.2 Local-Context Contrasting}}
\newline

\textbf{SUBG-CON} \cite{jiao2020sub} is proposed by utilizing the strong correlation between central (anchor) nodes and their surrounding subgraphs to capture contextual structure information. Given a graph $g=(\mathbf{A},\mathbf{X})$, SUBG-CON first picks up an anchor node set $\mathcal{S}$ from $\mathcal{V}$ and then samples their context subgraph $\{g_i=(\mathbf{A}^{(i)},\mathbf{X}^{(i)})\}_{i=1}^{|\mathcal{S}|}$ by the importance sampling strategy. Then a shared graph encoder $f_\theta(\cdot)$ and a $\operatorname{READOUT}$ function are applied to obtain node embedding matrices $\{\mathbf{H}^{(1)},\mathbf{H}^{(2)},\cdots,\mathbf{H}^{(|\mathcal{V}|)}\}$ where $\mathbf{H}^{(i)}=f_\theta(\mathbf{A}^{(i)},\mathbf{X}^{(i)})$ and graph-level representations $\{\mathbf{h}_{g_1},\mathbf{h}_{g_2},\cdots,\mathbf{h}_{g_{|\mathcal{V}|}}\}$ where $\mathbf{h}_{g_i}=\operatorname{READOUT}(\mathbf{H}^{(i)})$. Finally, the learning objective is defined as follows
\begin{equation}
\max _{\theta} \frac{1}{|\mathcal{S}|} \sum_{v_i\in\mathcal{S}}\mathcal{MI}\left(\mathbf{h}^{(i)}_i, \mathbf{h}_{g_i}\right)
\end{equation}

\noindent where $\mathbf{h}^{(i)}_i$ is the node representation of anchor node $v_i$ in the node embedding matrix $\mathbf{H}^{(i)}$. The negative samples to contrast with $\mathbf{h}^{(i)}_i$ is $Neg(\mathbf{h}^{(i)}_i)=\{\mathbf{h}_{g_j}\}_{v_j\in\mathcal{S},j\neq i}$.

\textbf{Graph InfoClust (GIC)} \cite{mavromatis2020graph} relies on a framework similar to DGI \cite{velickovic2019deep}. However, in addition to contrast local-global views, GIC also maximize the MI between node representations and their corresponding cluster embeddings. Given a graph $g=(\mathbf{A},\mathbf{X})$, it first applies an augmentation to obtain $\widetilde{g}=(\widetilde{\mathbf{A}},\widetilde{\mathbf{X}})=\mathcal{T}(\mathbf{A},\mathbf{X})$. Then a shared graph encoder $f_{\theta}(\cdot)$ is applied to obtain node embedding matrices $\mathbf{H}=f_{\theta}(\mathbf{A},\mathbf{X})$ and $\widetilde{\mathbf{H}}=f_{\theta}(\widetilde{\mathbf{A}},\widetilde{\mathbf{X}})$. Furthermore, an unsupervised clustering algorithm is used to group nodes into $K$ clusters $\mathcal{C}=\{C_1,C_2,\cdots,C_K\}$, and it obtains the cluster centers by $\bm{\mu}_k=\frac{1}{|C_k|}\sum_{v_i\in C_k}\mathbf{h}_i$ where $1 \leq k \leq K$. To compute the cluster embedding $\mathbf{z}_i$ for each node $v_i$, it applies a weighted average of the summaries of the cluster centers to which node $v_i$ belongs $\mathbf{z}_{i}=\sigma\left(\sum_{k=1}^{K} r_{i k} \bm{\mu}_{k}\right)$, where $r_{ik}$ is the probability that node $v_i$ is assigned to cluster $k$, and is a soft-assignment value with $\sum_{k} r_{i k}=1, \forall i$. For example, $r_{i,k}$ can be defined as $r_{i,k}=\frac{\exp(\mathbf{h}_i\bm{\mu}_k^T)}{\sum_{j=1}^K\exp(\mathbf{h}_i\bm{\mu}_j^T)}$. Finally, the learning objective is defined as follows
\begin{equation}
\max _{\theta} \frac{1}{N} \sum_{v_i\in\mathcal{V}}\mathcal{MI}\left(\mathbf{h}_i, \mathbf{z}_i\right)
\end{equation}

\noindent where the negative samples to contrast with $\mathbf{h}_i$ is $Neg(\mathbf{h}_i)=\{\mathbf{z}_j\}_{v_j\in\mathcal{V},j\neq i}$.
\newline

\noindent \textbf{\textit{3.3.2.3 Context-Global Contrasting}}
\newline

\textbf{MICRO-Graph} \cite{zhang2020motif}. The key challenge to conducting subgraph-level contrastive is to sample semantically informative subgraphs. For molecular graphs, the graph motifs, which are frequently-occurring subgraph patterns (e.g., functional groups) can be exploited for better subgraph sampling. Specifically, the motif learning is formulated as a differentiable clustering problem, and EM-clustering is adopted to group significant subgraphs into several motifs, thus obtaining a \emph{motifs table}. Given two graph $g^{(1)}=(\mathbf{A}^{(1)},\mathbf{X}^{(1)}), g^{(2)}=(\mathbf{A}^{(2)},\mathbf{X}^{(2)})\in\mathcal{G}$, it first applies a shared graph encoder $f_\theta(\cdot)$ to learn their node embedding matrices $\mathbf{H}^{(1)}=f_{\theta}(\mathbf{A}^{(1)},\mathbf{X}^{(1)})$ and $\mathbf{H}^{(2)}=f_{\theta}(\mathbf{A}^{(2)},\mathbf{X}^{(2)})$. Then it leverages learned motifs table to sample $K$ motif-like subgraphs from $g^{(1)}$ and $g^{(2)}$ and obtain their correspongding embedding matrices $\{\mathbf{H}^{(1)}_1,\mathbf{H}^{(1)}_2,\cdots,\mathbf{H}^{(1)}_K\}$ and $\{\mathbf{H}^{(2)}_1,\mathbf{H}^{(2)}_2,\cdots,\mathbf{H}^{(2)}_K\}$. Then a $\operatorname{READOUT}$ function is applied to obtain graph-level and subgraph-level representations, denoted as $\mathbf{h}^{(1)}_g$, $\{\mathbf{h}^{(1)}_1,\mathbf{h}^{(1)}_2,\cdots,\mathbf{h}^{(1)}_K\}$ and $\mathbf{h}^{(2)}_g$, $\{\mathbf{h}^{(2)}_1,\mathbf{h}^{(2)}_2,\cdots,\mathbf{h}^{(2)}_K\}$. Finally, the objective is defined as
\begin{equation}
\begin{small}
\begin{aligned}
\max _{\theta} \frac{1}{|\mathcal{G}|} \sum_{g \in \mathcal{G}}\sum_{k=1}^{K}\left[\mathcal{MI}(\mathbf{h}_{g}^{(1)}, \mathbf{h}_{k}^{(1)})+\mathcal{MI}(\mathbf{h}_{g}^{(2)}, \mathbf{h}_{k}^{(2)})\right]
\end{aligned}
\end{small}
\end{equation}

\noindent where the negative samples to contrast with $\mathbf{h}_{g}^{(1)}$ is $Neg(\mathbf{h}_{g}^{(1)})=\{\mathbf{h}_{j}^{(2)}\}_{j=1}^K$ and the negative samples to contrast with  $\mathbf{h}_{g}^{(2)}$ is $Neg(\mathbf{h}_{g}^{(2)})=\{\mathbf{h}_{j}^{(1)}\}_{j=1}^K$.

\textbf{InfoGraph} \cite{sun2019infograph} aims to obtain embeddings \emph{at the whole graph level} for self-supervised learning. Given a graph $g=(\mathbf{A},\mathbf{X})$, it first applies an augmentation to obtain $\widetilde{g}=(\widetilde{\mathbf{A}},\widetilde{\mathbf{X}})=\mathcal{T}(\mathbf{A},\mathbf{X})$. Then a shared $L$-layer graph encoder $f_\theta(\cdot)$ is applied to learn node embedding matrix sequences  $\{\mathbf{H}^{(l)}\}_{l=1}^L$ and $\{\widetilde{\mathbf{H}}^{(l)}\}_{l=1}^L$ obtain from \emph{each layer}. Then it concats the representations learned from each layer, $\mathbf{h}_i=\text{CONCAT}(\{\mathbf{h}^{(l)}_{i}\}_{l=1}^L)$ and $\widetilde{\mathbf{h}}_i=\text{CONCAT}(\{\widetilde{\mathbf{h}}^{(l)}_{i}\}_{l=1}^L)$, where $\mathbf{h}^{(l)}_{i}$ is the embedding of node $v_i$ in the node embedding matrix $\mathbf{H}^{(l)}$ obtained from the $l$-th layer of the graph encoder. In addition, a $\operatorname{READOUT}$ function is used to obatain the graph-level representation $\mathbf{h}_g=\operatorname{READOUT}(\{\mathbf{h}_i\}_{i=1}^N)$. Finally, the learning objective is defined as follows
\begin{equation}
\max _{\theta}\sum_{g \in \mathcal{G}}\frac{1}{|g|} \sum_{v_i\in g}\mathcal{MI}\left(\mathbf{h}_{g}, \mathbf{h}_{i}\right)
\end{equation}

\noindent where the negative samples to contrast with $\mathbf{h}_{g}$ is node representations on the augmented graph $Neg(\mathbf{h}_{g})=\{\widetilde{\mathbf{h}}_{i}\}_{v_i\in\mathcal{V}}$.

\textbf{BiGi} \cite{cao2021bipartite} is specifically designed for bipartite graph, where the class label $y_i\in \{0, 1\}$ of each node $v_i$ is already known. For a given $g=(\mathbf{A},\mathbf{X})$, it first applies a structure-based augmentation to obtain $\widetilde{g}=(\widetilde{\mathbf{A}},\mathbf{X})=\mathcal{T}(\mathbf{A},\mathbf{X})$. Then a shared graph encoder $f_\theta(\cdot)$ is applied to obtain $\mathbf{H}=f_{\theta}(\mathbf{A},\mathbf{X})$ and $\widetilde{\mathbf{H}}=f_{\theta}(\widetilde{\mathbf{A}},\mathbf{X})$. Beisde, it can obtain the graph-level representation from $\mathbf{H}$ directly as follows
\begin{equation}
\begin{small}
\begin{aligned}
\mathbf{h}_g=\Big[\sigma\Big(\frac{1}{|\mathcal{V}^{(1)}|}\sum_{v_i\in\mathcal{V}^{(1)}} \mathbf{h}\Big) \Big\| \sigma\Big(\frac{1}{|\mathcal{V}^{(2)}|}\sum_{v_i\in\mathcal{V}^{(2)}} \mathbf{h}_i\Big)\Big]
\end{aligned}
\end{small}
\end{equation}

\noindent where $\mathcal{V}^{(1)}=\{v_i|v_i\in\mathcal{V}, y_i=0\}$ and $\mathcal{V}^{(2)}=\{v_i|v_i\in\mathcal{V}, y_i=1\}$.
For a given edge $(v_i,v_j)\in\mathcal{E}$, it first performs the ege-nets sampling to obtain two subgraph (centered at node $v_i$ and $v_j$), and then gets their node feature matrix $\mathbf{H}^{(i)}$ and $\mathbf{H}^{(j)}$ from $\mathbf{H}$ directly.
Then a subgraph-level attention module (similar to GAT) is applied to obatain two subgraph-level representation $\mathbf{h}_i=Att_\gamma(\mathbf{H}^{(i)})$ and $\mathbf{h}_j=Att_\gamma(\mathbf{H}^{(j)})$. Finally, $\mathbf{h}_i$ and $\mathbf{h}_j$ are fused to obtain $\mathbf{h}_{i,j}$ = $[\mathbf{h}_i\|\mathbf{h}_j]$. Similarity, it can obtain the fused representation $\widetilde{\mathbf{h}}_{i,j}$ from $\widetilde{\mathbf{H}}$. Finally, the learning objective is defined as follows:
\begin{equation}
\max _{\theta, \gamma} \frac{1}{|\mathcal{E}|} \sum_{(v_i,v_j)\in\mathcal{E}}\mathcal{MI}\left(\mathbf{h}_{g}, \mathbf{h}_{i,j}\right)
\end{equation}

\noindent where the negative samples is defined as $Neg(\mathbf{h}_{g})=\widetilde{\mathbf{h}}_{i,j}$.

\subsection{Contrasive Objectives} \label{sec:3.4}
The main way to optimize the contrastive learning is to treat two representations \emph{(views)} $\mathbf{h}_i$ and $\mathbf{h}_j$ as random variables and maximize their mutual information, given by
\begin{equation}
\mathcal{MI}(\mathbf{h}_i, \mathbf{h}_j)=\mathbb{E}_{p(\mathbf{h}_i, \mathbf{h}_j)}\left[\log \frac{p(\mathbf{h}_i, \mathbf{h}_j)}{p(\mathbf{h}_i) p(\mathbf{h}_j)}\right]
\label{equ:kl}
\end{equation}

\noindent To computationally estimate the mutual information in contrastive learning, three lower-bound forms of the mutual information are derived, and then the mutual information is maximized indirectly by maximizing their lower-bounds. 

\textbf{Donsker-Varadhan Estimator} \cite{belghazi2018mutual} is one of the \emph{lower-bound} to the mutual information, defined as
\begin{equation}
\begin{small}
\begin{aligned}
\mathcal{MI}_{DV}\left(\mathbf{h}_{i}, \mathbf{h}_{j}\right)= & \mathbb{E}_{p\left(\mathbf{h}_{i}, \mathbf{h}_{j}\right)}\left[\mathcal{D}\left(\mathbf{h}_{i}, \mathbf{h}_{j}\right)\right] \\ & - \log \mathbb{E}_{p\left(\mathbf{h}_{i}\right) p\left(\mathbf{h}_{j}\right)}\left[e^{\mathcal{D}\left(\mathbf{h}_{i}, \mathbf{h}_{j}\right)}\right]
\end{aligned}
\end{small}
\end{equation}

\noindent where $p(\mathbf{h}_i, \mathbf{h}_j)$ denotes the joint distribution of two representations $\mathbf{h}_i$, $\mathbf{h}_j$, and $p(\mathbf{h}_i)p(\mathbf{h}_j)$ denotes the product of marginals. $\mathcal{D}: \mathbb{R}^{q} \times \mathbb{R}^{q} \rightarrow \mathbb{R}$ is a discriminator that maps two views $\mathbf{h}_i$, $\mathbf{h}_j$ to an agreement score. Generally, the discriminator $\mathcal{D}$ can optionally apply an additional prediction head $g_\omega(\cdot)$ to map $\mathbf{h}_i$ to $\mathbf{z}_i=g_\omega(\mathbf{h}_i)$ before computing agreement scores, where $g_\omega(\cdot)$ can be a linear mapping, a nonlinear mapping (e.g., MLP), or even a non-parametric identical mapping ($\mathbf{z}_i$ = $\mathbf{h}_i$). The discriminator $\mathcal{D}$ can be taken in various forms, i.e., the standard inner product $\mathcal{D}(\mathbf{z}_i,\mathbf{z}_j)=\mathbf{z}_i^T\mathbf{z}_j$, the inner product $\mathcal{D}(\mathbf{z}_i,\mathbf{z}_j)=\mathbf{z}_i^T\mathbf{z}_j/\tau$ with temperature parameter $\tau$, the cosine similarity $\mathcal{D}(\mathbf{z}_i,\mathbf{z}_j)=\frac{\mathbf{z}_i^T\mathbf{z}_j}{||\mathbf{z}_i|| ||\mathbf{z}_j||}$, or the gaussian similarity $\mathcal{D}(\mathbf{z}_i,\mathbf{z}_j)=\exp\left(-\frac{||\mathbf{z}_i-\mathbf{z}_j||_2^2}{2\sigma^2}\right)$.

\textbf{Jensen-Shannon Estimator.} Replacing the KL-divergence in Equ.~\ref{equ:kl} with the JS-divergence, it derives another Jensen-Shannon (JS) estimator \cite{nowozin2016f} which can estimate and optimize the mutual information more efficiently. The Jensen-Shannon (JS) estimator is defined as
\begin{equation}
\begin{small}
\begin{aligned}
\mathcal{MI}_{JS} & \left(\mathbf{h}_{i},\mathbf{h}_{j}\right)=\mathbb{E}_{p\left(\mathbf{h}_{i}, \mathbf{h}_{j}\right)}\bigg[\log\left(\mathcal{D}(\mathbf{h}_{i}, \mathbf{h}_{j}\right))\bigg] \\ & - \log \mathbb{E}_{p\left(\mathbf{h}_{i}\right) p\left(\mathbf{h}_{j}\right)}\bigg[log\Big(1-\mathcal{D}\left(\mathbf{h}_{i}, \mathbf{h}_{j}\right)\Big)\bigg]
\end{aligned}
\end{small}
\label{equ:JS}
\end{equation}

Let $\mathcal{D}\left(\mathbf{h}_{i}, \mathbf{h}_{j}\right)=\text{sigmod}(\mathcal{D}'\left(\mathbf{h}_{i}, \mathbf{h}_{j})\right)$, the Equ.$\ref{equ:JS}$ can be re-writed as a softplus (SP) version \cite{hassani2020contrastive,sun2019infograph}, as follows
\begin{equation}
\begin{small}
\begin{aligned}
\mathcal{MI}_{SP} & \left(\mathbf{h}_{i},\mathbf{h}_{j}\right)=\mathbb{E}_{p\left(\mathbf{h}_{i}, \mathbf{h}_{j}\right)}\bigg[-sp\Big(-\mathcal{D}'\left(\mathbf{h}_{i}, \mathbf{h}_{j}\right)\Big)\bigg] \\ & - \log \mathbb{E}_{p\left(\mathbf{h}_{i}\right) p\left(\mathbf{h}_{j}\right)}\bigg[sp\Big(\mathcal{D}'\left(\mathbf{h}_{i}, \mathbf{h}_{j}\right)\Big)\bigg]
\end{aligned}
\end{small}
\end{equation}

\noindent where $sp(x)=\log\left(1+e^x\right)$.

\textbf{InfoNCE Estimator.} InfoNCE \cite{gutmann2010noise} is one of the most popular lower-bound to the mutual information, defined as
\begin{equation}
\begin{small}
\begin{aligned}
\mathcal{MI}_{NCE} & \left(\mathbf{h}_{i},\mathbf{h}_{j}\right)=\mathbb{E}_{p\left(\mathbf{h}_{i}, \mathbf{h}_{j}\right)}\Bigg[\mathcal{D}\left(\mathbf{h}_{i}, \mathbf{h}_{j}\right) \\ & - \mathbb{E}_{K\sim\mathcal{P}^N}\Big[\log\frac{1}{N}\sum_{\mathbf{h}_{j}' \in K} e^{\mathcal{D}\left(\mathbf{h}_{i}, \mathbf{h}_{j}'\right)}\Big]\Bigg]
\end{aligned}
\end{small}
\label{equ:nce}
\end{equation}

\noindent where $K$ consists of $N$ random variables sampled from a n identical and independent distribution. NT-Xent loss \cite{sohn2016improved} is a special version of the InfoNCE loss, which defines the discriminator $\mathcal{D}$ as $\mathcal{D}(\mathbf{h}_i,\mathbf{h}_j)=\mathbf{h}_i^T\mathbf{h}_j/\tau$ with temperature parameter $\tau$. Taking graph classification as an example, $f_\gamma(\cdot)$ is a graph encoder that maps a graph $g=(\mathbf{A},\mathbf{X})\in\mathcal{G}$ to a graph-level representation $\mathbf{h}_g=f_\gamma(\mathbf{A},\mathbf{X})$. The InfoNCE is in practice computed on a mini-batch $\boldsymbol{B}$ of size $N+1$, then the Equ.~\ref{equ:nce} can be re-writed (with $\log N$ discarded) as 
\begin{equation}
\begin{small}
\begin{aligned}
\mathcal{MI}_{NCE} = -\frac{1}{N+1}\sum_{(\mathbf{A},\mathbf{X}) \in \boldsymbol{B}} \log \frac{e^{\mathcal{D}\left(\mathbf{h}_{i}, \mathbf{h}_{j}\right)}}{\sum_{(\mathbf{A}',\mathbf{X}') \in \boldsymbol{B}} e^{\mathcal{D}\left(\mathbf{h}_{i}, \mathbf{h}_{j}^{\prime}\right)}}
\end{aligned}
\end{small}
\end{equation}

\noindent where $\mathbf{h}_{i}$, $\mathbf{h}_{j}$ are the positive pair of views that comes from the same graph $g=(\mathbf{A},\mathbf{X})$, and $\mathbf{h}_{i}$ and $\mathbf{h}_{j}'$ are the negative pair of views that are computed from $g=(\mathbf{A},\mathbf{X})$ and $g'=(\mathbf{A}',\mathbf{X}')$ identically and independently.

\textcolor{mark}{\textbf{Triplet Margin Loss.} While all three MI estimators above estimate the lower bound on mutual information, mutual information maximization has been shown not to be a must for contrastive learning \cite{tschannen2019mutual}. For example, the triplet margin loss \cite{schroff2015facenet} can also be used to optimize the contrastive learning, but it is not an MI-based contrastive objective, and optimizing it does not guarantee the maximization of mutual information. The triplet margin loss is defined as}
\begin{equation}
\begin{small}
\begin{aligned}
\mathcal{L}_{triplet}  & \left(\mathbf{h}_{i},\mathbf{h}_{j}\right) = \mathbb{E}_{\left[(\mathbf{A},\mathbf{X}),(\mathbf{A}',\mathbf{X}')\right]\sim\mathcal{G}\times\mathcal{G}} \Big[ \\ & \max\{\mathcal{D}\left(\mathbf{h}_{i}, \mathbf{h}_{j}'\right)-\mathcal{D}\left(\mathbf{h}_{i}, \mathbf{h}_{j}\right)+\epsilon, 0\}\Big]
\end{aligned}
\end{small}
\end{equation}

\noindent \textcolor{mark}{where the triplet margin loss does not directly minimize the agreement of the negative sample pair $\mathcal{D}(\mathbf{h}_{i}, \mathbf{h}_{j}')$, but only ensures that the agreement of the negative sample pair is smaller than that of the positive sample pair by a margin value $\epsilon$. The idea behind is that when the negative samples are sufficiently far apart, i.e., the agreement between them is small enough, there is no need to further reduce their agreement, which helps to focus the training more on those hard samples that are hard to distinguish. The quadruplet loss \cite{chen2017beyond} further considers imposing constraints on inter-class samples on top of the triplet margin loss, defined as:}
\begin{equation}
\begin{small}
\begin{aligned}
\mathcal{L}_{Quadruplet}  & \left(\mathbf{h}_{i},\mathbf{h}_{j}\right) = \mathcal{MI}_{triplet} + \mathbb{E}_{\left[(\mathbf{A},\mathbf{X}),(\mathbf{A}',\mathbf{X}')\right]\sim\mathcal{G}\times\mathcal{G}} \Big[ \\ & \max\{\mathcal{D}\left(\mathbf{h}_{i}', \mathbf{h}_{j}'\right)-\mathcal{D}\left(\mathbf{h}_{i}, \mathbf{h}_{j}\right)+\epsilon', 0\}\Big]
\end{aligned}
\end{small}
\end{equation}

\noindent \textcolor{mark}{where $\epsilon'$ is a smaller margin value than $\epsilon$. The quadruplet loss differs from the triplet margin loss in that it not only uses an anchor-based sampling strategy but also samples negative samples in a more random way, which helps to learn more distinguishable inter-class boundaries.}

\textcolor{mark}{\textbf{RankMI Loss.} While both triplet margin loss and quadruplet loss ignore the lower bound of the mutual information, the RankMI Loss \cite{kemertas2020rankmi} seamlessly incorporates information-theoretic approaches into the representation learning and maximizes the mutual information among samples belonging to the same category, defined as:}
\begin{equation}
\begin{aligned}
\mathcal{MI}_{RankMI}  & \left(\mathbf{h}_{i},\mathbf{h}_{j}\right) = \mathbb{E}_{\left[(\mathbf{A},\mathbf{X}),(\mathbf{A}',\mathbf{X}')\right]\sim\mathcal{G}\times\mathcal{G}} \Big[ \\ & 
\mathcal{D}(\mathbf{h}_{i}, \mathbf{h}_{j}) + \log(2 - e^{\mathcal{D}(\mathbf{h}_{i}', \mathbf{h}_{j}')})
\Big]
\end{aligned}
\end{equation}

\noindent \textcolor{mark}{As RankMI can incorporate margins based on random positive and negative pairs, the quadruple loss can be considered as a special case of RankMI with a fixed margin.}
\section{Generative Learning}
Compared with contrastive methods, the generative methods shown in Fig.~\ref{fig:1}(b) are based on generative models and treat rich information embedded in the data as a natural self-supervision. In generative methods, the prediction head $g_\omega(\cdot)$ is usually called the graph decoder, which is used to perform graph reconstruction. Categorized by how the reconstruction is performed, we summarize generative methods into two categories: (1) graph autoencoding that performs reconstruction in a once-for-all manner; (2) graph autoregressive that iteratively performs reconstruction. In this section, due to space limitations, we present only some representative generative methods and place those {relatively less important} works in \textbf{Appendix B}.

\subsection{Graph Autoencoding}
Since the autoencoder \cite{hinton2006reducing} was proposed, it has been widely used as a basic architecture for a variety of image and text data. Given restricted access to the graph, the graph autoencoder is trained to reconstruct certain parts of the input data. Depending on which parts of the input graph are given or restricted, various pretext tasks have been proposed, which will be reviewed one by one next.

\textbf{Graph Completion} \cite{you2020does}. Motivated by the success of image inpainting, graph completion is proposed as a pretext task for graph data. It first masks one node by removing part of its features, and then aims to reconstruc masked features by feeding \emph{unmasked} node features in the neighborhood. For a given node $v_i$, it randomly masks its features  $\mathbf{x}_i$ with $\widehat{\mathbf{x}}_i=\mathbf{x}_i\odot\mathbf{m}_i$ to obtain a new node feature matrix $\widehat{\mathbf{X}}$, and then aim to reconstruct masked features. More formally,
\begin{equation}
\mathcal{L}_{ssl}\left(\theta, \mathbf{A}, \widehat{\mathbf{X}} \right)=\left\|f_{\theta}(\mathbf{A},\widehat{\mathbf{X}})_{v_{i}} - \mathbf{x}_i\right\|^2
\end{equation}

\noindent Here, it just takes one node as an example, and the reconstruction of multiple nodes can be considered in practice. Note that only those \emph{unmasked} neighborhood nodes can be used to reconstruct the target node for graph completion. 

\textbf{Node Attribute Masking} \cite{jin2020self} is similar to Graph Completion, but it reconstructs the features of multiple nodes \emph{simultaneously}, and it no longer requires that the neighboring node features used for reconstruction must be unmasked.

\textbf{Edge Attribute Masking} \cite{hu2019strategies}. This pretext task is specifically designed for graph data with known edge features, and it enables GNN to learn more edge relation information. Similarly, it first randomly masks the features of a edge set $\mathcal{M}_e$. Specifically, it obtains a masked edge feature matrix $\widehat{\mathbf{X}}^e$ where $\widehat{\mathbf{x}}^e_{i,j}=\mathbf{x}^e_{i,j}\odot\mathbf{m}_{i,j}$ for $(v_i, v_j)\in\mathcal{M}_e$. More formally,
\begin{equation}
\mathcal{L}_{ssl}\left(\theta, \mathbf{A}, \mathbf{X}, \widehat{\mathbf{X}}^e \right)=\frac{1}{|\mathcal{M}_e|}\sum_{(v_i,v_j)\in\mathcal{M}_e}\left\| \overline{\mathbf{x}}^e_{i,j} - \mathbf{x}^e_{i,j}\right\|^2
\end{equation}

\noindent where $\overline{\mathbf{x}}^e_{i,j}=(\overline{\mathbf{X}}^e)_{i,j}$ and $\overline{\mathbf{X}}^e=f_{\theta}(\mathbf{A},\mathbf{X}, \widehat{\mathbf{X}}^e)$.

\textbf{Node Attribute Denoising} \cite{manessi2020graph}. Different from Node Attribute Masking, this pretext task aims to add noise to the node features to obtain a noisy node feature matrix $\widehat{\mathbf{X}}=\mathbf{X}+N(\mathbf{0}, \mathbf{\Sigma})$, and then ask the model to reconstruct the clean node features $\mathbf{X}$. More formally,
\begin{equation}
\mathcal{L}_{ssl}\left(\theta, \mathbf{A}, \widehat{\mathbf{X}} \right)=\frac{1}{N}\sum_{v_i\in\mathcal{V}}\left\|f_{\theta}(\mathbf{A},\widehat{\mathbf{X}})_{v_{i}} - \mathbf{x}_i\right\|^2
\end{equation}

\noindent where adding noise is only one means of corrupting the image, in addition to blurring, graying, etc. Inspired by this, it can use \emph{arbitrary} corruption operations $\mathcal{C}(\cdot)$ to obtain the corrupted features and then force the model to reconstruct them. Different from Node Attribute Denoising, which reconstructs raw features from noisy inputs, Node Embedding Denoising aims to reconstructs clean node features $\mathbf{X}$ from noisy embeddings $\widehat{\mathbf{H}}=\mathbf{H}+N(\mathbf{0}, \mathbf{\Sigma})$.

\textbf{Adjacency Matrix Reconstruction} \cite{zhu2020self}. The graph adjacency matrix is one of the most important information in graph data, which stores the graph structure information and the relations between nodes. This pretext task randomly perturbs parts of the edges in a graph $\mathbf{A}$ to obtain $\widehat{\mathbf{A}}$, then requires the model to reconstruct the adjacency matrix of the input graph. More formally,
\begin{equation}
\mathcal{L}_{ssl}\left(\theta, \widehat{\mathbf{A}}, \mathbf{X}\right)=\frac{1}{N^2}\sum_{i,j}\left(\overline{\mathbf{A}}_{i,j} - \mathbf{A}_{i,j}\right)^2
\end{equation}

\noindent where $\overline{\mathbf{A}}=f_\theta(\widehat{\mathbf{A}},\mathbf{X})$. During the training process, since the adjacency matrix $\mathbf{A}$ is usually a sparse matrix, it can also use cross-entropy instead of MAE as loss in practice.

\begin{figure*}[ht]
	\begin{center}
		\subfigure[Node Property Prediction]{\includegraphics[width=0.48\linewidth]{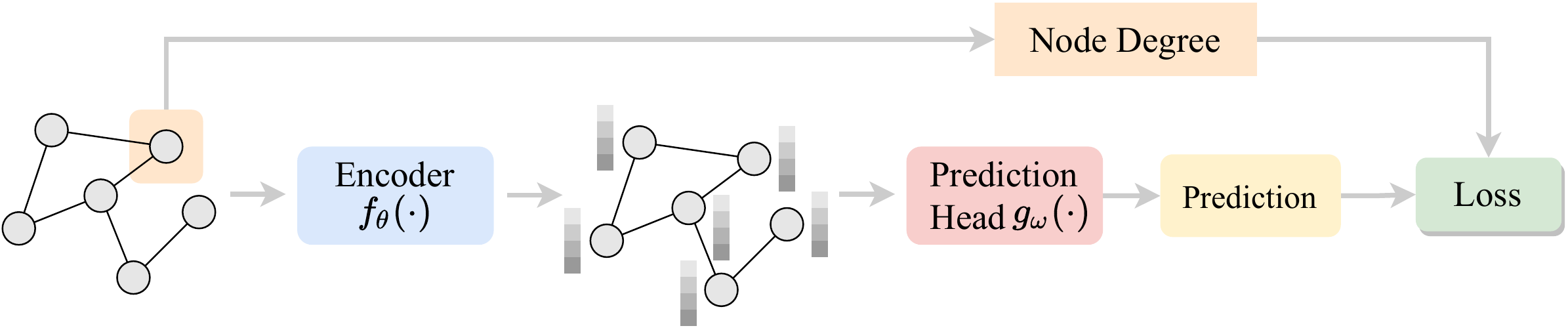}}
		\subfigure[Context-based Prediction]{\includegraphics[width=0.48\linewidth]{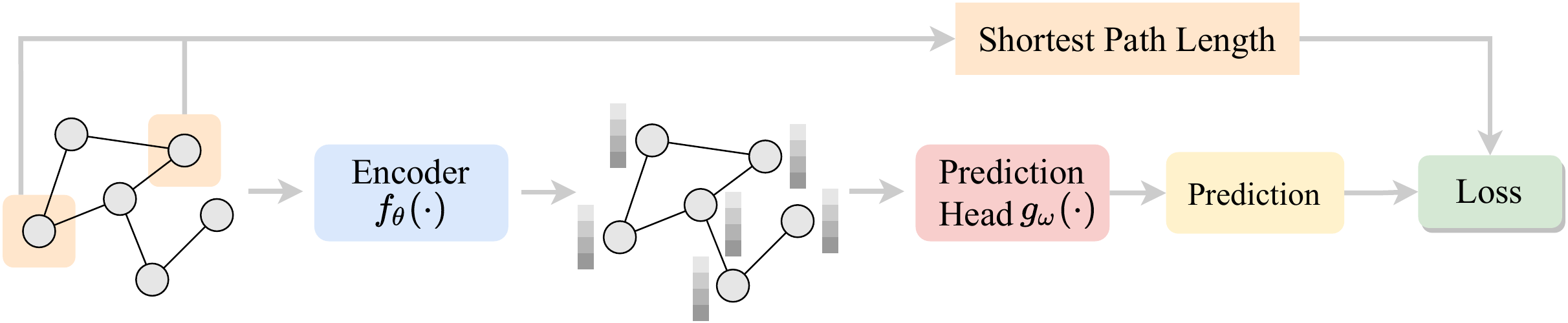}}
		\subfigure[Self-training]{\includegraphics[width=0.52\linewidth]{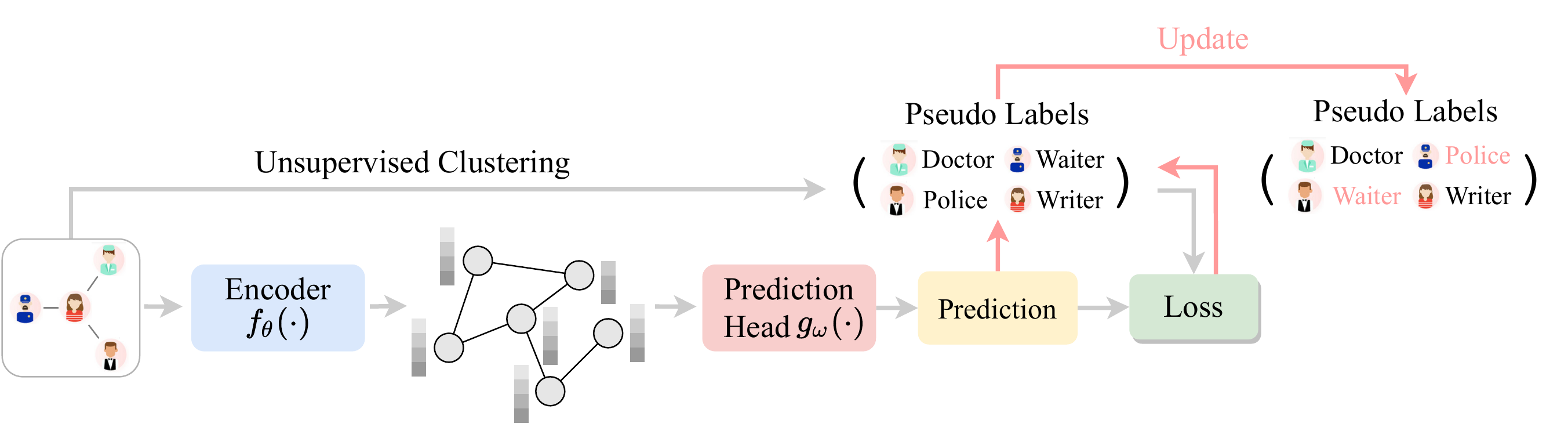}}
		\subfigure[Domain Knowledge-based]{\includegraphics[width=0.45\linewidth]{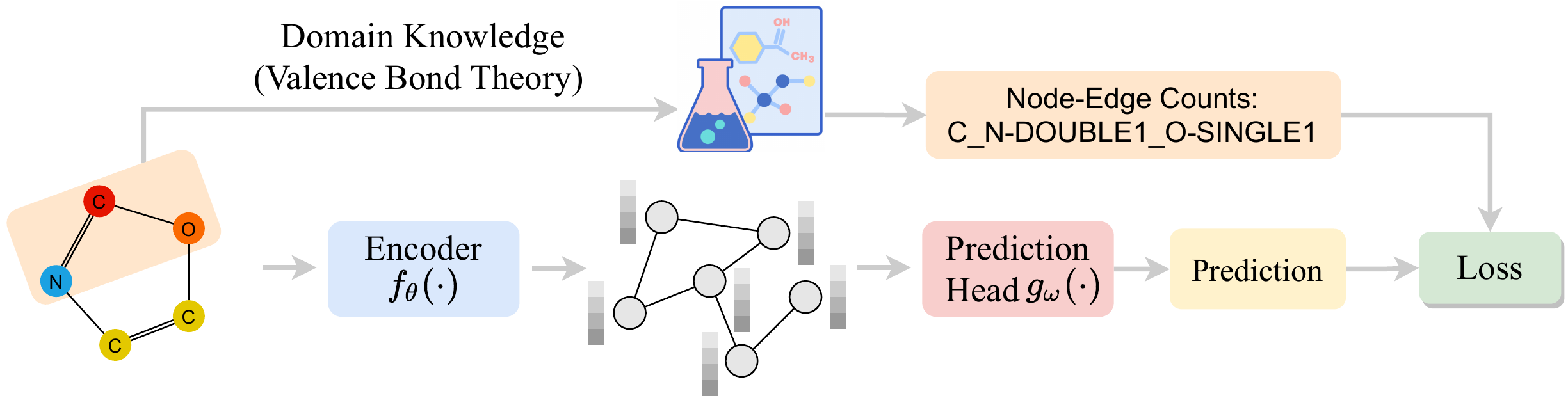}}
	\end{center}
	\vspace{-1em}
	\caption{A comparison of predictive learning methods. Categorized by how the labels are obtained, we summarize predictive methods for graph data into four categories: node property prediction, context-based prediction, self-training, and domain knowledge-based prediction. \textbf{Fig.~(a)}: the node property prediction pre-calculates the node properties, such as node degree, and used them as self-supervised labels. \textbf{Fig.~(b)}: for the context-based prediction, the local or global contextual information in the graph, such as the shortest path length between nodes, can be extracted as labels to help with self-supervised learning. \textbf{Fig.~(c)}: The self-learning method applies algorithms such as unsupervised clustering to obtain pseudo-labels and then updates the pseudo-label set of the previous stage based on the prediction results or losses. \textbf{Fig.~(d)}: for the domain knowledge-based prediction, the domain knowledge, such as expert knowledge or specialized tools, can be used in advance to obtain informative labels.}
	\vspace{-1em}
	\label{fig:5}
\end{figure*}

\subsection{Graph Autoregressive}
The autoregressive model is a linear regression model that uses a combination of random variables from previous moments to represent random variables at a later moment.

\textbf{GPT-GNN} \cite{hu2020gpt}. In recent years, the idea of GPT \cite{radford2019language} has also been introduced into the GNN domain. For example, GPT-GNN proposes an autoregressive framework to perform node and edge reconstruction on given graph \emph{iteratively}. Given a graph $g_t=(\mathbf{A}_t,\mathbf{X}_t)$ with its nodes and edges randomly masked in iteration $t$, GPT-GNN generates one masked node $X_i$ and its connected edges $E_i$ to obtain a updated graph $g_{t+1}=(\mathbf{A}_{t+1},\mathbf{X}_{t+1})$ and optimizes the likelihood of the node and edges generation in the next iteration $t+1$, with the learning objective defined as
\begin{equation}
\begin{aligned}
& p_{\theta}\left(\mathbf{X}_{t+1}, \mathbf{A}_{t+1} \mid \mathbf{X}_{t}, \mathbf{A}_{t}\right) \\
=& \sum_{o} p_{\theta}\left(X_{i}, E_{i}^{ \neg o} \mid E_{i}^o, \mathbf{X}_{t}, \mathbf{A}_{t}\right) \cdot p_{\theta}\left(E_{i}^o \mid \mathbf{X}_{t}, \mathbf{A}_{t}\right) \\
=& \mathbb{E}_{o}\big[p_{\theta}\left(X_{i}, E_{i}^{ \neg o} \mid E_{i}^o, \mathbf{X}_{t}, \mathbf{A}_{t}\right)\big] \\
=& \mathbb{E}_{o}\left[p_{\theta}\big(\mathbf{X}_{t+1} \mid E_{i}^o, \mathbf{X}_{t}, \mathbf{A}_{t}\right)p_{\theta}(E_{i}^{ \neg o} \mid E_{i}^o, \mathbf{X}_{t+1}, \mathbf{A}_{t})\big]
\end{aligned}
\end{equation}

\noindent where $o$ is a variable to denote the index vector of all edges within $E_t$ in the iteration $t$. Thus, $E_{t}^o$ denotes the observed edges in the iteration $t$, and $E_{i}^{ \neg o}$ denotes the the masked edges (to be generated) in the iteration $t+1$. Finally, the graph generation process is factorized into a node \emph{attribute generation} step $p_{\theta}\left(\mathbf{X}_{t+1} \mid E_{i}^o, \mathbf{X}_{t}, \mathbf{A}_{t}\right)$ and an \emph{edge generation} step $p_{\theta}(E_{i}^{ \neg o} \mid E_{i}^o, \mathbf{X}_{t+1}, \mathbf{A}_{t})$. In practice, GPT-GNN performs node and edge generation iteratively.

\section{Predictive Learning}
The contrastive methods deal with the \emph{inter-data} information (data-data pairs), the generative methods focus on the \emph{intra-data} information, while the predictive methods aim to \emph{self-generate informative labels} from the data as supervision and handle the \emph{data-label} relationships. Categorized by how labels are obtained, we summarize predictive methods into four categories: (1) Node Property Prediction. The properties of nodes, such as node degree, are pre-calculated and used as self-supervised labels to perform prediction tasks. (2) Context-based Prediction. Local or global contextual information in the graph can be extracted as labels to aid self-supervised learning, e.g., by predicting the shortest path length between nodes, the model can capture long-distance dependencies, which is beneficial for downstream tasks such as link prediction. (3) Self-Training. Learning with the pseudo-labels obtained from the prediction or clustering in a previous stage or even randomly assigned. (4) Domain Knowledge-based Prediction. Expert knowledge or specialized tools are used in advance to analyze graph data (e.g., biological or chemical data) to obtain informative labels. A comparison of four predictive methods is shown in Fig.~\ref{fig:5}. In this section, due to space limitations, we present only some representative predictive methods and place those {relatively less important} works in \textbf{Appendix C}.

\subsection{Node-Property Prediction (NP)}
An effective way to perform predictive learning is to take advantage of the extensive implicit numerical properties within the graph, e.g. node properties, such as node degree and local clustering coefficient. Specifically, it first defines a mapping $\Omega: \mathcal{V} \rightarrow \mathcal{Y}$ to denote the extraction of \emph{statistical labels} $y_i=\Omega(\mathbf{A}, \mathbf{X})_{v_i}$ for each node $v_i$ from graph $g=(\mathbf{A}, \mathbf{X})$. The learning objective is then formulated as
\begin{equation}
\mathcal{L}_{ssl}\left(\theta, \mathbf{A}, \mathbf{X}\right)=\frac{1}{N}\sum_{v_i\in\mathcal{V}}\Big( f_{\theta}(\mathbf{A},\mathbf{X})_{v_i} - y_{i}\Big)^2
\end{equation}

\noindent where $f_{\theta}(\mathbf{A},\mathbf{X})_{v_i}$ is the predicted label of node $v_i$. With different node properties, the mapping function $\Omega(\cdot)$ can have different designs. If it use node degree as a local node property for self-supervision, we have $y_i=\Omega(\mathbf{A}, \mathbf{X})_{v_i}=\sum_{j=1}^{N} \mathbf{A}_{i,j}$. For the local clustering coefficients, we have
\begin{equation}
y_i=\Omega(\mathbf{A}, \mathbf{X})_{v_i}=\frac{2\big|\{(v_m,v_n)|v_m\in\mathcal{N}_i, v_n\in\mathcal{N}_i\}\big|}{|\mathcal{N}_i|\dot(|\mathcal{N}_i|-1)}
\end{equation}

\noindent where the local clustering coefficient is a local coefficient describing the level of node aggregation in a graph. Beyond the above two properties, any other node property (or even a combination of them) can be used as statistical labels to perform the pretext task of Node Property Prediction.

\subsection{Context-based Prediction (CP)}
Apart from Node Property Prediction, the underlying graph structure information can be further explored to construct a variety of regression-based or classification-based pretext tasks and thus provide self-supervised signals. We refer to this branch of methods as context-based predictive methods because it generally explores contextual information.

\textbf{$\text{S}^2\text{GRL}$} \cite{peng2020self}. Motivated by the observation that two arbitrary nodes in a graph can interact with each other through paths of different lengths, $\text{S}^2\text{GRL}$ treats the contextual position of one node relative to the other as a source of free and effective supervisory signals. Specifically, it defines the $k$-hop context of node $v_i$ as $\mathcal{C}_i^k=\{v_j|d(v_i,v_j)=k\}(k=1,2,\cdots,K)$, where $d(v_i,v_j)$ is the shortest path length between node $v_i$ and node $v_j$. In this way, for each target node $v_i$, if a node $v_j \in \mathcal{C}_i^k$, then the hop count $k$ (relative contextual position) will be assigned to node $v_j$ as pseudo-label $y_{i,j}=k$. The learning objective is defined as predicting the hop count between pairs of nodes, as follows
\begin{equation}
\begin{small}
\begin{aligned}
\mathcal{L}_{ssl}\left(\theta, \omega, \mathbf{A}, \mathbf{X}\right)&=\frac{1}{NK}\sum_{v_i\in\mathcal{V}} \sum_{k=1}^K\sum_{v_j\in\mathcal{C}_i^k} \ell\Big(f_w\big(\\ & f_{\theta}(\mathbf{A},\mathbf{X})_{v_{i}},f_{\theta}(\mathbf{A},\mathbf{X})_{v_{j}}\big), k\Big) 
\end{aligned}
\end{small}
\end{equation}

\noindent where $\ell(\cdot)$ denotes the cross entropy loss and $f_\omega(\cdot)$ linearly maps the input to a 1-dimension value. Compared with the task of $\text{S}^2\text{GRL}$, the \textbf{PairwiseDistance} \cite{jin2020self} has truncated the shortest path longer than 4, mainly to avoid the excessive computational burden and to prevent very noisy ultra-long pairwise distances from dominating the optimization.

\textbf{PairwiseAttrSim} \cite{jin2020self}. Due to the neighborhood-based message passing mechanism, the learned representations of two similar nodes in the graph are not necessarily similar, as opposed to two identical images that will yield the same representations in the image domain. Though we would like to utilize local neighborhoods in GNNs to enhance node feature transformation, we still wishes to preserve the node pairwise similarity to some extent, rather than allowing a node's neighborhood to drastically change it. Thus, the pretext task of PairwiseAttrSim can be established to achieve node similarity preservation. Specifically, it first samples node pairs with the $K$ highest and lowest similarities $\mathcal{S}_{i,h}$ and $\mathcal{S}_{i,s}$ for node $v_i$, given by
\begin{equation}
\begin{small}
\begin{aligned}
\mathcal{S}_{i,h}&=\left\{\left(v_{i}, v_{j}\right) \mid s_{i j} \text { in top-K of }\left\{s_{i k}\right\}_{k=1,k\neq i}^{N}\right\} \\
\mathcal{S}_{i,l}&=\left\{\left(v_{i}, v_{j}\right) \mid s_{i j} \text { in bottom-K of }\left\{s_{i k}\right\}_{k=1,k \neq i}^{N}\right\}
\end{aligned}
\end{small}
\end{equation}

\noindent where $s_{i,j}$ measures the node feature similarity between node $v_i$ and node $v_j$ (according to cosine similarity). Let $\mathcal{S}_{i}=\mathcal{S}_{i,h}\cup\mathcal{S}_{i,h}$, the learning objective can then be formulated as a regression problem, as follows
\begin{equation}
\begin{small}
\begin{aligned}
\mathcal{L}_{ssl}\left(\theta, \omega, \mathbf{A}, \mathbf{X}\right)&=\frac{1}{2NK}\sum_{v_i\in\mathcal{V}}\sum_{(v_{m}, v_{n}) \in \mathcal{S}_{i}} \Big(f_w\big(\\&f_{\theta}(\mathbf{A},\mathbf{X})_{v_{m}},f_{\theta}(\mathbf{A},\mathbf{X})_{v_{n}}\big) - s_{m,n}\Big)^2
\end{aligned}
\end{small}
\end{equation}

\noindent where $f_\omega(\cdot)$ linearly maps the input to a 1-dimension value.

\textbf{Distance2Clusters} \cite{jin2020self}. The PairwiseAttrSim applies a sampling strategy to reduce the time complexity, but still involves sorting the node similarities, which is a very time-consuming operation. Inspired by various unsupervised clustering algorithms \cite{Macqueen67somemethods,gao2020clustering,wu2020deep,li2020consistent,yang2019deep,yang2016joint,xie2016unsupervised,mcconville2019n2d,caron2018deep}, if a set of clusters can be pre-obtained, the PairwiseAttrSim can be further simplified to predict the shortest path from each node to the anchor nodes associated with cluster centers, resulting in a novel pretext task - Distance2Clusters. Specifically, it first partitions the graph into $K$ clusters $\{C_1,C_2,\cdots,C_K\}$ by applying some classical unsupervised clustering algorithms. Inside each cluster $C_k$ , the node with the highest degree will be taken as the corresponding cluster center, denoted as $c_k$ ($1\leq k \leq K$). Then it can calculate the distance $\mathbf{d}_i \in \mathbb{R}^K$ from node $v_i$ to cluster centers $\{c_k\}_{k=1}^K$. The learning objective of Distance2Clusters is defined as
\begin{equation}
\begin{aligned}
\mathcal{L}_{ssl}\left(\theta, \mathbf{A}, \mathbf{X}\right)  = \frac{1}{N}\sum_{v_{i} \in \mathcal{V}} \Big\|f_{\theta}(\mathbf{A},\mathbf{X})_{v_{i}}-\mathbf{d}_i\Big\|^2
\end{aligned}
\end{equation}

\textbf{Meta-path Prediction} \cite{hwang2020self}. A meta-path of length $l$ is a sequence of nodes connected with heterogeneous edges, i.e., $v_{1} \stackrel{t_{1}}{\longrightarrow} v_{2} \stackrel{t_{2}}{\longrightarrow} \ldots \stackrel{t_{l}}{\longrightarrow} v_{l}$, where $t_l\in\mathcal{T}^e$ denote the type of $l$-th edge in the meta-path. Given a set of node pair $\mathcal{S}$ sampled from the heterogeneous graph and $K$ \emph{pre-defined} meta-path types $\mathcal{M}$, this pretext task aims to predict if the two nodes $(v_i,v_j)\in\mathcal{S}$ are connected by one of the meta-path type $m\in\mathcal{M}$. Finally, the predictions of the $K$ meta-paths are formulated as $K$ binary classification tasks, as follows
\begin{equation}
\begin{small}
\begin{aligned}
\mathcal{L}_{ssl}\left(\theta, \mathbf{A}, \mathbf{X}\right)&=\frac{1}{K|\mathcal{S}|}\sum_{m\in\mathcal{M}}\sum_{(v_i,v_j)\in\mathcal{S}} \\ & \ell\Big(f_w\big(f_{\theta}(\mathbf{A},\mathbf{X})_{v_{i}},f_{\theta}(\mathbf{A},\mathbf{X})_{v_{j}}\big), \mathbf{Y}_{i,j}^m\Big) 
\end{aligned}
\end{small}
\end{equation}

\noindent where $\ell(\cdot)$ deontes the cross entropy loss, and $\mathbf{Y}_{i,j}^m$ is the ground-truth label where $\mathbf{Y}_{i,j}^m=1$ if there exits a meta-path $m$ between node $v_i$ and node $v_j$, otherwise $\mathbf{Y}_{i,j}^m=0$. 

\textbf{SLiCE} \cite{wang2020self}. Different from the pretext task of Meta-path Prediction \cite{hwang2020self} that requires pre-defined mate-paths, SLiCE automatically learns the composition of different meta-paths for a specific task. Specifically, it first samples a set of nodes $\mathcal{S}$ from the node set $\mathcal{V}$. Given a node in $v_i\in\mathcal{S}$, it generates a context subgraph $g_i=(\mathbf{A}_i,\mathbf{X}_i)$ around $v_i$ and encodes the context as a low-dimensional embedding matrix $\mathbf{H}_i$. Then it randomly masks a node $v_{i}^m$ in graph $g_i$ for prediction. Therefore, the pretext task aims to maximize the probability of observing this masked node $v_{i}^m$ based on the context $g_i$,
\begin{equation}
\begin{split}
\mathcal{L}_{ssl}\left(\theta, \mathbf{A}, \mathbf{X}\right)=\prod_{v_i \in \mathcal{S}} \prod_{v_{i}^m \in g_{i}} p\left(v_{i}^m \mid \mathbf{H}_{i}, \theta\right)
\end{split}
\end{equation}

\noindent where $p(\cdot\mid\theta)$ can in practice be approximated by a graph neural networks model $f_\theta(\cdot)$ parameterized by $\theta$.

\textbf{Distance2Labeled} \cite{jin2020self}. Recent work provides deep insight into existing self-supervised pretext tasks that utilize only attribute and structure information and finds that they are not always beneficial in improving the performance of downstream tasks, possibly because the information mined by the pretext tasks may have been fully exploited during the message passing by the GNN model. Thus, given partial information about downstream tasks, such as a small set of labeled nodes, we can explore label-specific self-supervised tasks. For example, we directly modify the pretext task of Distance2Cluster by combining label information to create a new pretext task - Distance2Labeled. Specifically, it first calculates the average, minimum, and maximum shortest path length from node $v_i$ to all labeled nodes in class $\{C_k\}_{k=1}^K$, resulting in a distance vector $\mathbf{d}_i\in\mathbb{R}^{3K}$. Finally, the learning objective of Distance2Labeled can be formulated as a distance regression problem, as follows
\begin{equation}
\begin{aligned}
\mathcal{L}_{ssl}\left(\theta, \mathbf{A}, \mathbf{X}\right)  = \frac{1}{N}\sum_{v_{i} \in \mathcal{V}} \Big\|f_{\theta}(\mathbf{A},\mathbf{X})_{v_{i}}-\mathbf{d}_i\Big\|^2
\end{aligned}
\end{equation}

Compared with Distance2Cluster, Distance2Labeled utilizes task-specific label information rather than additional unsupervised clustering algorithms to find cluster centers, showing advantages in both efficiency and performance.

\begin{figure*}[!htbp]
	\begin{center}
	\includegraphics[width=1.0\linewidth]{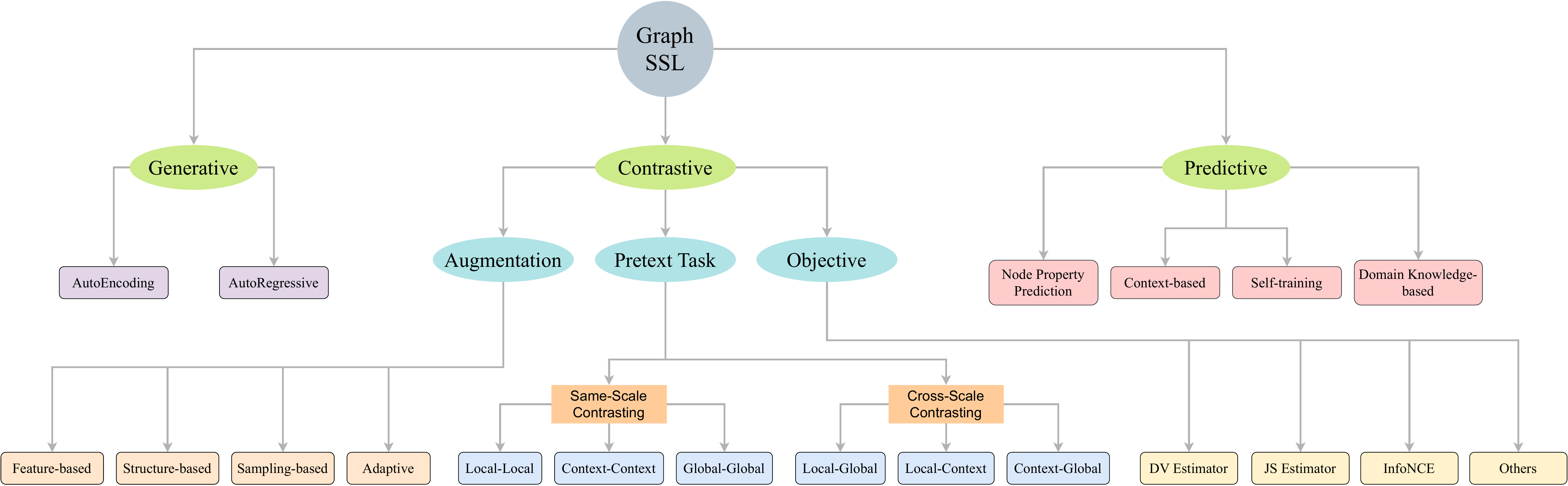}
	\end{center}
	\vspace{-1em}
	\caption{An summary of graph self-supervised learning (SSL) methods. We categorize them into three branches: contrastive, generative, and predictive. For contrastive methods, they contrast different views and deal with data-data pairs (inter-data) information, and we further categorize it from three aspects: augmentation strategy, pretext task, and objective. In terms of augmentation strategy, it can be divided into four major categories: feature-based augmentation, structure-based augmentation, sampling-based augmentation, and adaptive augmentation. From the perspective of pretext tasks, it can be divided into same-scale contrasting and cross-scale contrasting. The same-scale contrasting includes Local-Local (L-L), Context-Cotext (C-C), and Global-Global (G-G) methods, while the cross-scale contrasting includes the Local-Global (L-G), Local-Cotext (L-C), and Context-Global (C-G) methods. For generative methods, they focus on the intra-data information and can be divided into Autoencoding and Autoregressive methods. For predictive methods, it handles the data-label relationship, which can be further divided into four major categories: Node Property Prediction (NP), Content-based Prediction (CP), Self-Training (ST), and Domain Knowledge-based Prediction (DK).}
	\vspace{-1em}
	\label{fig:6}
\end{figure*}

\subsection{Self-Training (ST)}
For self-training methods, the prediction results from the previous stage can be used as labels to guide the training of next stage, thus achieving self-training in an iterative way.

\textbf{Multi-stage Self-training} \cite{li2018deeper}. This pretext task is proposed to leverage the abundant unlabeled nodes to help training. Given both the labeled set $\mathcal{D}_L^t$ and unlabeled set $\mathcal{D}_U^t$ in the iteration step $t$, the graph encoder $f_\theta(\cdot)$ is first trained on the labeled set $\mathcal{D}_L^t$, as follows
\begin{equation}
\mathcal{L}_{node}\left(\theta, \mathbf{A}, \mathbf{X}, \mathcal{D}_{L}^t\right)=\sum_{\left(v_{i}, y_{i}^t\right) \in \mathcal{D}_{L}^t} \ell\Big(f_{\theta}(\mathbf{A},\mathbf{X})_{v_{i}}, y_{i}^t\Big)
\label{equ:ms}
\end{equation}

\noindent and then applied to make predictions $\widehat{\mathcal{Y}}^t=\{\widehat{y}_i^t \mid v_i\in\mathcal{V}_U^t\}$ on the unlabeled set $\mathcal{V}_U^t$. Then the predicted labels (as well as corresponding nodes) with $K$-top high confidence
\begin{equation}
\mathcal{D}_N^t=\left\{(v_i, \widehat{y}_i^t)| \mid \widehat{y}_i^t \text { in top-K confidence of }\widehat{\mathcal{Y}}^t\right\}
\end{equation} 

\noindent are considered as the pseudo-labels and moved to the labeled node set $\mathcal{D}_L^t$ to obtain an updated labeled set $\mathcal{D}_L^{t+1}=\mathcal{D}_L^t \bigcup \mathcal{D}_N^t$ and an updated unlabeled set $\mathcal{D}_U^{t+1}=\mathcal{D}_U^t / \mathcal{D}_N^t$. Finally, a fresh graph encoder is trained on the updated labeled set $\mathcal{D}_L^{t+1}$, and the above operations are performed multiple times in an iterative manner.

\textbf{Node Clustering or Partitioning} \cite{you2020does}. Compared to Multi-stage Self-training, the pretext task of Node Clustering pre-assigns a pseudo-label $y_i$, e.g., the cluster index, to each node $v_i$ by some unsupervised clustering algorithms. The learning objective of this pretext task is then formulated as a classification problem, as follows
\begin{equation}
\mathcal{L}_{ssl}\left(\theta, \mathbf{A}, \mathbf{X}\right)=\frac{1}{N}\sum_{v_i \in \mathcal{V}} \ell\Big(f_{\theta}(\mathbf{A},\mathbf{X})_{v_{i}}, y_{i}\Big)
\end{equation}

When node attributes are not available, another choice to obtain pseudo-labels is based on the topology of a given graph structure or adjacency matrix. Specifically, graph partitioning \cite{blondel2008fast,traag2019louvain} is to partition the nodes of a graph into roughly equal subsets, such that the number of edges connecting nodes across subsets is minimized. To absorb the advantages of both attributive- and structural-based clustering, \textbf{CAGNN} \cite{zhu2020cagnn} combines the node clustering and node partitioning to proposed a new pretext task. Concretely, it first assigns cluster indices as pseudo labels but follows a topology refining process that refines the clusters by minimizing the inter-cluster edges.

\textbf{M3S} \cite{sun2020multi}. Combining Multi-stage Self-training with Node Clustering, M3S applies DeepCluster \cite{caron2018deep} and the alignment mechanism as a self-checking mechanism, thus providing stronger self-supervision. Specifically, a $K$-mean clustering algorithm is performed on node representations $\mathbf{H}^{(t)}$ learned in the iteration step $t$ (rather than $\mathbf{X}$) and the clustered pseudo-label $\mathcal{D}_N^t$ that matches the prediction of the classifier in the last iteration step $t-1$ will added to the labeled set to obtain an updated labeled set $\mathcal{D}_L^{t+1}=\mathcal{D}_L^t \bigcup \mathcal{D}_N^t$. Finally, a fresh model will be trained on the labeled set $\mathcal{D}_L^{t+1}$ with the objective defined as Equ.~\ref{equ:ms}.

\textbf{Cluster Preserving} \cite{hu2019pre}. An important characteristic of real-world graphs is the cluster structure, so we can consider the cluster preservation as a self-supervised pretext task. The unsupervised clustering algorithms are first applied to group nodes in a graph into $K$ non-overlapping clusters $\{C_k\}_{k=1}^K$, then the cluster prototypes can be obtained by $c_k=\text{AGGRATE}(\{f_{\theta}(\mathbf{A},\mathbf{X})_{v_{i}} \mid v_i\in C_k\})$. The mapping function $g_\omega(\cdot)$ is used to estimate the similarity of node $v_i$ with the cluster prototype $c_k$, e.g., the probability $\widehat{y}_{i,k}$ that node $v_i$ belongs to cluster $C_k$ is defined as follows,
\begin{equation}
\widehat{y}_{i,k}=\frac{\exp\Big(g_\omega(f_{\theta}\big(\mathbf{A},\mathbf{X})_{v_{i}}, c_k\big)\Big)}{\sum_{k=1}^K\exp\Big(g_\omega(f_{\theta}\big(\mathbf{A},\mathbf{X})_{v_{i}}, c_k\big)\Big)}
\end{equation}

\noindent Finally, the objective of Cluster Preserving is defined as
\begin{equation}
\mathcal{L}_{ssl}\left(\theta, \mathbf{A}, \mathbf{X}\right)=-\frac{1}{N}\sum_{v_i\in\mathcal{V}}\sum_{k=1}^K y_{i,k}\log(\widehat{y}_{i,k})
\end{equation}

\noindent where the ground-truth label $y_{i,k}=1$ if node $v_i$ is grouped into cluster $C_k$, otherwise $y_{i,k}=0$. 


\subsection{Domain Knowledge-based Prediction (DK)}
The formation of real-world graphs usually obeys specific rules, e.g., the links between atoms in molecular graphs are bounded by valence bonding theory, while cross-cited papers in citation networks generally have the same topic or authors. Therefore, extensive expert knowledge can be incorporated as a \emph{prior} into the design of pretext tasks.

\textbf{Contextual Molecular Property Prediction} \cite{rong2020self} incorporates domain knowledge about biological macromolecules to design molecule-specific pretext tasks. Given a node $v_i$, it samples its $k$-hop neighborhood nodes and edges as a local subgraph and then extracts statistical properties of this subgraph. Specifically, it counts the number of occurrence of (node, edge) pairs around the center node $v_i$ and then list all the node-edge count terms in alphabetical order, which constitutes the final property, e.g., C$\_$N-DOUBLE1$\_$O-SINGLE1 in Fig.~\ref{fig:5} (d). With plenty of context-aware properties $\mathcal{P}=\{p_k\}_{k=1}^K$ pre-defined, the contextual property prediction can be defined as a multi-class
prediction problem with one class corresponds to one contextual property, as follows
\begin{equation}
\mathcal{L}_{ssl}\left(\theta, \mathbf{A}, \mathbf{X}\right)=\frac{1}{N}\sum_{v_i\in\mathcal{V}}\ell\Big( f_{\theta}(\mathbf{A},\mathbf{X})_{v_i}, y_{i}\Big)
\end{equation}

\noindent where $\ell(\cdot)$ denotes the cross entropy loss, and $y_i=k$ if the molecular property of node $v_i$ is $p_k$.

\textbf{Graph-level Motif Prediction} \cite{rong2020self}. Motifs are recurrent sub-graphs among the input graph, which are prevalent in molecular graphs. One important class of motifs in molecules are functional groups, which encodes the rich domain knowledge of molecules and can be easily detected by professional softwares, such as RDKit. Suppose considering the presence of $K$ motifs $\{m_k\}_{k=1}^K$ in the molecular data, then for one specific molecule graph $g_i=(\mathbf{A}_{i},\mathbf{X}_{i})\in\mathcal{G}$, it detects whether each motif shows up in $g_i$ and use it as the label $\mathbf{y}_i\in\mathbb{R}^K$. Specifically, if $m_k$ shows up in $g_i$, the $k$-th elements $\mathbf{y}_{i,k}$ will be set to 1, otherwise 0. Formally, the learning objective of the motif prediction task is formulated as a multi-label classification problem, as follows
\begin{equation}
\mathcal{L}_{ssl}\left(\gamma, \mathcal{G}\right)=\frac{1}{|\mathcal{G}|}\sum_{g_{i}\in \mathcal{G}} \ell\Big(f_{\gamma}(\mathbf{A}_{i},\mathbf{X}_{i}), \mathbf{y}_{i}\Big)
\end{equation}

\noindent where $\ell(\cdot)$ deontes the binary cross entropy loss.
\section{Summary of the Implementation}
A summary of the surveyed works is presented in Fig.~\ref{fig:6}, and \textbf{Appendix D} lists their properties, including graph property, pretext task, augmentation, objective function, training strategy, and publication year. Furthermore, we show in \textbf{Appendix E} the implementation details of surveyed works, such as downstream tasks, evaluation metrics, and datasets.


\subsection{Downstream Tasks}
The graph SSL methods are generally evaluated on three levels of graph tasks: node-level, link-level, and graph-level. Among them, the \textbf{graph-level tasks} are usually performed on multiple graphs in the form of inductive learning. Commonly used graph-level tasks include graph classification and graph regression. The \textbf{link-level tasks} mainly focus on link prediction, that is, given two nodes, the objective is to predict whether a link (edge) exists between them. On the other hand, the \textbf{node-level tasks} are generally performed on a large graph in the form of transductive learning. Depending on whether labels are provided, it can be divided into three categories: node regression, node classification, and node clustering. The node classification and node regression are usually performed with partial known labels. Instead, the node clustering is performed in a more challenging unsupervised manner and adopted when the performance of the node classification is not sufficiently distinguishable.

\subsection{Evaluation Metrics}
For \textbf{graph classification} tasks, the commonly used evaluation metrics include ROC-AUC and Accuracy (Acc); while for \textbf{graph regression} tasks, Mean Absolute Error (MAE) is used. In terms of \textbf{link prediction} tasks, ROC-AP, ROC-PR, and ROC-AUC are usually used as evaluation metrics. Besides, \textbf{node regression} tasks are usually evaluated by metrics including MAE, Mean Square Error (MSE), and Mean Absolute Percentage Error (MAPE). In addition to Accuracy, \textbf{node classification} tasks also adopt F1-score for single-label classification and Micro-F1 (or Macro-F1) for multi-label classification. Moreover, \textbf{node clustering} tasks often adopt the same metrics used to evaluate the unsupervised clustering, such as Normalized Mutual Information (NMI), Adjusted Rand Index (ARI), Accuracy, etc.

\subsection{Datasets}
The statistics of a total of 41 datasets are available in \textbf{Appendix F}. Commonly used datasets for graph self-supervised learning tasks can be divided into five categories: citation networks, social networks, protein networks, molecule graphs, and others. (1) \textit{Citation Networks.} In citation networks, nodes usually denote papers, node attributes are some keywords in papers, edges denote cross-citation, and categories are topics of papers. Note that nodes in the citation networks may also sometimes indicate authors, institutions, etc. (2) \textit{Social Networks.} The social network datasets consider entities (e.g., users or authors) as nodes, their interests and hobbies as attributes, and their social interactions as edges. The widely used social network datasets for self-supervised learning are mainly some classical graph datasets, such as Reddit \cite{hamilton2017inductive}, COLLAB \cite{yanardag2015deep}. (3) \textit{Molecule Graphs.} In molecular graphs, nodes represent atoms in the molecule, the atom index is indicated by the node attributes, and edges represent bonds. Molecular graph datasets typically contain multiple graphs and are commonly used for tasks such as graph classification and graph regression, e.g., predicting molecular properties. (4) \textit{Protein Networks.} The protein networks can be divided into two main categories - Protein Molecule Graph and Protein Interaction Network - based on the way they are modeled. The Protein Molecule Graph is a particular type of molecule graph, where nodes represent amino acids, and an edge indicates the two connected nodes are less than 6 angstroms apart. The commonly used datasets include PROTEINS \cite{dobson2003distinguishing} and D\&D \cite{dobson2003distinguishing}, used for chemical molecular property prediction. The other branch is Protein Interaction Networks, where nodes denote protein molecules, and edges indicate their interactions. The commonly used dataset is PPI \cite{zitnik2017predicting}, used for graph biological function prediction. (5) \textit{Other Graphs.} In addition to the four types of datasets mentioned above, there are some datasets that are less common or difficult to categorize, such as image, traffic, and co-purchase datasets.


\subsection{Codes in Github}
The open-source codes are beneficial to the development of the deep learning community. A summary of the open-source codes of 71 surveyed works is presented in \textbf{Appendix G}, where we provide hyperlinks to their open-source codes. Most of these methods are implemented on GPUs based on Pytorch or Tensorflow libraries. Moreover, we have created a GitHub repository \url{https://github.com/LirongWu/awesome-graph-self-supervised-learning} to summarize the latest advances in graph SSL, which will be updated in real-time as more papers and their codes become available.

\subsection{Experimental Study}
\textcolor{mark}{To make a fair comparison, we select two important downstream tasks, node classification and graph classification, and provide the classification performance of various classical algorithms on 15 commonly used graph datasets in \textbf {Appendix H}. Due to space limitations, please refer to the appendix for more experimental results and analysis.}

\section{Discussion}
\textcolor{mark}{We begin with some discussion and summary of the connections and developments between various methods. To present a clearer picture of the development lineage of various graph SSL methods, we provide a complete timeline in \textbf{Appendix I}, listing the publication dates of some key milestones. Besides, we provide the inheritance connections between methods to show how they are developed. Furthermore, we provide short descriptions of contributions to some seminal works to highlight their importance.}

\textcolor{mark}{DGI \cite{velickovic2019deep} is a pioneering work for graph contrastive learning, originally designed specifically for node classification on attribute graphs, and later extended to other types of graphs, resulting in new variants such as HDGI \cite{ren2019heterogeneous} for heterogeneous graphs, STDGI \cite{opolka2019spatio} for spatial-temporal graphs, DMGI \cite{park2020unsupervised} for multiplexed graphs, and BiGI \cite{cao2021bipartite} for bipartite graphs. Besides, InfoGraph \cite{sun2019infograph} extends DGI to global-context contrasting, achieving state-of-the-art performance on multiple graph classification datasets. With the focus shifted from local nodes and global graphs to subgraphs, GCC \cite{qiu2020gcc} proposes the first subgraph-level context-context contrasting framework, where subgraphs sampled from the same graph are considered as the positive pair.}

\textcolor{mark}{Different from DGI, GRACE \cite{zhu2020deep} focuses on contrasting views at the node-level by generating multiple augmented graphs through handcrafted augmentations and then encouraging consistency between the same nodes in different views. GCA \cite{zhu2020graph} adopts a similar framework to GRACE, but focuses on designing the \textit{adaptive} augmentation strategy. Similarly, GROC \cite{jovanovic2021towards} claims that gradient information can be used to guide data augmentation and proposes a gradient-based graph topology augmentation that further improves the performance of GRACE and GCA. The same local-local contrasting as GRACE, but BGRL \cite{thakoor2021bootstrapped} is inspired by BYOL \cite{zbontar2021barlow} and explores for the first time whether negative samples are a must for graph contrastive learning.}

\textcolor{mark}{The focus on three different levels of nodes, edges, and structures has led to different generative methods such as Graph Completion \cite{you2020does}, Edge Feature Masking \cite{hu2019strategies}, and Adjacency Matrix Reconstruction \cite{zhu2020self}, respectively. Moreover, due to their simplicity and effectiveness, these methods have been widely used in algorithms such as GPT-GNN \cite{hu2020gpt} and recommendation applications such as Pretrain-Recsys \cite{hao2021pre}. In terms of predictive methods, the basic difference between methods is how to obtain pseudo-labels, and there are three main means: (1) numerical statistics, such as Node Property Prediction and PairwiseAttrSim \cite{jin2020self}; (2) prediction results from the previous training stage, such as CAGNN \cite{zhu2020cagnn} and M3S \cite{sun2020multi}; (3) domain knowledge, such as Molecular Property Prediction and Global Motif Prediction \cite{rong2020self}.}

\textcolor{mark}{\textbf{Discussion on Pros and Cons.}
Next, we will discuss the advantages and disadvantages of some classical algorithms on four aspects: innovation, accessibility, effectiveness (performance), and efficiency, based on which we divide existing algorithms into four categories: (1) Pioneering work. Representative works include DGI \cite{velickovic2019deep}, InfoGraph \cite{sun2019infograph}, HDGI  \cite{ren2019heterogeneous}, STDGI \cite{opolka2019spatio}, etc. These methods, for the first time, apply contrastive learning to a novel downstream task or graph type, showing promising innovations and inspiring many follow-up researches. However, as early attempts, they often perform relatively poorly on downstream tasks and with high computational complexity, compared to some subsequent works. (2) Knowledge-based work. There are some works that combine self-supervised learning techniques with prior knowledge to obtain excellent performance on downstream tasks. For example, Molecular Property Prediction \cite{rong2020self} combines domain knowledge, i.e., molecular properties, while LCC \cite{ren2021label} introduces label information into the computation of supervision signals. These knowledge-based methods usually achieve fairly good performance due to the introduction of additional information, but this also limits their applicability to other tasks and graph types. (3) There is some work aimed at building on existing work and pursuing state-of-the-art performance on a variety of datasets, among which representative works include K2SL \cite{yu2020self}, BGRL \cite{thakoor2021bootstrapped}, and SUGAR \cite{sun2021sugar}. While these works have achieved state-of-the-art performance on a variety of downstream tasks, they are largely incremental contributions to previous seminal work (e.g., DGI) and are relatively weak on innovation and accessibility. (4) Most generative and predictive methods are less effective than contrastive methods, but are generally very simple to implement, easy to combine with existing frameworks, have lower computational complexity, and exhibit better applicability and efficiency.}

\textbf{Pretext Tasks for Complex Types of Graph.}
Most of the existing work is focused on the design of pretext tasks, especially on attribute graphs, with little effort to other more complex graph types, such as spatial-temporal and heterogeneous graphs. Moreover, these pretext tasks usually utilize only node-level or graph-level features, limiting their ability to exploit richer information, such as temporal information in spatial-temporal graphs and relation information in heterogeneous graphs. \textcolor{mark}{As a result, how to design more suitable pretext tasks can be considered from three aspects: (1) designing \emph{graph type-specific} pretext tasks that adaptively pick the most suitable tasks depending on the type of graph; (2) incorporating temporal or heterogeneous information (in the form of prior knowledge) into the pretext task design; (3) taking the automated design of pretext tasks as a new research topic from the perspective of automatic learning.}

\textbf{Lack of Theoretical Foundation.}
Despite the great success of graph SSL on various tasks, they mostly draw on the successful experience of SSL on CV and NLP domains. In other words, most existing graph SSL methods are designed with \emph{intuition}, and their performance gains are evaluated by \emph{empirical experiments}. The lack of sufficient theoretical foundations behind the design has led to both performance bottlenecks and poor explainability. Therefore, we believe that building a solid theoretical foundation for graph SSL from a graph theory perspective and minimizing the gap between the theoretical foundation and empirical design is also a promising future direction. \textcolor{mark}{For example, an important problem for graph SSL is whether mutual information maximization is the only means to achieve graph contrastive learning? Such problems have been explored in \cite{tschannen2019mutual} for image data, but how to extend them to the graph domain is not yet available. In Sec.~\ref{sec:3.4}, in addition to MI estimators such as InforNCE, we have introduced some contrastive objectives that are not based on mutual information, such as triplet margin and quadruplet loss. However, how to theoretically analyze the connection between these losses and mutual information needs to be further explored.}

\textbf{Insufficient Augmentation Strategy.}
Recent advances in the field of visual representation learning \cite{he2020momentum,chen2020simple} are mainly attributed to a variety of data augmentation strategies, such as resize, rotation, coloring, etc. However, due to the inherent non-Euclidean nature of graph data, it is difficult to directly apply existing image-based augmentation to graphs. Moreover, most augmentation strategies on graphs are limited to adding/removing nodes and edges or their combination to achieve the asserted SOTA. To further improve the performance of SSL on graphs, it is a promising direction to design more efficient augmentation strategies. \textcolor{mark}{More importantly, the design of the augmentation strategy should follow some well-designed guidelines instead of relying entirely on subjective intuition. In summary, we argue that the design of graph augmentation should be based on the following four guidelines: (1) Applicability, graph augmentation should ideally be a plug-and-play module that can be easily combined with the existing self-supervised learning frameworks. (2) Adaptability, some work \cite{jin2020self, thakoor2021bootstrapped} have pointed out that different datasets and task types may require different augmentations, so how to design the date-specific and task-specific augmentation strategy is a potential research topic.  (3) Efficiency, data augmentation should be a lightweight module that does not bring a huge computational burden to the original implementation. (4) Dynamic, with the ability to dynamically update the augmentation strategy as the training proceeds.}

\textcolor{mark}{\textbf{Inefficient Negative Sampling Strategy.}
The selection of high-quality negative samples is a crucial issue. The most common sampling strategy is uniform sampling, but this has been shown to be very informative \cite{chuang2020debiased,kalantidis2020hard,robinson2020contrastive}. The problem of how to better obtain negative samples has been well explored in the field of computer vision. For example, \cite{chuang2020debiased} presents the debiased contrastive learning that directly corrects the sampling bias of negative samples. Besides, \cite{kalantidis2020hard} takes advantage of mixup techniques to directly synthesize hard negative samples in the embedding space. Moreover, \cite{robinson2020contrastive} develops a family of unsupervised sampling strategies for user-controllable negative sample selection. Despite the great success, these methods, specifically designed for image data, may be difficult to apply directly to non-Euclidean graph data. More importantly, accurate estimation of hard negative samples becomes more difficult when label information is not available. Therefore, how to reduce the gap between ideal and practical contrastive learning by a decent negative sampling strategy requires more exploration.}

\textbf{Lack of Explainability.}
Though existing graph SSL methods have achieved excellent results on various downstream tasks, we still do not know exactly what has been learned from self-supervised pretext tasks. Which of the feature patterns, significant structures, or feature-structure relationships has been learned by self-supervision? Is this learning explicit or implicit? Is it possible to find interpretable correspondences on the input data? These are important issues for understanding and interpreting model behavior but are missing in current graph SSL works. Therefore, we need to explore the interpretability of graph SSL and perform a deep analysis of model behavior to improve the generalization and robustness of existing methods for security- or privacy-related downstream tasks.

\textbf{Margin from Pre-training to Downstream Tasks.}
Pre-training with self-supervised tasks and then using the pre-trained model for specific downstream tasks, either by fine-tuning or freezing the weights, is a common training strategy in graph SSL \cite{hao2021pre,shang2019pre,zhang2021pre}. However, how shall we transfer the pre-trained knowledge to downstream tasks? Though numerous strategies have been proposed to address this problem in the CV and NLP domains \cite{zhuang2020comprehensive}, they are difficult to apply directly to graphs due to the inherent non-Euclidean structure of graphs. Therefore, it is an important issue to design graph-specific techniques to minimize the margin between pre-training and downstream tasks.
\section{Conclusion}
\textcolor{mark}{A comprehensive survey of the literature on graph self-supervised learning techniques is conducted in this paper. We develop a unified mathematical framework for graph SSL. Moreover, we summarize the implementation details in each work and show their similarities and differences. More importantly, we are the first survey to provide a detailed experimental study on self-supervised learning, setting the stage for the future development of graph SSL. Finally, we point out the technical limitations of the current research and provide promising directions for future work on graph SSL. We hope this survey will inspire follow-up researchers to focus on other important but easy-to-miss details such as theoretical foundations, explainability, etc., in addition to model performance on downstream tasks.}

\clearpage

\renewcommand\thefigure{A\arabic{figure}}
\renewcommand\thetable{A\arabic{table}}
\setcounter{table}{0}
\setcounter{figure}{0}

\appendix
\section*{A. Contrastive Methods}
In this section, we will continue with some contrastive methods for graph SSL, but to avoid over-redundancy, we will not provide detailed mathematical formulas for some \emph{relatively less important} works. These methods include (1) methods that are essentially the same as the frameworks already presented in the main paper with only minor differences; (2) methods that have not been accepted and are only publicly available on platforms such as arxiv, OpenReview, etc.; (3) application methods, where graph SSL is only one of the adopted techniques or tricks and is not the focus of their works; (4) methods with only minor performance improvement on the benchmark datasets.

\subsection*{A.1 Global-Global Contrasting}

\noindent \textbf{Iterative Graph Self-Distillation (IGSD)} \cite{zhang2020iterative} is a graph self-supervised learning paradigm that iteratively performs the teacher-student distillation with graph augmentations. Specifically, given a graph $g_i=(\mathbf{A}_i,\mathbf{X}_i)\in\mathcal{G}$, it performs a series of augmentation transformations $\mathcal{T}(\cdot)$ to generate an augmented graph $\widetilde{g}_i=(\widetilde{\mathbf{A}}_i,\widetilde{\mathbf{X}}_i)=\mathcal{T}(\mathbf{A}_i,\mathbf{X}_i)$. Then two graph encoder $f_{\theta_1}(\cdot)$ and $f_{\theta_2}(\cdot)$ as well as a $\text{READOUT}$ function are applied to obtain the graph-level representations $\mathbf{h}_{g_i} = \text{READOUT}\left(f_{\theta_1}(\mathbf{A}_i,\mathbf{X}_i)\right)$ and $\widetilde{\mathbf{h}}_{\widetilde{g}_i} = \text{READOUT}(f_{\theta_2}(\widetilde{\mathbf{A}}_i,\widetilde{\mathbf{X}}_i))$, respectively. Moreover, a prediction head $g_\omega(\cdot)$ is used in the student network to obtain $g_\omega(\mathbf{h}_{g_i})$ and $g_\omega(\widetilde{\mathbf{h}}_{\widetilde{g}_i})$. To contrast 
representations $\mathbf{h}_{g_i}$ and $\widetilde{\mathbf{h}}_{\widetilde{g}_i}$, the symmetric consistency loss is defined as
\begin{equation}
\mathcal{D}\left(\mathbf{h}_{g_i},\widetilde{\mathbf{h}}_{\widetilde{g}_i}\right)=\left\|g_\omega(\mathbf{h}_{g_i})-\widetilde{\mathbf{h}}_{\widetilde{g}_i}\right\|^{2} + \left\|g_\omega(\widetilde{\mathbf{h}}_{\widetilde{g}_i})-\mathbf{h}_{g_i}\right\|^{2}
\end{equation}

\noindent Finally, the learning objective is defined as follows
\begin{equation}
\max_{\theta_1, \omega} \frac{1}{|\mathcal{G}|} \sum_{g_i\in\mathcal{G}}\log \frac{\exp\left(\mathcal{D}(\mathbf{h}_{g_i},\widetilde{\mathbf{h}}_{\widetilde{g}_i})\right)}{\sum_{j=1}^N\exp\left(\mathcal{D}(\mathbf{h}_{g_i},\widetilde{\mathbf{h}}_{\widetilde{g}_j})\right)}
\end{equation}

\noindent where the negative samples to contrast with $\mathbf{h}_{g_{i}}$ is $Neg(\mathbf{h}_{g_{i}})=\{\widetilde{\mathbf{h}}_{\widetilde{g}_j}\}_{g_j\in\mathcal{G},j\neq i}$. The parameter $\theta_2$ of teacher network $f_{\theta_2}(\cdot)$ are updated as an exponential moving average (EMA) of the student network parameters $\theta_1$, as follows
\begin{equation}
\theta_{2} \longleftarrow \alpha \theta_{2}+(1-\alpha) \theta_{1}
\end{equation}

\noindent where $\alpha\in[0,1)$ is the momentum weight to control the evolving speed of $\theta_2$.
\newline

\noindent \textbf{Domain-Agnostic Contrastive Learning (DACL)} \cite{verma2020towards} uses the \emph{Mixup} noise to create positive and negative examples by mixing data samples either at the input or embedding spaces. Specifically, given a graph $g_i=(\mathbf{A}_i,\mathbf{X}_i)\in\mathcal{G}$, it applies a graph encoder $f_{\theta}(\cdot)$ and a $\text{READOUT}$ function to obtain the graph-level representation $\mathbf{h}_{g_i} = \text{READOUT}( f_{\theta}(\mathbf{A}_i,\mathbf{X}_i))$. Then it performs the mixup transformations to generate two positive pairs , e.g., $\mathbf{h}^{(1)}_{g_i} = \lambda_1\mathbf{h}_{g_i}+(1-\lambda_1)\mathbf{h}_{g_m}$ and $\mathbf{h}^{(2)}_{g_i} = \lambda_2\mathbf{h}_{g_i}+(1-\lambda_2)\mathbf{h}_{g_n}$, where $\lambda_1$ and $\lambda_2$ are sampled from a Gaussian distribution and $\mathbf{h}_{g_m}$ and $\mathbf{h}_{g_n}$ are randomly sampled from $\{\mathbf{h}_{g_k}\}_{k=1,k\neq i}^N$, respectively. Finally, the learning objective is defined as
\begin{equation}
\max _{\theta, \omega} \frac{1}{|\mathcal{G}|} \sum_{g_i\in\mathcal{G}}\mathcal{MI}\left(g_\omega(\mathbf{h}^{(1)}_{g_{i}}), g_\omega(\mathbf{h}^{(2)}_{g_i})\right)
\end{equation}

\noindent where $g_\omega(\cdot)$ is a nonlinear prediction head, and the contrasted negative samples is $Neg\left(g_\omega(\mathbf{h}^{(1)}_{g_{i}})\right)=\{g_\omega(\mathbf{h}^{(c)}_{g_{j}})\}_{j\neq i; c = 1, 2}$.

\subsection*{A.2 Local-Local Contrasting}

\noindent \textbf{KS2L} \cite{yu2020self} is a novel self-supervised knowledge distillation framework, with two complementary intra- and cross-model knowledge distillation modules. Given a graph $g=(\mathbf{A},\mathbf{X})$, it first applies two linear mapping functions to obatain $\mathbf{Z}^{(1)}=g_{\omega_1}(\mathbf{X})$ and $\mathbf{Z}^{(2)}=g_{\omega_2}(\mathbf{X})$ and then uses two graph encoder $f_{\theta_1}(\cdot)$ and $f_{\theta_2}(\cdot)$ to obtain node embedding matrices $\mathbf{H}^{(1)}=f_{\theta_1}(\mathbf{A},\mathbf{Z}^{(1)})$ and $\mathbf{H}^{(2)}=f_{\theta_2}(\mathbf{A},\mathbf{Z}^{(2)})$. Finally, the learning objective is defined as follows
\begin{equation}
\max _{\theta_1,\theta_2, \omega_1, \omega_2} \frac{1}{2|\mathcal{E}|} \sum_{e_{i,j}\in\mathcal{E}}\mathcal{MI}\left[\left(\mathbf{h}_i^{(1)},\mathbf{z}_j^{(1)}\right)+\left(\mathbf{h}_i^{(1)},\mathbf{z}_j^{(2)}\right)\right]
\end{equation}

\noindent where the first term is used for \emph{intra-model} knowledge distillation, and the negative samples to contrast with $\mathbf{h}_i^{(1)}$ is $Neg(\mathbf{h}_i^{(1)})=\{\mathbf{z}_k^{(1)}\}_{e_{i,k}\notin\mathcal{E}}$. The second term is used for \emph{cross-model} knowledge distillation, and the negative samples to contrast with $\mathbf{h}_i^{(1)}$ is $Neg(\mathbf{h}_i^{(1)})=\{\mathbf{z}_k^{(2)}\}_{e_{i,k}\notin\mathcal{E}}$.
\newline

\noindent \textbf{$\text{C}\text{G}^3$} \cite{wan2020contrastive} adopts a similar framework to GRACE \cite{zhu2020deep}, but perfroms the same label-level contasting as LCC \cite{ren2021label}. Given a graph $g=(\mathbf{A},\mathbf{X})$, it first applies a localized graph convolution encoder $f_{\theta_1}(\cdot)$ and a hierarchical graph convolution encoder $f_{\theta_2}(\cdot)$ to generate their node embedding matrices $\mathbf{H}^{(1)}=f_{\theta_1}(\mathbf{A},\mathbf{X})$ and $\mathbf{H}^{(2)}=f_{\theta_2}(\mathbf{A},\mathbf{X})$. To incorporate the scarce yet valuable label information for training, it uses a supervised label contrastive loss as follows:
\begin{equation}
\max_{\theta} \frac{1}{2N}\sum_{i=1}^{N}\left[\mathcal{L}^{(1)}_{lc}+\mathcal{L}^{(2)}_{lc}\right]
\end{equation}

\noindent where $\mathcal{L}^{(1)}_{lc}$ and $\mathcal{L}^{(2)}_{lc}$ are defined as
\begin{equation}
\mathcal{L}^{(1)}_{lc}=\log \frac{\sum_{k=1}^m\mathbb{I}_{y_k=y_i}\cdot\exp\left(\mathbf{h}_i^{(1)}\cdot\mathbf{h}_k^{(2)}\right)}{\sum_{j=1}^m\exp\left(\mathbf{h}_i^{(1)}\cdot\mathbf{h}_j^{(2)}\right)}
\end{equation}

\begin{equation}
\mathcal{L}^{(2)}_{lc}=\log \frac{\sum_{k=1}^m\mathbb{I}_{y_k=y_i}\cdot\exp\left(\mathbf{h}_i^{(2)}\cdot\mathbf{h}_k^{(1)}\right)}{\sum_{j=1}^m\exp\left(\mathbf{h}_i^{(2)}\cdot\mathbf{h}_j^{(1)}\right)}
\end{equation}

\noindent where $m$ is the number of all labeled nodes, and $\mathbb{I}_{y_k=y_i}$ is an indicator function to determine whether the label of node $v_k$ is the same as that of node $v_i$.
\newline

\noindent \textbf{PT-DGNN} \cite{zhang2021pre} is proposed to perfrom pre-training on dynamic graph $g=(\mathbf{A}^{(t)}, \mathbf{X})$, where $\mathbf{A}^{(t)}_{i,j}=1$ denotes the interaction between node $v_i$ and $v_j$ at time $t$, that is, $t$ is the timestamp of edge $e_{i,j}$. It first applies the dynamic subgraph sampling (DySS) to obtain a subgraph $g_S=(\mathbf{A}_s, \mathbf{X}_s)$ for pre-training. DySS mainly includes three steps: 1) Randomly select $s$ nodes as the sampling initial points; 2) Put the first-order neighbors of these nodes into the candidate pool and save their timestamp as weight to calculate sampling probability; 3) Select final $s$ nodes according to sampling probability. Furthermore, it genetates an augmentated graph $g_a=(\mathbf{A}_a, \mathbf{X}_a)=\mathcal{T}(\mathbf{A}_s, \mathbf{X}_s)$ by performing the attribute masking and edge perturbation to obtain , where the masked node and edge set are denoted as $\mathcal{S}$ and $\mathcal{M}$, respectively. Then a shared graph encoder $f_{\theta}(\cdot)$ is applied to obtain the node embedding matrix $\mathbf{H}=f_{\theta}(\mathbf{A}_a,\mathbf{X}_a)$. Finally, the objective of \emph{dynamic edge generation} is defined as
\begin{equation}
\max_{\theta} \frac{1}{|\mathcal{V}_a|} \sum_{v_i\in \mathcal{V}_a}  \sum_{e_{i,j}\in\mathcal{M}}\log\frac{\exp\Big(\mathcal{D}(\mathbf{h}_i, \mathbf{h}_j)\Big)}{\sum_{v_{k}\in \mathcal{V}_a} \exp\Big(\mathcal{D}(\mathbf{h}_i, \mathbf{h}_k)\Big)}
\end{equation}

\noindent where $\mathcal{V}_a$ is the node set of graph $g_a$. The learning objective of the \emph{node attribute generation} is defined as follows
\begin{equation}
\min_{\theta, \omega} \frac{1}{|\mathcal{S}|}\sum_{v_i\in\mathcal{S}}\big\|g_\omega(\mathbf{h}_i)-\mathbf{x}_i\big\|^2
\end{equation}

\noindent where $g_\omega(\cdot)$ is a node-level nonlinear prediction head.
\newline

\noindent \textbf{COAD} \cite{chen2020coad} applies graph SSL to expert linking, which aims at linking any external information of persons to experts in AMiner. Specifically, it first pre-trains the model by \emph{local-local} contrastive learning with the triplet margin loss and then fine-tunes the model by adversarial learning to improve the model transferability. 
\newline

\noindent \textbf{Contrast-Reg} \cite{ma2021improving} is a lightweight \emph{local-local} contrastive regularization term that adopts the InfoNCE loss to contrasts the node representation similarities of semantically similar (positive) pairs against those of negative pairs. Extensive theoretical analysis demonstrates that Contrast-Reg avoids the high scale of node representation norms and the high variance among them to improve the generalization performance.
\newline

\noindent \textbf{C-SWM} \cite{kipf2019contrastive} models the structured environments, such as multi-object systems, as graphs, and then utilizes a \emph{local-local} contrastive approach to perform the representation learning from environment interactions without supervision.

\subsection*{A.3 Local-Global Contrasting}

\noindent \textbf{High-order Deep Multiplex Infomax (HDMI)} \cite{jing2021hdmi} is proposed to achieve \emph{higher-order} mutual information maximization. Given a graph $g=(\mathbf{A},\mathbf{X})$, it first applies an augmentation transformation $\mathcal{T}(\cdot)$ to obtain $\widetilde{g}=(\widetilde{\mathbf{A}},\widetilde{\mathbf{X}}) = \mathcal{T}(\mathbf{A},\mathbf{X})$. Then a shared graph encoder $f_{\theta}(\cdot)$ is applied to obtain node embedding matrices $\widetilde{\mathbf{H}}=f_{\theta}(\widetilde{\mathbf{A}},\widetilde{\mathbf{X}})$ and $\mathbf{H}=f_{\theta}(\mathbf{A},\mathbf{X})$. Beside, a $\operatorname{READOUT}()$ function is used to obtain the graph-level representaion $\mathbf{h}_g=\operatorname{READOUT}(\mathbf{H})$. Finally, the learning objective is defined as follows
\begin{equation}
\begin{small}
\begin{aligned}
\max _{\theta,\lambda} \frac{1}{N} \sum_{v_i\in\mathcal{V}}
[\mathcal{MI}\left(\mathbf{h}_g, \mathbf{h}_i\right)
+\mathcal{MI}\left(\mathbf{x}_i, \mathbf{h}_i\right)
+\mathcal{MI}\left(\mathbf{h}_i,\mathbf{m}_i\right)]
\end{aligned}
\end{small}
\end{equation}

\noindent where $\mathbf{m}_i=\sigma(\mathbf{W}_m[\sigma(\mathbf{W}_h \mathbf{x}_i);\sigma(\mathbf{W}_g \mathbf{h}_g)])$, and $\lambda=(\mathbf{W}_m$, $\mathbf{W}_h$, $\mathbf{W}_g)$ are the weight parameters. The negative samples to contrast with $\mathbf{h}_g$ in the \emph{first} term is $Neg(\mathbf{h}_g)=\widetilde{\mathbf{h}}_{i}$. The negative samples to contrast with $\mathbf{x}_i$ in the \emph{second} term is $Neg(\mathbf{x}_i)=\widetilde{\mathbf{h}}_{i}$. The negative samples to contrast with $\mathbf{h}_i$ in the \emph{third }term is $Neg(\mathbf{h}_i)=\widetilde{\mathbf{m}}_{i}=\sigma(\mathbf{W}_m[\sigma(\mathbf{W}_h \widetilde{\mathbf{x}}_i);\sigma(\mathbf{W}_g \mathbf{h}_g)])$.
\newline

\noindent \textbf{Deep Multiplex Graph Infomax (DMGI)} \cite{park2020unsupervised} extents the idea of DGI to multiplex graphs where nodes are connected by multiple types of relations. For each relation type $r\in\mathcal{R}$ (corresponding to a relation graph $g^{(r)}=(\mathbf{A}^{(r)},\mathbf{X})$), a relation-type graph encoder $f_{\theta_r}(\cdot)$ is applied to obtain the relation-specific node embedding matrix $\mathbf{H}^{(r)}=f_{\theta_r}(\mathbf{A}^{(r)},\mathbf{X})$. Besides, it employs a $\operatorname{READOUT}$ function to obatain the graph-level representation $\mathbf{h}_{g^{(r)}}=\operatorname{READOUT}(\mathbf{H}^{(r)})$. Finally, it independently maximizes the mutual Information between the node embeddings $\mathbf{H}^{(r)}=\left\{\mathbf{h}_{1}^{(r)}, \mathbf{h}_{2}^{(r)}, \ldots, \mathbf{h}_{N}^{(r)}\right\}$ and the graph-level summary $\mathbf{h}_{g^{(r)}}$ pertaining to each graph $g^{(r)}$, defined as
\begin{equation}
\max _{\theta_r} \frac{1}{N} \sum_{v_i\in\mathcal{V}}\mathcal{MI}\left(\mathbf{h}_{g^{(r)}}, \mathbf{h}^{(r)}_i\right), r\in\mathcal{R}
\end{equation}

\noindent where the negative samples to contrast with $\mathbf{h}_{g^{(r)}}$ is $Neg(\mathbf{h}_{g^{(r)}})=\{\mathbf{h}_{j}^{(r)}\}_{v_j\in\mathcal{V},j\neq i}$. Similarly, it can learn another embedding matrix from the augmented graph $\widetilde{g}^{(r)}=(\widetilde{\mathbf{A}}^{(r)},\widetilde{\mathbf{X}})= \mathcal{T}(\mathbf{A}^{(r)},\mathbf{X})$ and also maximize the MI of the node embeddings $\widetilde{\mathbf{H}}^{(r)}=\left\{\widetilde{\mathbf{h}}_{1}^{(r)}, \widetilde{\mathbf{h}}_{2}^{(r)}, \ldots, \widetilde{\mathbf{h}}^{(r)}_N\right\}$ and the graph-level summary $\widetilde{\mathbf{h}}_{\widetilde{g}^{(r)}}$. However, as each $\mathbf{H}^{(r)}$ is trained independently for each $r\in\mathcal{R}$, these embedding matrices may fail to take advantage of the multiplexity of the network. Therefore, DMGI learns another consensus embedding matrix $\mathbf{Z}$ on which relation-specific node embedding matrices $\mathbf{H}^{(r)}$ and $\widetilde{\mathbf{H}}^{(r)}$ can agree with each other by optimizing the learning objective, as follows
\begin{equation}
\ell_{\mathrm{cs}}=\Big[\mathbf{Z}-\frac{1}{|\mathcal{R}|} \sum_{r \in \mathcal{R}} \mathbf{H}^{(r)}\Big]^{2}-\Big[\mathbf{Z}-\frac{1}{|\mathcal{R}|} \sum_{r \in \mathcal{R}} \widetilde{\mathbf{H}}^{(r)}\Big]^{2}
\end{equation}

\noindent where $\mathbf{Z}$ is defined as a set of \emph{trainable} parameters.
\newline

\noindent \textbf{HDGI} \cite{ren2019heterogeneous}. A meta-path of length $l$ is a sequence of nodes connected with heterogeneous edges, i.e., $\Phi: v_{1} \stackrel{t_{1}}{\longrightarrow} v_{2} \stackrel{t_{2}}{\longrightarrow} \ldots \stackrel{t_{l}}{\longrightarrow} v_{l}$, where $t_l\in\mathcal{T}^e$ denote the type of $l$-th edge in the meta-path. Given a meta-path $\Phi$, if there exist a meta-path $\Phi$ between node $v_i$ and node $v_j$, it defines that $v_i$ and $v_j$ are connected neighbors based on the meta-path $\Phi$, thus obtaining a meta-path based adjacent matrix $\mathbf{A}^{\Phi}$. Given a meta-path set $\{\Phi_k\}_{k=1}^K$, we can obtain $K$ meta-path based adjacent matrix $\left\{\mathbf{A}^{\Phi_{k}}\right\}_{k=1}^K$. HDGI first applies the augmentation to obtain $\widetilde{\mathbf{X}}, \{\widetilde{\mathbf{A}}^{\Phi_{k}}\}_{k=1}^K=\mathcal{T}(\mathbf{X}, \{\mathbf{A}^{\Phi_{k}}\}_{k=1}^K)$. Then $K$ graph encoder are applied to obtain node embedding matrices $\mathbf{H}^{(k)}=f_{\theta_k}(\mathbf{A}^{\Phi_{k}},\mathbf{X})$ and $\widetilde{\mathbf{H}}^{(k)}=f_{\theta_k}(\widetilde{\mathbf{A}}^{\Phi_{k}},\widetilde{\mathbf{X}})$ $(1\leq k \leq K)$. To obtain the more general node representations, the attention mechanism can be applied to fuse these representations $\mathbf{H}^{fuse}=\sum_{k=1}^K s_k \mathbf{H}^{(k)}$, where the attention scores $\{s_k\}_{k=1}^N$ are defined as follow
\begin{equation}
\begin{small}
\begin{aligned}
s_k\!=\!\frac{\exp(e^{(k)})}{\sum_{k=1}^K \exp(e^{(k)})}, e^{(k)}\!=\!\frac{1}{N}\sum_{i=1}^N \tanh(\mathbf{q}^T[\mathbf{W}\mathbf{h}^{(k)}_i+\mathbf{b}])
\end{aligned}
\end{small}
\end{equation}

\noindent where $1\leq k \leq K$, $\mathbf{W}$ and $\mathbf{b}$ are the shared weight matrix and shared bias vector, and $\mathbf{q}$ is a shared attention vector. Similarity, it can obtain the fused representation $\widetilde{\mathbf{H}}^{fuse}=\sum_{k=1}^K \widetilde{s}_k \widetilde{\mathbf{H}}^{(k)}$. Besides, a $\operatorname{READOUT}(\cdot)$ function are applied to obtain the global-level representation $\mathbf{h}_g=\operatorname{READOUT}(\mathbf{H}^{fuse})$. The objective is defined as:
\begin{equation}
\max _{\theta_1, \theta_2, \cdots, \theta_K, \gamma} \frac{1}{N} \sum_{v_i\in\mathcal{V}}\mathcal{MI}\left(\mathbf{h}_{g}, \mathbf{h}^{fuse}_i\right)
\end{equation}

\noindent where $\gamma=(\mathbf{W}, \mathbf{b}, \mathbf{q})$ are the weight parameters, and the negative samples to contrast with $\mathbf{h}_{g}$ is $Neg(\mathbf{h}_{g})=\widetilde{\mathbf{h}}^{fuse}_i$.
\newline

\noindent \textbf{DITNet} \cite{cheng2021drug} propose an end–to–end model to predict drug-target interactions on heterogeneous graphs. Specifically, it learns high-quality representations for downstream tasks by performing \emph{local-gobal} and \emph{context-global} contrasting. 

\subsection*{A.4 Local-Context Contrasting}

\noindent \textbf{Context Prediction} \cite{hu2019strategies} is proposed to maps nodes appearing in similar structural contexts to similar embeddings. It first picks up an anchor node set $\mathcal{S}$ from $\mathcal{V}$. Given an anchor node $v_i\in\mathcal{S}$, it defines its \emph{$K$-hop neighborhood graph} $g_i^{(1)}=(\mathbf{A}_i^{(1)},\mathbf{X}_i^{(1)})$ as all nodes and edges that are at most $K$-hops away from $v_i$. The \emph{context graph} represents a subgraph $g_2=(\mathbf{A}_i^{(2)},\mathbf{X}_i^{(2)})$ that is between $r_1$-hops (it requires $r_1 < K$) and $r_2$-hops away from $v_i$ (i.e., it is a ring of width $r_2 - r_1$). Two graph encoder $f_{\theta_1}(\cdot)$ and $f_{\theta_2}(\cdot)$ are then applied to encode two graphs as node embedding matrices $\mathbf{H}_i^{(1)}=f_{\theta_1}(\mathbf{A}_i^{(1)},\mathbf{X}_i^{(1)})$ and $\mathbf{H}_i^{(2)}=f_{\theta_2}(\mathbf{A}_i^{(2)},\mathbf{X}_i^{(2)})$. Besides, a $\operatorname{READOUT}(\cdot)$ function are further applied to obtain the subgraph-level representation, as follows
\begin{equation}
\mathbf{h}^{(g)}_i=\frac{1}{|\mathcal{M}_i|}\sum_{v_j\in\mathcal{M}_i}(\mathbf{H}_i^{(2)})_{v_j}
\end{equation}

\noindent where $\mathcal{M}_i$ is the set of those nodes that are shared between the neighborhood graph $g_i^{(1)}$ and the context graph $g_i^{(2)}$, and we refers to those nodes as \emph{context anchor nodes.} Finally, the learning objective is defined as follows
\begin{equation}
\max _{\theta_1,\theta_2} \frac{1}{|\mathcal{S}|} \sum_{v_i\in\mathcal{S}}\mathcal{MI}\left((\mathbf{H}^{(1)}_i)_{v_i}, \mathbf{h}_i^{(g)}\right)
\end{equation}

\noindent where $(\mathbf{H}^{(1)}_i)_{v_i}$ is the node representation of anchor node $v_i$ in the node embedding matrix $\mathbf{H}^{(1)}_i$. The negative samples to contrast with $(\mathbf{H}^{(1)}_i)_{v_i}$ is $Neg((\mathbf{H}^{(1)}_i)_{v_i})=\{\mathbf{h}_j^{(g)}\}_{v_j\in\mathcal{S},j\neq i}$.
\newline

\noindent \textbf{GraphLog} \cite{xu2021selfsupervised} is a unified graph SSL framework. In addition to the \emph{local-local} contrasting (similar to GRACE \cite{zhu2020deep}) and \emph{global-global} contrasting (similar to GraphGL \cite{you2020graph}), GraphLoG leverages an RPCL\cite{xu1993rival}-based clustering , e.g., $K$-means clustering, to learn hierarchical prototypes of graph data and then perform \emph{local-context} contrasting (similar to GIC \cite{mavromatis2020graph}) for each hierarchical layer, respectively. 
\newline

\noindent \textbf{MHCN} \cite{yu2021self} adopts a DGI-like auxiliary task to enhance social recommendation by leveraging high-order user relations. However, considering that the DGI-like local-global contrasting \cite{velickovic2019deep} stays at a coarse level and there is no guarantee that the encoder can distill sufficient information from the input data, MHCN extends the DGI to a fine-grained \emph{local-context} contrasting by exploiting the hierarchical structure in the input hypergraphs. 
\newline

\noindent \textbf{EGI} \cite{zhu2020transfer} considers transfer learning on graphs, i.e., the pre-training of GNNs. Specifically, unlike DGI that models the local-global mutual information, EGI samples a set of ego-subgraph and then directly optimizes the \emph{local-context} MI maximization between the structural input and output of GNN, with a particular focus on the structural information.

\subsection*{A.5 Context-Global Contrasting}

\noindent \textbf{SUGAR} \cite{sun2021sugar} is a novel hierarchical subgraph-level framework, which encourages subgraph embedding to be mindful of the global structural properties by maximizing their mutual information. For a given $g=(\mathbf{A},\mathbf{X})\in\mathcal{G}$, it first applies an augmentation transformation to obtain $\widetilde{g}=(\widetilde{\mathbf{A}},\widetilde{\mathbf{X}})=\mathcal{T}(\mathbf{A},\mathbf{X})$. Then it samples subgraphs from each graph and collect them into a subgraph candidate pool. Besides, a shared graph encoder $f_\theta(\cdot)$ and a \text{READOUT} funtion are applied to encode these subgraphs into subgraph-level representations $\{\mathbf{h}_1,\mathbf{h}_2,\cdots\}$ and $\{\widetilde{\mathbf{h}}_1,\widetilde{\mathbf{h}}_2,\cdots\}$. Next, $K$ striking subgraphs are selected from candidate pool by a reinforcement learning module and pooled into a sketched graph where each node $v_k$ ($1 \leq k \leq K$) corresponds to a subgraph with embedding $\mathbf{h}_k$ for graph $g$ ($\widetilde{\mathbf{h}}_k$ for graph $\widetilde{g}$). Finally, the learning objective is defined as follows
\begin{equation}
\max _{\theta}\frac{1}{|\mathcal{G}|} \sum_{g \in \mathcal{G}} \sum_{k=1}^K\mathcal{MI}\left(\mathbf{h}_{g}, \mathbf{h}_{k}\right)
\end{equation}

\noindent where $\mathbf{h}_g=\text{READOUT}(\{\mathbf{h}_k\}_{k=1}^K)$ and the negative samples to contrast with $\mathbf{h}_{g}$ is $Neg(\mathbf{h}_{g})=\{\widetilde{\mathbf{h}}_{j}\}_{j=1}^K$.
\newline

\noindent \textbf{Head-Tail Contrastive (HTC)} \cite{wang2021graph} is proposed to enhance the graph-level representations learned by GNNs. Given a graph $g_i=(\mathbf{A}_i, \mathbf{X}_i)\in\mathcal{G}$, it first applies an augmentation transformation $\mathcal{T}(\cdot)$ to obtain $\widetilde{g}_i=(\widetilde{\mathbf{A}}_i,\widetilde{\mathbf{X}}_i)=\mathcal{T}(\mathbf{A}_i,\mathbf{X}_i)$. Then a shared graph encoder $f_{\theta_1}(\cdot)$ are applied to $g_i$ and $\widetilde{g}_i$ to obtain node embedding matirces $\mathbf{H}_i$ and $\widetilde{\mathbf{H}}_i$, respectively. Besides, a $\text{READOUT}(\cdot)$ function is used to obtain graph-level representations $\mathbf{h}_{g_i}$ and $\widetilde{\mathbf{h}}_{\widetilde{g}_i}$. Moreover, HTC samples $S$ subsets from $\mathbf{H}_i\text{, }\{\mathbf{H}_i^{(1)},\mathbf{H}_i^{(1)},\cdots,\mathbf{H}_i^{(S)}\}$ which corresponds to subgraphs $\{g_i^{(1)},g_i^{(1)},\cdots,g_i^{(S)}\}$. Then HTC calculates a subgraph-level representation by
\begin{equation}
    \overline{\mathbf{h}}_{g_i} = f_{\theta_2}\Big(\big[\mathbf{h}_{g_i^{(1)}};\mathbf{h}_{g_i^{(2)}};\cdots;\mathbf{h}_{g_i^{(S)}}\big]\Big)
\end{equation}

\noindent where $\mathbf{h}_{g_i^{(s)}}=\operatorname{READOUT}(\mathbf{H}_i^{(s)}) (1\leq s \leq S)$ and $f_{\theta_2}(\cdot)$ is a $(S,1)$ size convolution kernel. Finally, the learning objective is defined as follows
\begin{equation}
\max _{\theta_1, \theta_2} \frac{1}{|\mathcal{G}|} \sum_{g_i\in\mathcal{G}}\mathcal{MI}\left(\mathbf{h}_{g_i}, \overline{\mathbf{h}}_{g_i}\right)
\end{equation}

\noindent where the negative samples to contrast with $\mathbf{h}_{g_i}$ is $Neg(\mathbf{h}_{g_i})=\{\mathbf{h}_{g_j}\}_{g_j\in\mathcal{G},j\neq i}\cup\widetilde{\mathbf{h}}_{\widetilde{g}_i}$.

\section*{B. Generative Methods}
In this section, we will continue with some generative methods for graph SSL. However, we will not provide detailed mathematical formulas for some \emph{relatively less important} works to avoid over-redundancy.

\subsection*{B.1 Graph Autoencoding}

\noindent \textbf{Node Attribute Masking} \cite{jin2020self}. This task is similar to Graph Completion \cite{you2020does}, but in this case it masks and reconstructs the features of multiple nodes \emph{simultaneously}, and it no longer requires that the neighboring node features used for message passing must be unmasked features. It first randomly masks (i.e., set equal to zero) the features of a node set $\mathcal{S}$. Specifically, it obtains a masked node feature matrix $\widehat{\mathbf{X}}$ where $\widehat{\mathbf{x}}_i=\mathbf{x}_i\odot\mathbf{m}_i$ for $v_i\in\mathcal{S}$, and then ask the model to reconstruct these masked features. More formally,
\begin{equation}
\mathcal{L}_{ssl}\left(\theta, \mathbf{A}, \widehat{\mathbf{X}} \right)=\frac{1}{|\mathcal{S}|}\sum_{v_i\in\mathcal{S}}\left\|f_{\theta}(\mathbf{A},\widehat{\mathbf{X}})_{v_{i}} - \mathbf{x}_i\right\|^2
\end{equation}

\noindent  \textbf{G-BERT} \cite{shang2019pre} combines the power of GNNs and BERT to performs representation learning for medication recommendation. Specifically, G-BERT uses two BERT-like self-supervised pretext tasks, \emph{self-prediction and dual-prediction}. The self-prediction takes the learned embedding as input to reconstruct the masked medical codes with the same type, while dual-prediction takes one type of embedding as input and tries to reconstruct the other type. 
\newline

\noindent \textbf{SLAPS} \cite{fatemi2021slaps} is a graph learning framework that tasks the \emph{node attribute reconstruction} as the self-supervised pretext task to infer a task-specific latent structure and then apply a graph neural networks on the inferred graph structure.
\newline

\noindent To tackle the cold-start problem for recommendation tasks, \textbf{Pretrain-Recsys} \cite{hao2021pre} pre-trains the model by simulating the cold-start scenarios. Specifically, it applies an attention-based meta aggregator to reduce the impact from the cold-start neighbors and takes the target \emph{embedding reconstruction} as the self-supervised pretext task. 
\newline

\noindent \textbf{Graph-Bert} \cite{zhang2020graph} is a novel graph neural network that pre-trains the model with the self-supervised \emph{node attribute reconstruction} and \emph{structure recovery} tasks, and then transfers the model to other downstream tasks directly or with necessary fine-tuning if any supervised label information is available.

\section*{C. Predictive Methods}
In this section, we will continue with some predictive methods for graph SSL. However, we will not provide detailed mathematical formulas for some \emph{relatively less important} works to avoid over-redundancy.

\subsection*{C.1 Context-based Prediction (CP)}

\noindent \textbf{PairwiseDistance} \cite{jin2020self}. The pretext task of PairwiseDistance aims to guide the model to preserve global topology information by predicting the shortest path length between different node pairs. More specifically, it first randomly samples a certain amount of node pairs $\mathcal{S}$ from all node pairs $\{(v_i,v_j)|v_i,v_j\in\mathcal{V}\}$ and calculates the pairwise node shortest path length $d_{i,j}=d(v_i,v_j)$ for node pairs $(v_i,v_j)\in\mathcal{S}$. Furthermore, it groups the shortest path lengths into four categories: $C_{i,j}=0, C_{i,j}=1, C_{i,j}=2$, and $C_{i,j}=3$ corresponding to $d_{i,j}=1, d_{i,j}=2, d_{i,j}=3$, and $d_{i,j}\geq3$, respectively. The learning objective can then be formulated as a multi-class classification problem, as follows
\begin{equation}
\begin{small}
\begin{aligned}
\mathcal{L}_{ssl}\left(\theta, \omega, \mathbf{A}, \mathbf{X}\right)=\frac{1}{|\mathcal{S}|}&\sum_{(v_{i}, v_{i}) \in \mathcal{S}} \ell\Big(f_w\big(|f_{\theta}(\mathbf{A},\mathbf{X})_{v_{i}}\\&-f_{\theta}(\mathbf{A},\mathbf{X})_{v_{j}}|\big), C_{i,j}\Big) 
\end{aligned}
\end{small}
\end{equation}

\noindent where $\ell(\cdot)$ denotes the cross entropy loss and $f_\omega(\cdot)$ linearly maps the input to a 1-dimension value. Compared with the task of $\text{S}^2\text{GRL}$ \cite{peng2020self}, PairwiseDistance truncates the shortest path longer than 4, mainly to avoid the excessive computational burden and to prevent very noisy ultra-long pairwise distances from dominating the optimization.
\newline

\noindent \textbf{EdgeMask} \cite{jin2020self}. Different from masking edge features, we can also mask edge connections, but instead of reconstructing the entire adjacency matrix directly, EdgeMask takes the link prediction as a pretext task. More specifically, it first masks $m$ edges $\mathcal{M} \in \mathcal{E}$ and also samples $m$ edges $\overline{\mathcal{M}} \in \{(v_i,v_j)|v_i,v_j\in\mathcal{V}$ and $(v_i,v_j)\notin\mathcal{E}\}$. Then, learning objective of EdgeMask is to predict whether there exists a link between a given node pair. More formally, 
\begin{equation}
\begin{aligned}
& \mathcal{L}_{ssl}\left(\theta, \omega,  \mathbf{A}, \mathbf{X}\right)  = \frac{1}{2m}\Big(\\ & \sum_{(v_{i}, v_{j}) \in \mathcal{M}} \ell\big(f_w(|f_{\theta}(\mathbf{A},\mathbf{X})_{v_{i}}-f_{\theta}(\mathbf{A},\mathbf{X})_{v_{j}}|), 1\big)+ \\ &  \sum_{(v_{i}, v_{j}) \in \overline{\mathcal{M}}}  \ell\big(f_w(|f_{\theta}(\mathbf{A},\mathbf{X})_{v_{i}}-f_{\theta}(\mathbf{A},\mathbf{X})_{v_{j}}|), 0\big)\Big)
\end{aligned}
\end{equation}

\noindent where $\ell(\cdot)$ denotes the cross entropy loss and $f_\omega(\cdot)$ linearly maps the input to a 1-dimension value. The EdgeMask task aims to help GNN learn more local structural information.
\newline

\noindent \textbf{TopoTER} \cite{gao2021topoter} amis to maximize the mutual information between topology transformations and node representations before and after the transformations, which can be relaxed to minimizing the cross entropy between the applied topology transformation types and its estimation from node representations. Given a graph $g=(\mathbf{A},\mathbf{X})$, it first randomly samples a subset $\mathcal{S}_1$ of $M$ connected edges and a subset $\mathcal{S}_2$ of $M$ disconnected edges. Then it randomly removes $r\cdot M$ edges from $\mathcal{S}_1$ and adds $r\cdot M$ edges to $\mathcal{S}_2$, where $r$ is the edge perturbation rate, to obatain a topology-perturbation graph $\widetilde{g}=(\widetilde{\mathbf{A}},\mathbf{X})$. Next, it applies a shared graph encoder $f_\theta(\cdot)$ to obtain their node embedding matrices $\mathbf{H}=f_\theta(\mathbf{A},\mathbf{X})$ and $\widetilde{\mathbf{H}}=f_\theta(\widetilde{\mathbf{A}},\mathbf{X})$. Finally, a mapping function $g_\omega(\cdot)$ is applied to the difference $\Delta\mathbf{H}=\mathbf{H}-\widetilde{\mathbf{H}}$ to predict transformation types $\widehat{\mathbf{Y}}_{i,j}=g_\omega(\Delta\mathbf{H}_i, \Delta\mathbf{H}_j)\in\mathbb{R}^4$ for any $e_{i,j} \in \mathcal{S}_1\bigcup\mathcal{S}_2$. Then, the learning objective of TopoTER is defined as
\begin{equation}
\mathcal{L}_{ssl}\left(\theta, \omega,  \mathbf{A}, \mathbf{X}\right)  = -\sum_{e_{i,j} \in \mathcal{S}_1\bigcup\mathcal{S}_2}\sum_{c=0}^3 \mathbf{Y}^{(c)}_{i,j}\log(\widehat{\mathbf{Y}}^{(c)}_{i,j})
\end{equation}

\noindent where $c$ denotes four transformation types, $c=0$: add an edge to a disconnected node pair in $\mathcal{S}_2$; $c=1$: reomove an edge from a connected node pair in $\mathcal{S}_1$; $c=2$: keep the connection between node pairs in $\mathcal{S}_1$; $c=3$: keep the disconnection between node pairs in $\mathcal{S}_2$. $\mathbf{Y}^{(c)}_{i,j}$ is the ground-truth binary indicator (0 or 1) for each transformation type.
\newline

\noindent \textbf{Centrality Score Ranking} \cite{hu2019pre}. Node centrality is an important metric for graphs, which measures the importance of nodes based on their structural roles in the whole graph. However, different from node property, centrality scores are not comparable among different graphs with different scales. Therefore, it resorts to rank the \emph{relative orders} between nodes and consider them as pseudo labels. Specifically, four different centrality scores $\mathcal{S}=\{$eigencentrality, betweenness, closeness, subgraph centrality$\}$ are used. For a given centrality score $s\in\mathcal{S}$ and a node pair $(v_i,v_j)$ with relative order $\mathbf{Y}_{i,j}^s=\mathbb{I}(s_i>s_j)$, a mapping function $f_{\theta^{(s)}}(\cdot)$ is applied to estimate its rank score by $\mathbf{r}^s=f_{\theta^{(s)}}(\mathbf{A},\mathbf{X})$, where $\mathbf{r}^s_{i}$ denotes the rank score of node $v_i$. The probability of estimated rank order is defined as
\begin{equation}
\widehat{\mathbf{Y}}_{i,j}^s=\frac{\exp(\mathbf{r}^s_{i}-\mathbf{r}^s_{j})}{1+\exp(\mathbf{r}^s_{i}-\mathbf{r}^s_{j})}
\end{equation}

\noindent The learning objective of this pretext task is defined as 
\begin{equation}
\begin{aligned}
\mathcal{L}_{ssl} & \left(\{\theta^{(s)}\}_{s=1}^4, \mathbf{A}, \mathbf{X}\right)= -\frac{1}{4N^2}\sum_{s\in\mathcal{S}} \\ & \sum_{i,j\in\mathcal{V}}\Big[\mathbf{Y}^s_{i,j}\log(\widehat{\mathbf{Y}}^s_{i,j})  +(1-\mathbf{Y}^s_{i,j})\log(1-\widehat{\mathbf{Y}}^s_{i,j})\Big]
\end{aligned}
\end{equation}

\noindent \textbf{ContextLabel} \cite{jin2020self}. Different from Distance2Labeled, which uses relative position as a self-supervised signal, the pretext task of ContextLabel works by constructing a \emph{local distribution} for each node and then asking the model to regress these distributions. It first assigns labels for unlabeled node with the Label Propagation (LP) algorithm \cite{zhu2003semi}, and then defines the label distribution $\mathbf{y}_i$ for node $v_i$ within its $k$-hop neighborhood, where the $c$-th element of the label distribution vector $\mathbf{y}_{i,c}$ can be defined as
\begin{equation}
\mathbf{y}_{i, c}=\frac{\left|\mathcal{N}_i^{\mathcal{V}_L}(c)\right|+\left|\mathcal{N}_i^{\mathcal{V}_U}(c)\right|}{\left|\mathcal{N}_i^{\mathcal{V}_L}\right|+\left|\mathcal{N}_i^{\mathcal{V}_{U}}\right|}, c=1,2, \cdots, C
\end{equation}

\noindent where $\mathcal{N}_i^{\mathcal{V}_L}$ and $\mathcal{N}_i^{\mathcal{V}_U}$ are the labeled nodes and unlabeled nodes within $k$-hop neighborhood of node $v_i$, respectively. $\mathcal{N}_i^{\mathcal{V}_L}(c)$ denotes only those in the neighborhood set with ground-truth label $c$, and $\mathcal{N}_i^{\mathcal{V}_U}(c)$ denotes those in the neighborhood set that are assigned label $c$ by Label Propagation (LP) algorithm. The learning objective is defined as
\begin{equation}
\mathcal{L}_{ssl}\left(\theta, \mathbf{A}, \mathbf{X}\right)  = \frac{1}{N}\sum_{v_{i} \in \mathcal{V}} \Big\|f_{\theta}(\mathbf{A},\mathbf{X})_{v_{i}}-\mathbf{y}_i\Big\|^2
\end{equation}

\noindent In addition to Label Propagation, there are numerous algorithms that can be used to compute pseudo-labels, such as the Iterative Classification Algorithm (ICA) \cite{sen2008collective}, and even a combination of LP and ICA for better performance.
\newline

\noindent Motivated by the observation that anomalous nodes differ from normal nodes in structures and attributes, \textbf{Hop-Count based Model (HCM)} propose to use \emph{global context prediction} as a self-supervised task for modeling both local and global contextual information and then achieve better anomaly detection on the attributed networks. 

\subsection*{C.2 Self-Training (ST)}
\noindent \textbf{SEF} \cite{sehanobish2020self} presents a framework that first trains a graph attention network model with \emph{pseudo labels} obtained from unsupervised Louvain clustering \cite{blondel2008fast} and then uses the learned edge attention coefficients as self-supervised edge features. Besides, it encodes the learned edge features via a Set Transformer and combines them with node features for node classification in an end-to-end training manner.

\subsection*{C.3 Domain Knowledge-based Prediction (DK)}
\noindent \textbf{DrRepair} \cite{yasunaga2020graph} considers the problem of learning to repair programs from diagnostic feedback by building a program feedback graph, which connects symbols relevant to program repair in source code and diagnostic feedback. Besides, it leverages unlabeled programs available online to create a large amount of \emph{extra program repair examples}, which are used to pre-train a graph neural network for program repair.

\section*{D. Summary of Surveyed Works}
A summary of all the surveyed works (\emph{a total of 71 methods}) is presented in Table.~\ref{tabA1:methods_1} and Table.~\ref{tabA1:methods_2}, including graph property, pretext task type, augmentation strategy, objective function, training strategy, and the year of publication.

\section*{E. Summary of Implementation Details}
The implementation details of all the surveyed works are presented in Table.~\ref{tabA2:task_1} and Table.~\ref{tabA2:task_2}, such as the (node/link/graph) level of downstream tasks, evaluation metrics with specific tasks, and the commonly used datasets.

\section*{F. Summary of Common Datasets}
A summary and statistics of common graph datasets is shown in Table.~\ref{tabA3:dataset}, including category, graph number, node number \textit{per graph}, edge number \textit{per graph}, dimensionality of node attributes, class number and citation papers.

\section*{G. Summary of Open-source Codes}
A summary of the open-source codes of all the surveyed works is presented in Table.~\ref{tabA4:github}, where we provide hyperlinks to their open-source codes, and those for which no open-source code is found are indicated by ``N.A.". Moreover, we have created a GitHub repository \url{https://github.com/LirongWu/awesome-graph-self-supervised-learning} to summarize the latest advances in graph SSL, which will be updated in real-time as more papers and their codes become available.

\begin{table*}[bp]
\caption{A summary of the surveyed papers. The summary is organized based on the underlying motivation behind pretext task design.}
\label{tabA1:methods_1}
\centering
\begin{tabular}{lcccccc}

\hline
\multicolumn{1}{c}{Methods} & \multicolumn{1}{c}{Graph Property} & \multicolumn{1}{c}{Pretext Task} & \multicolumn{1}{c}{Data Augmentation} & \multicolumn{1}{c}{Objective Function} &\multicolumn{1}{c}{Training Strategy} & \multicolumn{1}{c}{Year} \\ 
\hline

Graph Completion \cite{you2020does} & Attributed & Generative/AE & Attribute Masking & MAE & P\&F/JL & 2020 \\ \hline
\tabincell{l}{Node Attribute \\ Masking \cite{jin2020self}} & Attributed & Generative/AE & Attribute Masking & MAE & P\&F/JL & 2020 \\ \hline
\tabincell{l}{Edge Attribute \\ Masking \cite{hu2019strategies}} & Attributed & Generative/AE & Attribute Masking & MAE & P\&F & 2019 \\ \hline
\tabincell{l}{Node Attribute \\ Denoising \cite{manessi2020graph}}  & Attributed & Generative/AE & Attribute Masking & MAE & JL & 2020 \\ \hline
\tabincell{l}{Adjacency Matrix \\ Reconstruction \cite{zhu2020self}} & Attributed & Generative/AE & \tabincell{c}{Attribute Masking \\ Edge Perturbation} & MAE & JL & 2020 \\ \hline
Graph Bert \cite{zhang2020graph} & Attributed & Generative/AE & \tabincell{c}{Attribute Masking \\ Edge Perturbation} & MAE & P\&F & 2020 \\ \hline
Pretrain-Recsys \cite{hao2021pre} & Attributed & Generative/AE & Edge Perturbation & MAE & P\&F & 2021 \\ \hline
GPT-GNN \cite{hu2020gpt} & Heterogeneous & Generative/AR & \tabincell{c}{Attribute Masking \\ Edge Perturbation} & MAE/InfoNCE & P\&F & 2020 \\ \hline

GraphCL \cite{you2020graph} & Attributed & Contrastive/G-G & \tabincell{c}{Attribute Masking \\ Edge Perturbation \\ Random Walk Sampling} & InfoNCE & URL & 2020 \\ \hline
IGSD \cite{zhang2020iterative} & Attributed & Contrastive/G-G & \tabincell{c}{Edge Perturbation \\ Edge Doffisopm} & InfoNCE & JL/URL & 2020 \\ \hline
DACL \cite{verma2020towards} & Attributed & Contrastive/G-G & Mixup & InfoNCE & URL & 2020 \\ \hline
LCC \cite{ren2021label} & Attributed & Contrastive/G-G & None & InfoNCE & JL & 2021 \\ \hline
CSSL \cite{zeng2020contrastive} & Attributed & Contrastive/G-G & \tabincell{c}{Node Insertion \\ Edge Perturbation \\ Uniform Sampling} & InfoNCE & P\&F/JL/URL & 2020 \\ \hline
GCC \cite{qiu2020gcc} & Unattributed & Contrastive/C-C & Random Walk Sampling & InfoNCE & P\&F/URL & 2020 \\ \hline

GRACE \cite{zhu2020deep} & Attributed & Contrastive/L-L & \tabincell{c}{Attribute Masking \\ Edge Perturbation} & InfoNCE & URL & 2020 \\ \hline
GCA \cite{zhu2020graph} & Attributed & Contrastive/L-L & Attention-based & InfoNCE & URL & 2020 \\ \hline
GROC \cite{jovanovic2021towards} & Attributed & Contrastive/L-L & Gradient-based & InfoNCE & URL & 2021 \\ \hline
SEPT \cite{yu2021socially} & Attributed & Contrastive/L-L & Edge Perturbation & InfoNCE & JL & 2021 \\ \hline
STDGI \cite{opolka2019spatio} & Spatial-Temporal & Contrastive/L-L & Attribute Shuffling & JS Estimator & URL & 2019 \\ \hline
GMI \cite{peng2020graph} & Attributed & Contrastive/L-L & None & SP Estimator & URL & 2020 \\ \hline
KS2L \cite{yu2020self} & Attributed & Contrastive/L-L & None & InfoNCE & URL & 2020 \\ \hline
$\text{C}\text{G}^3$ \cite{wan2020contrastive} & Attributed & Contrastive/L-L & None & InfoNCE & JL & 2020 \\ \hline
BGRL \cite{thakoor2021bootstrapped} & Attributed & Contrastive/L-L & \tabincell{c}{Attribute Masking \\ Edge Perturbation} & Inner Product & URL & 2021 \\ \hline

\end{tabular}
\end{table*}

\begin{table*}[!htbp]
\caption{A summary of the surveyed papers. The summary is organized based on the underlying motivation behind pretext task design.}
\label{tabA1:methods_2}
\centering
\begin{tabular}{lcccccc}

\hline
\multicolumn{1}{c}{Methods} & \multicolumn{1}{c}{Graph Property} & \multicolumn{1}{c}{Pretext Task} & \multicolumn{1}{c}{Data Augmentation} & \multicolumn{1}{c}{Objective Function} &\multicolumn{1}{c}{Training Strategy} & \multicolumn{1}{c}{Year} \\ 
\hline

SelfGNN \cite{kefato2021self} & Attributed & Contrastive/L-L & \tabincell{c}{Attribute Masking \\ Edge Diffusion} & MSE & URL & 2021 \\ \hline
HeCo \cite{wang2021self} & Heterogeneous & Contrastive/L-L & None & InfoNCE & URL & 2021 \\ \hline
PT-DGNN \cite{zhang2021pre} & Dynamic & Contrastive/L-L & \tabincell{l}{Attribute Masking \\ Edge Perturbation} & InforNCE  & P\&F & 2021 \\ \hline
COAD \cite{chen2020coad} & Attributed & Contrastive/L-L & None & Triplet Margin Loss & P\&F & 2020 \\ \hline
Contrst-Reg \cite{ma2021improving}& Attributed & Contrastive/L-L & Attribute Shuffling & InfoNCE & JL & 2021 \\ \hline

DGI \cite{velickovic2019deep} & Attributed & Contrastive/L-G & Arbitrary & JS Estimator & URL & 2019 \\ \hline
HDMI \cite{jing2021hdmi} & Attributed & Contrastive/L-G & Attribute Shuffling & JS Estimator & URL & 2021 \\ \hline
DMGI \cite{park2020unsupervised} & Heterogeneous & Contrastive/L-G & Attribute Shuffling & \tabincell{c}{JS Estimator \\ MAE} & URL & 2020 \\ \hline
MVGRL \cite{hassani2020contrastive} & Attributed & Contrastive/L-G & \tabincell{c}{Attribute Masking \\ Edge Perturbation \\ Edge Diffusion \\ Random Walk Sampling} & \tabincell{c}{DV Estimator \\ JS Estimator \\ NT-Xent \\ InfoNCE} & URL & 2020 \\ \hline
HDGI \cite{ren2019heterogeneous} & Heterogeneous & Contrastive/L-G & Attribute Shuffling & JS Estimator & URL & 2019 \\ \hline
Subg-Con \cite{jiao2020sub} & Attributed & Contrastive/L-C & Importance Sampling & Triplet Margin Loss & URL & 2020 \\ \hline
Cotext Prediction \cite{hu2019strategies} & Attributed & Contrastive/L-C & Ego-nets Sampling & Cross Entropy & P\&F & 2019 \\ \hline
GIC \cite{mavromatis2020graph} & Attributed & Contrastive/L-C & Arbitrary & JS Estimator & URL & 2020 \\ \hline
GraphLoG \cite{xu2021selfsupervised} & Attributed & \tabincell{l}{Contrastive/L-L \\ Contrastive/G-G \\ Contrastive/L-C} & Attribute Masking & InfoNCE & URL & 2021 \\ \hline
MHCN \cite{yu2021self} & Heterogeneous & Contrastive/L-C & Attribute Shuffling & InfoNCE & JL & 2021 \\ \hline
EGI \cite{zhu2020transfer} & Attributed & Contrastive/L-C & Ego-nets Sampling & SP Estimator & P\&F & 2020 \\ \hline
MICRO-Graph \cite{zhang2020motif} & Attributed & Contrastive/C-G & Knowledge Sampling & InfoNCE & URL & 2020 \\ \hline
InfoGraph \cite{sun2019infograph} & Attributed & Contrastive/C-G & None & SP Estimator & URL & 2019 \\ \hline
SUGAR \cite{sun2021sugar} & Attributed & Contrastive/C-G & BFS Sampling & JS Estimator & JL & 2021 \\ \hline
BiGI \cite{cao2021bipartite} & Heterogeneous & Contrastive/C-G & \tabincell{c}{Edge Perturbation \\ Ego-nets Sampling} & JS Estimator & JL & 2021 \\ \hline
HTC \cite{wang2021graph} & Attributed & Contrastive/C-G & Attribute Shuffling & \tabincell{l}{SP Estimator \\ DV Estimator}  & URL & 2021 \\ \hline

\tabincell{l}{Node Property \\ Prediction \cite{jin2020self}} & Attributed & Predictive/NP & None & MAE & P\&F/JL & 2020 \\ \hline
\tabincell{l}{$\text{S}^2\text{GRL}$ \cite{peng2020self}} & Attributed & Predictive/CP & None & Cross Entropy & URL & 2020 \\ \hline
PairwiseDistance \cite{jin2020self} & Attributed & Predictive/CP & None & Cross Entropy & P\&F/JL & 2020 \\ \hline
PairwiseAttrSim \cite{jin2020self} & Attributed & Predictive/CP & None & MAE & P\&F/JL & 2020 \\ \hline
Distance2Cluster \cite{jin2020self} & Attributed & Predictive/CP & None & MAE & P\&F/JL & 2020 \\ \hline
EdgeMask \cite{jin2020self} & Attributed & Predictive/CP & None & Cross Entropy & P\&F/JL & 2020 \\ \hline
TopoTER \cite{gao2021topoter} & Attributed & Predictive/CP & Edge Perturbation & Cross Entropy & URL & 2021 \\ \hline
\tabincell{l}{Centrality Score \\ Ranking \cite{hu2019pre}} & Attributed & Predictive/CP & None & Cross Entropy & P\&F & 2019 \\ \hline
Meta-path Prediction \cite{hwang2020self} & Heterogeneous & Predictive/CP & None & Cross Entropy & JL & 2020 \\ \hline
SLiCE \cite{wang2020self} & Heterogeneous & Predictive/CP & None & Cross Entropy & P\&F & 2020 \\ \hline
Distance2Labeled \cite{jin2020self} & Attributed & Predictive/CP & None & MAE & P\&F/JL & 2020 \\ \hline
ContextLabel \cite{jin2020self} & Attributed & Predictive/CP & None & MAE & P\&F/JL & 2020 \\ \hline
HCM \cite{huang2021hop} & Attributed & Predictive/CP & Edge Perturbation & Bayesian Inference & URL & 2021 \\ \hline

\tabincell{l}{Contextual Molecular \\ Property Prediction \cite{rong2020self}} & Attributed & Predictive/DK & None & Cross Entropy & P\&F & 2020 \\ \hline
\tabincell{l}{Graph-level \\ Motif Prediction \cite{rong2020self}} & Attributed & Predictive/DK & None & Cross Entropy & P\&F & 2020 \\ \hline
\tabincell{l}{Multi-stage \\ Self-training \cite{li2018deeper}} & Attributed & Predictive/ST & None & None & JL & 2018 \\ \hline
Node Clustering \cite{you2020does} & Attributed & Predictive/ST & None & Clustering & P\&F/JL & 2020 \\ \hline
Graph Partitioning \cite{you2020does} & Attributed & Predictive/ST & None & Graph Partitioning & P\&F/JL & 2020 \\ \hline
CAGNN \cite{zhu2020cagnn} & Attributed & Predictive/ST & None & Clustering & URL & 2020 \\ \hline
M3S \cite{sun2020multi} & Attributed & Predictive/ST & None & Clustering & JL & 2020 \\ \hline
Cluster Preserving \cite{hu2019pre} & Attributed & Predictive/ST & None & Cross Entropy & P\&F & 2019 \\ \hline

\end{tabular}
\end{table*}

\section*{H. Experimental Study Results}
\textcolor{mark}{Two downstream tasks, namely node and graph classification, and one evaluation metric, namely accuracy, are selected to conduct the experimental study. For node classification, we provide the classification performance of 23 classical methods on 8 commonly used datasets; for graph classification, we select 7 datasets and report the comparison results of 11 representative methods. The results of node classification and graph classification are shown in Table.~\ref{tabA5:node} and Table.~\ref{tabA6:graph}, respectively, and the best result in each dataset is marked in \textbf{bold}. The metrics are taken directly from their original papers or retrained with the open-source code. When the code is not publicly available, or when it is impractical to run the released code from scratch, we replace the corresponding results with a dash ``-". Moreover, fixed dataset splits are adopted for all methods to make a fair comparison (1) the data splits of Cora, Citeseer, and Pubmed are consistent with that in \cite{kipf2016semi}, (2) the data splits of Wiki-CS, Amazon-Computers, Amazon-Photo, Coauthor-CS, and Coauthor-Physics are consistent with that in \cite{thakoor2021bootstrapped}, (3) the data splits of MUTAG, PTC, RDT-B, RDT-M5K, IMDB-B, and IMDB-M are consistent with that in \cite{sun2019infograph}, and (4) the data splits of NCI1 are consistent with that in \cite{you2020graph}.}

\textcolor{mark}{After analyzing the results in Table.~\ref{tabA5:node} and Table.~\ref{tabA6:graph}, we have the following observations: (1) There is no single method that can achieve the best performance on all datasets, and the state-of-the-art method on one dataset may be suboptimal on another. This suggests that existing graph self-supervised algorithms are somewhat dependent on the properties of the graph and are not sufficiently general. (2) Algorithms that achieve excellent performance on small-scale datasets (e.g., Cora and Citeseer), such as Graph-Bert \cite{zhang2020graph} and Sub-Con \cite{jiao2020sub}, may have suboptimal performance on large-scale datasets (e.g., Coauthor-CS and Coauthor-Physics). This inspires follow-up researchers to evaluate model performance on datasets with different scales, as experimental results on small-scale datasets may be biased and therefore less reliable. (3) Most generative and predictive learning methods are specifically designed for node-level tasks, such as node classification, but nevertheless, they are still not comparable to those state-of-the-art contrastive methods. This suggests that the potential of generative and predictive learning needs to be further explored, and that domain knowledge-based learning may be a promising direction. (4) Due to the lack of comparison with contemporaneous work, many state-of-the-art methods achieve roughly comparable performance on some datasets, such as Amazon-Computers and Coauthor-Physics, with differences smaller than one percent. The purpose of this experimental study is to provide a good foundation for follow-up researchers and to facilitate a fairer experimental comparison for graph self-supervised learning.}

\section*{I. TimeLine}
\begin{figure*}[!htbp]
	\begin{center}
		\includegraphics[width=0.9\linewidth]{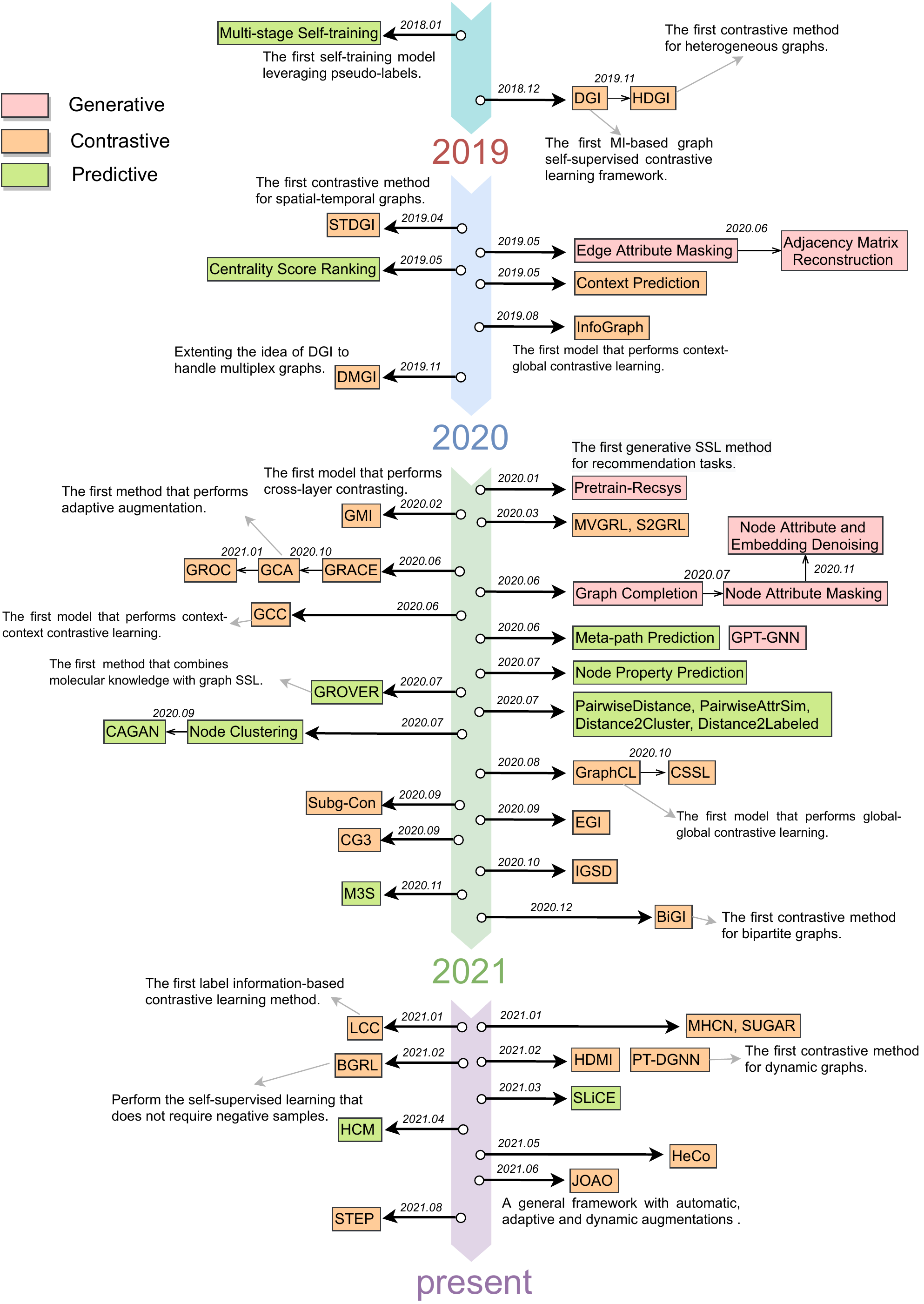}
	\end{center}
	\caption{\textcolor{mark}{Key milestones in the development of graph self-supervised learning. The generative, contrastive, and predictive learning methods are marked with red, orange, and green blocks, respectively. Best viewed in color.}}
	\label{fig:A1}
\end{figure*}

\textcolor{mark}{A complete timeline is provided in Fig.~\ref{fig:A1} to present a clear development lineage of various graph SSL methods, listing the publication dates (based on the arxiv) of key milestones. Besides, we provide inheritance connections between methods to show how they are developed. More importantly, we provide short descriptions of contributions for some pioneering methods to highlight their importance.}

\begin{table*}[!htbp]
\caption{A summary of implementation details about the surveyed papers, including downstream task level, evaluation metric, and dataset.}
\label{tabA2:task_1}
\centering
\begin{tabular}{lcll}

\hline
\multicolumn{1}{c}{Methods} & \multicolumn{1}{c}{Task Level} &
\multicolumn{1}{c}{Evaluation Metric} & \multicolumn{1}{c}{Dataset} \\ 
\hline

Graph Completion \cite{you2020does} & Node & Node Classification (Acc) & Cora, Citeseer, Pubmed \\ \hline
\tabincell{l}{Node Attribute \\ Masking \cite{jin2020self}} & Node & Node Classification (Acc) & Cora, Citeseer, Pubmed, Reddit \\ \hline
\tabincell{l}{Edge Attribute \\ Masking \cite{hu2019strategies}} & Graph & \tabincell{l}{Graph Classification (ROC-AUC)} & \tabincell{l}{MUTAG, PTC, PPI, BBBP, Tox21, ToxCast, ClinTox, MUV, \\ HIV, SIDER, BACE} \\ \hline
\tabincell{l}{Node Attribute \\ Denoising \cite{manessi2020graph}} & Node & Node Classification (Acc) & Cora, Citeseer, Pubmed \\ \hline
\tabincell{l}{Adjacency Matrix \\ Reconstruction \cite{zhu2020self}} & Node & Node Classification (Acc) & Cora, Citeseer, Pubmed \\ \hline
Graph Bert \cite{zhang2020graph} & Node & \tabincell{l}{Node Classification (Acc), \\ Node Clustering (NMI)} & Cora, Citeseer, Pubmed \\ \hline
Pretrain-Recsys \cite{hao2021pre} & Node/Link & - & ML-1M, MOOCs, Last-FM \\ \hline
GPT-GNN \cite{hu2020gpt} & Node/Link & \tabincell{l}{Node Classification (F1-score), \\ Link Prediction (ROC-AUC)} & OAG, Amazon, Reddit \\ \hline

GraphCL \cite{you2020graph} & Graph & \tabincell{l}{Graph Classification \\ (Acc, ROC-AUC)} & \tabincell{l}{NCI1, PROTEINS, D\&D, COLLAB, RDT-B, RDT-M5K, \\ GITHUB, MNIST, CIFAR10, MUTAG, IMDB-B, BBBP, \\ Tox21, ToxCast, SIDER, ClinTox, MUV, HIV, BACE, PPI} \\ \hline
IGSD \cite{zhang2020iterative} & Graph & Graph Classification (Acc) & \tabincell{l}{MUTAG, PTC, NCI1, IMDB-B, QM9, COLLAB, IMDB-M} \\ \hline
DACL \cite{verma2020towards} & Graph & Graph Classification (Acc) & \tabincell{l}{MUTAG, PTC, IMDB-B, IMDB-M, RDT-B, RDT-M5K} \\ \hline
LCC \cite{ren2021label} & Graph & Graph Classification (Acc) & \tabincell{l}{IMDB-B, IMDB-M, COLLAB, MUTAG, PROTEINS, PTC, \\ NCI1, D\&D} \\ \hline
CSSL \cite{zeng2020contrastive} & Graph & Graph Classification (Acc) & PROTEINS, D\&D, NCI1, NCI109, Mutagenicity \\ \hline
GCC \cite{qiu2020gcc} & Node/Graph & \tabincell{l}{Node Classification (Acc), \\ Graph Classification (Acc)} & \tabincell{l}{US-Airport, H-index, COLLAB, IMDB-B, IMDB-M, RDT-B, \\ RDT-M5K} \\ \hline
GRACE \cite{zhu2020deep} & Node & \tabincell{l}{Node Classification \\ (Acc, Micro-F1)} & Cora, Citeseer, Pubmed, DBLP, Reddit, PPI \\ \hline
GCA \cite{zhu2020graph} & Node & Node Classification (Acc) & \tabincell{l}{Amazon-Computers, Amazon-Photo, Coauthor-Physics, \\ Coauthor-CS, Wiki-CS} \\ \hline
GROC \cite{jovanovic2021towards} & Node & Node Classification (Acc) &  Cora, Citeseer, Pubmed, Amazon-Photo, Wiki-CS \\ \hline
SEPT \cite{yu2021socially} & Node/Link & - &  Last-FM, Douban, Yelp \\ \hline
STDGI \cite{opolka2019spatio} & Node & \tabincell{l}{Node Regression \\ (MAE, RMSE, MAPE)} & METR-LA \\ \hline
GMI \cite{peng2020graph} & Node/Link & \tabincell{l}{Node Classification \\ (Acc, Micro-F1), \\ Link Prediction (ROC-AUC)} & Cora, Citeseer, PubMed, Reddit, PPI, BlogCatalog, Flickr \\ \hline
KS2L \cite{yu2020self} & Node/Link & \tabincell{l}{Node Classification (Acc), \\ Link Prediction (ROC-AUC)} & \tabincell{l}{Cora, Citeseer, Pubmed, Amazon-Computers, \\ Amazon-Photo, Coauthor-CS} \\ \hline
$\text{C}\text{G}^3$ \cite{wan2020contrastive} & Node & Node Classification (Acc) & \tabincell{l}{Cora, Citeseer, Pubmed, Amazon-Computers, \\ Amazon-Photo, Coauthor-CS} \\ \hline
BGRL \cite{thakoor2021bootstrapped} & Node & \tabincell{l}{Node Classification \\ (Acc, Micro-F1)} & \tabincell{l}{Wiki-CS, Amazon-Computers, Amazon-Photo, PPI, \\ Coauthor-CS, Coauthor-Physics, ogbn-arxiv} \\ \hline
SelfGNN \cite{kefato2021self} & Node & \tabincell{l}{Node Classification (Acc)} & \tabincell{l}{Cora, Citeseer, Pubmed, Amazon-Computers, \\ Amazon-Photo, Coauthor-CS, Coauthor-Physics} \\ \hline
HeCo \cite{kefato2021self} & Node & \tabincell{l}{Node Classification (ROC-AUC, \\  Micro-F1, Macro-F1), \\ Node Clustering (NMI, ARI)} & ACM, DBLP, Freebase, AMiner \\ \hline
PT-DGNN \cite{zhang2021pre} & Link & Link Prediction (ROC-AUC) & HepPh, Math Overflow, Super User \\ \hline
COAD \cite{chen2020coad} & Node/Link & \tabincell{l}{Node Clustering \\ (Precision, Recall, F1-score), \\ Link Prediction \\ (HitRatio@K, MRR)} & \tabincell{l}{AMiner, News, LinkedIn} \\ \hline
Contrast-Reg \cite{ma2021improving} & Node/Link & \tabincell{l}{Node Classification (Acc), \\ Node Clustering \\ (NMI, Acc, Macro-F1), \\ Link Prediction (ROC-AUC)} & \tabincell{l}{Cora, Citeseer, Pubmed, Reddit, ogbn-arxiv,  Wikipedia, \\ ogbn-products, Amazo-Computers, Amazo-Photo} \\ \hline

DGI \cite{velickovic2019deep} & Node & \tabincell{l}{Node Classification \\ (Acc, Micro-F1)} & Cora, Citeseer, Pubmed, Reddit, PPI \\ \hline
HDMI \cite{jing2021hdmi} & Node & \tabincell{l}{Node Classification \\ (Micro-F1, Macro-F1), \\ Node Clustering (NMI)} & ACM, IMDB, DBLP, Amazon \\ \hline
DMGI \cite{park2020unsupervised} & Node & \tabincell{l}{Node Clustering (NMI), \\ Node Classification (Acc)} & ACM, IMDB, DBLP, Amazon \\ \hline
MVGRL \cite{hassani2020contrastive} & Node/Graph & \tabincell{l}{Node Classification (Acc), \\ Node Clustering (NMI, ARI), \\ Graph Classification (Acc)} & \tabincell{l}{Cora, Citeseer, Pubmed, MUTAG, PTC, IMDB-B, \\ IMDB-M, RDT-B} \\ \hline

\end{tabular}
\end{table*}

\begin{table*}[!htbp]
\caption{A summary of implementation details about the surveyed papers, including downstream task level, evaluation metric, and dataset.}
\label{tabA2:task_2}
\centering
\begin{tabular}{lcll}

\hline
\multicolumn{1}{c}{Methods} & \multicolumn{1}{c}{Task Level} &
\multicolumn{1}{c}{Evaluation Metric} & \multicolumn{1}{c}{Dataset} \\ 
\hline

HDGI \cite{ren2019heterogeneous} & Node & \tabincell{l}{Node Classification \\ (Micro-F1, Macro-F1), \\ Node Clustering (NMI, ARI)} & \tabincell{l}{ACM, DBLP, IMDB} \\ \hline
Subg-Con \cite{jiao2020sub} & Node & \tabincell{l}{Node Classification \\ (Acc, Micro-F1)} & Cora, Citeseer, Pubmed, PPI, Flickr, Reddit \\ \hline
Cotext Prediction \cite{hu2019strategies} & Graph & \tabincell{l}{Graph Classification (ROC-AUC)} & \tabincell{l}{MUTAG, PTC, PPI, BBBP, Tox21, ToxCast, ClinTox, \\ MUV, HIV, SIDER, BACE} \\ \hline
GIC \cite{mavromatis2020graph} & Node/Link & \tabincell{l}{Node Classification (Acc), \\ Node Clustering \\ (Acc, NMI, ARI), \\ Link Prediction \\ (ROC-AUC, ROC-AP)} & \tabincell{l}{Cora, Citeseer, Pubmed, Amazon-Computers, \\ Amazon-Photo, Coauthor-CS, Coauthor-Physics} \\ \hline
GraphLoG \cite{xu2021selfsupervised} & Graph & \tabincell{l}{Graph Classification (ROC-AUC)} & \tabincell{l}{BBBP, Tox21, ToxCast, ClinTox, MUV, HIV, SIDER, \\ BACE} \\ \hline
MHCN \cite{yu2021self} & Node/Link & - & Last-FM, Douban, Yelp \\ \hline
EGI \cite{zhu2020transfer} & Node/Link & \tabincell{l}{Node Classification (Acc), \\ Link Prediction \\ (ROC-AUC, MRR)} & YAGO, Airport \\ \hline
MICRO-Graph \cite{zhang2020motif} & Graph & \tabincell{l}{Graph Classification (ROC-AUC)} & \tabincell{l}{BBBP, Tox21, ToxCast, ClinTox, HIV, SIDER, BACE} \\ \hline
InfoGraph \cite{sun2019infograph} & Graph & Graph Classification (Acc) & \tabincell{l}{MUTAG, PTC, RDT-B, RDT-M5K, IMDB-B, QM9, \\ IMDB-M} \\ \hline
SUGAR \cite{sun2021sugar} & Graph & Graph Classification (Acc) & MUTAG, PTC, PROTEINS, D\&D, NCI1, NCI109 \\ \hline
BiGI \cite{cao2021bipartite} & Link & \tabincell{l}{Link Prediction \\ (AUC-ROC, AUC-PR)} & \tabincell{l}{DBLP, ML-100K, ML-1M, Wikipedia} \\ \hline
HTC \cite{wang2021graph} & Graph & Graph Classification (Acc) & \tabincell{l}{MUTAG, PTC, IMDB-B, IMDB-M, RDT-B, QM9, \\ RDT-M5K} \\ \hline

\tabincell{l}{Node Property \\ Prediction \cite{jin2020self}} & Node & Node Classification (Acc) & Cora, Citeseer, Pubmed \\ \hline
\tabincell{l}{$\text{S}^2\text{GRL}$ \cite{peng2020self}} & Node/Link & \tabincell{l}{Node Classification \\ (Acc, Micro-F1), \\ Node Clustering (NMI), \\ Link Prediction (ROC-AUC)} & \tabincell{l}{Cora, Citeseer, Pubmed, PPI, Flickr, BlogCatalog, \\ Reddit} \\ \hline
PairwiseDistance \cite{jin2020self} & Node & Node Classification (Acc) & Cora, Citeseer, Pubmed \\ \hline
PairwiseAttrSim \cite{jin2020self} & Node & Node Classification (Acc) & Cora, Citeseer, Pubmed \\ \hline
Distance2Cluster \cite{jin2020self} & Node & Node Classification (Acc) & Cora, Citeseer, Pubmed \\ \hline
EdgeMask \cite{jin2020self} & Node & Node Classification (Acc) & Cora, Citeseer, Pubmed \\ \hline
TopoTER \cite{gao2021topoter} & Node/Graph & \tabincell{l}{Node Classification (Acc), \\ Graph Classification (Acc)} & \tabincell{l}{Cora, Citeseer, Pubmed, MUTAG, PTC, RDT-B, \\ RDT-M5K, IMDB-B, IMDB-M} \\ \hline
\tabincell{l}{Centrality Score \\ Ranking \cite{hu2019pre}} & Node/Link/Graph & \tabincell{l}{Node Classification (Micro-F1), \\ Link Prediction (Micro-F1), \\ Graph Classification (Micro-F1)} & Cora, Pubmed, ML-100K, ML-1M, IMDB-M, IMDB-B \\ \hline
Meta-path Prediction \cite{hwang2020self} & Node/Link & \tabincell{l}{Node Classification (F1-score), \\ Link Prediction (ROC-AUC)} & ACM, IMDB, Last-FM, Book-Crossing \\ \hline
SLiCE \cite{wang2020self} & Link & \tabincell{l}{Link Prediction \\ (ROC-AUC, Micro-F1)} & Amazon, DBLP, Freebase, Twitter, Healthcare \\ \hline
Distance2Labeled \cite{jin2020self} & Node & Node Classification (Acc) & Cora, Citeseer, Pubmed \\ \hline
ContextLabel \cite{jin2020self} & Node & Node Classification (Acc) & Cora, Citeseer, Pubmed, Reddit \\ \hline
HCM \cite{huang2021hop} & Node & Node Classification (ROC-AUC) & ACM, Amazon, Enron, BlogCatalog, Flickr \\ \hline

\tabincell{l}{Contextual Molecular \\ Property Prediction \cite{rong2020self}} & Graph & \tabincell{l}{Graph Classification (Acc), \\ Graph Regression (MAE)} & \tabincell{l}{BBBP, SIDER, ClinTox, BACE, Tox21, ToxCast, ESOL, \\ FreeSolv, Lipo, QM7, QM8} \\ \hline
\tabincell{l}{Graph-level \\ Motif Prediction \cite{rong2020self}} & Graph & \tabincell{l}{Graph Classification (Acc), \\ Graph Regression (MAE)} & \tabincell{l}{BBBP, SIDER, ClinTox, BACE, Tox21, ToxCast, ESOL, \\ FreeSolv, Lipo, QM7, QM8} \\ \hline
\tabincell{l}{Multi-stage \\ Self-training \cite{li2018deeper}} & Node & Node Classification (Acc) & Cora, Citeseer, Pubmed \\ \hline
Node Clustering \cite{you2020does} & Node & Node Classification (Acc) & Cora, Citeseer, Pubmed \\ \hline
Graph Partitioning \cite{you2020does} & Node & Node Classification (Acc) & Cora, Citeseer, Pubmed \\ \hline
CAGNN \cite{zhu2020cagnn} & Node & \tabincell{l}{Node Classfication \\ (Micro-F1, Macro-F1), \\ Node Clustering \\ (Micro-F1, Macro-F1, NMI)} & Cora, Citeseer, Pubmed \\ \hline
M3S \cite{sun2020multi} & Node & Node Classification (Acc) & Cora, Citeseer, Pubmed \\ \hline
Cluster Preserving \cite{hu2019pre} & Node/Link/Graph & \tabincell{l}{Node Classification (Micro-F1), \\ Link Prediction (Micro-F1), \\ Graph Classification (Micro-F1)} & Cora, Pubmed, ML-100K, ML-1M, IMDB-M, IMDB-B \\ \hline

\end{tabular}
\end{table*}
\begin{table*}[!htbp]
\caption{A summary and statistics of common graph datasets for graph self-supervised learning.}
\label{tabA3:dataset}
\centering
\begin{tabular}{llllllll}

\hline
\multicolumn{1}{c}{Dataset} & \multicolumn{1}{c}{Category} & \multicolumn{1}{c}{\#Graph} & \multicolumn{1}{c}{\#Node (Avg.)} & \multicolumn{1}{c}{\#Edge (Avg.)} & \multicolumn{1}{c}{\#Feature} & \multicolumn{1}{c}{\#Class} & \multicolumn{1}{c}{Citation} \\ 
\hline

Cora \cite{sen2008collective} & Citation Network & 1 & 2708 & 5429 & 1433 & 7 & \tabincell{l}{\cite{you2020does}, \cite{jin2020self}, \cite{manessi2020graph}, \cite{zhu2020self}, \cite{zhang2020graph}, \cite{zhu2020deep}, \\ \cite{jovanovic2021towards}, \cite{peng2020graph}, \cite{wan2020contrastive}, \cite{velickovic2019deep}, \cite{hassani2020contrastive}, \cite{jiao2020sub}, \\ \cite{mavromatis2020graph}, \cite{peng2020self}, \cite{hu2019pre}, \cite{li2018deeper}, \cite{zhu2020cagnn}, \cite{sun2020multi}, \\ \cite{ma2021improving}, \cite{kefato2021self}, \cite{yu2020self}, \cite{gao2021topoter}} \\ \hline
Citeseer \cite{giles1998citeseer} & Citation Network & 1 & 3327 & 4732 & 3703 & 6 & \tabincell{l}{\cite{you2020does}, \cite{jin2020self}, \cite{manessi2020graph}, \cite{zhu2020self}, \cite{zhang2020graph}, \cite{zhu2020deep}, \\ \cite{jovanovic2021towards}, \cite{peng2020graph}, \cite{wan2020contrastive}, \cite{velickovic2019deep}, \cite{hassani2020contrastive}, \cite{jiao2020sub}, \\ \cite{mavromatis2020graph}, \cite{peng2020self}, \cite{li2018deeper}, \cite{zhu2020cagnn}, \cite{sun2020multi}, \cite{ma2021improving}, \\ \cite{kefato2021self}, \cite{yu2020self}, \cite{gao2021topoter}} \\ \hline
Pubmed \cite{mccallum2000automating} & Citation Network & 1 & 19717 & 44338 & 500 & 3 & \tabincell{l}{\cite{you2020does}, \cite{jin2020self}, \cite{manessi2020graph}, \cite{zhu2020self}, \cite{zhang2020graph}, \cite{zhu2020deep}, \\ \cite{jovanovic2021towards}, \cite{peng2020graph}, \cite{wan2020contrastive}, \cite{velickovic2019deep}, \cite{hassani2020contrastive}, \cite{jiao2020sub}, \\ \cite{mavromatis2020graph}, \cite{peng2020self}, \cite{hu2019pre}, \cite{li2018deeper}, \cite{zhu2020cagnn}, \cite{sun2020multi}, \\ \cite{ma2021improving}, \cite{kefato2021self}, \cite{yu2020self}, \cite{gao2021topoter}} \\ \hline
Wiki-CS \cite{mernyei2020wiki} & Citation Network & 1 & 11701 & 216123 & 300 & 10 & \cite{zhu2020graph}, \cite{jovanovic2021towards}, \cite{thakoor2021bootstrapped} \\ \hline
Coauthor-CS \cite{shchur2018pitfalls} & Citation Network & 1 & 18333 & 81894 & 6805 & 15 & \cite{zhu2020graph}, \cite{wan2020contrastive}, \cite{mavromatis2020graph}, \cite{thakoor2021bootstrapped}, \cite{kefato2021self}, \cite{yu2020self} \\ \hline
Coauthor-Physics \cite{shchur2018pitfalls} & Citation Network & 1 & 34493 & 247962 & 8415 & 5 & \cite{zhu2020graph}, \cite{mavromatis2020graph}, \cite{thakoor2021bootstrapped}, \cite{kefato2021self} \\ \hline
DBLP (v12) & Citation Network  & 1 & 4894081 & 45564149 & - & - & \tabincell{l}{\cite{zhu2020deep}, \cite{park2020unsupervised}, \cite{wang2020self}, \cite{cao2021bipartite}, \cite{ren2019heterogeneous}, \cite{jing2021hdmi}, \\ \cite{wang2021self}} \\ \hline
ogbn-arxiv \cite{hu2020open} & Citation Network & 1 & 169343 & 1166243 & 128 & 40 & \cite{thakoor2021bootstrapped}, \cite{ma2021improving} \\ \hline

Reddit \cite{hamilton2017inductive} & Social Network & 1 & 232965 & 11606919 & 602 & 41 & \tabincell{l}{\cite{jin2020self}, \cite{hu2020gpt}, \cite{zhu2020deep}, \cite{peng2020graph}, \cite{velickovic2019deep}, \cite{jiao2020sub}, \\ \cite{peng2020self}, \cite{ma2021improving}} \\ \hline
BlogCatalog \cite{li2015unsupervised} & Social Network & 1 & 5196 & 171743 & 8189 & 6 & \cite{peng2020graph}, \cite{peng2020self}, \cite{huang2021hop} \\ \hline
Flickr \cite{li2015unsupervised} & Social Network & 1 & 7575 & 239738 & 12047 & 9 & \cite{peng2020graph}, \cite{jiao2020sub}, \cite{peng2020self}, \cite{huang2021hop} \\ \hline
COLLAB \cite{yanardag2015deep} & Social Networks & 5000 & 74.49 & 2457.78 & - & 2 & \cite{you2020graph}, \cite{zhang2020iterative}, \cite{ren2021label}, \cite{qiu2020gcc} \\ \hline
RDT-B \cite{yanardag2015deep} & Social Networks & 2000 & 429.63 & 497.75 & - & 2 & \tabincell{l}{\cite{you2020graph}, \cite{qiu2020gcc}, \cite{hassani2020contrastive}, \cite{sun2019infograph}, \cite{wang2021graph}, \cite{gao2021topoter}, \\ \cite{verma2020towards}} \\ \hline
RDT-M5K \cite{yanardag2015deep} & Social Networks & 4999 & 508.52 & 594.87 & - & 5 & \cite{you2020graph}, \cite{qiu2020gcc}, \cite{sun2019infograph}, \cite{wang2021graph}, \cite{gao2021topoter}, \cite{verma2020towards} \\ \hline
IMDB-B \cite{yanardag2015deep} & Social Networks & 1000 & 19.77 & 96.53 & - & 2 & \tabincell{l}{\cite{you2020graph}, \cite{zhang2020iterative}, \cite{ren2021label}, \cite{qiu2020gcc}, \cite{hassani2020contrastive}, \cite{sun2019infograph}, \\ \cite{hu2019pre}, \cite{park2020unsupervised}, \cite{hwang2020self}, \cite{ren2019heterogeneous}, \cite{wang2021graph}, \cite{jing2021hdmi}, \\ \cite{gao2021topoter}, \cite{verma2020towards}} \\ \hline
IMDB-M \cite{yanardag2015deep} & Social Networks & 1500 & 13.00 & 65.94 & - & 3 & \tabincell{l}{\cite{zhang2020iterative}, \cite{ren2021label}, \cite{qiu2020gcc}, \cite{hassani2020contrastive}, \cite{sun2019infograph}, \cite{hu2019pre}, \\  \cite{park2020unsupervised}, \cite{hwang2020self}, \cite{ren2019heterogeneous}, \cite{wang2021graph}, \cite{jing2021hdmi}, \cite{gao2021topoter}, \\ \cite{verma2020towards}} \\ \hline
ML-100K \cite{harper2015movielens} & Social Networks & 1 & 2625 & 100000 & - & 5 & \cite{hu2019pre}, \cite{cao2021bipartite} \\ \hline
ML-1M \cite{harper2015movielens} & Social Networks & 1 & 9940 & 1000209 & - & 5 & \cite{hu2019pre}, \cite{cao2021bipartite}, \cite{hao2021pre} \\ \hline

PPI \cite{zitnik2017predicting} & Protein Networks & 24 & 56944 & 818716 & 50 & 121 & \tabincell{l}{\cite{hu2019strategies}, \cite{you2020graph}, \cite{zhu2020deep}, \cite{peng2020graph}, \cite{peng2020graph}, \cite{velickovic2019deep}, \\ \cite{jiao2020sub}, \cite{peng2020self}, \cite{thakoor2021bootstrapped}} \\ \hline
D\&D \cite{dobson2003distinguishing, shervashidze2011weisfeiler} & Protein Networks & 1178 & 284.32 & 715.65 & 82 & 2 & \cite{you2020graph}, \cite{ren2021label}, \cite{zeng2020contrastive}, \cite{sun2021sugar} \\ \hline
PROTEINS \cite{dobson2003distinguishing,borgwardt2005protein} & Protein Networks & 1113 & 39.06 & 72.81 & 4 & 2 & \cite{you2020graph}, \cite{ren2021label}, \cite{zeng2020contrastive}, \cite{sun2021sugar} \\ \hline

NCI1 \cite{wale2008comparison,yanardag2015deep} & Molecule Graphs & 4110 & 29.87 & 32.30 & 37 & 2 & \cite{you2020graph}, \cite{zhang2020iterative}, \cite{ren2021label}, \cite{zeng2020contrastive}, \cite{sun2021sugar} \\ \hline
MUTAG \cite{debnath1991structure,kriege2012subgraph} & Molecule Graphs & 188 & 17.93 & 19.79 & 7 & 2 & \tabincell{l}{\cite{hu2019strategies}, \cite{you2020graph}, \cite{zhang2020iterative}, \cite{ren2021label}, \cite{hassani2020contrastive}, \cite{sun2019infograph}, \\ \cite{sun2021sugar}, \cite{wang2021graph}, \cite{gao2021topoter}, \cite{verma2020towards}} \\ \hline
QM9 (QM7, QM8) \cite{ramakrishnan2014quantum} & Molecule Graphs & 133885 & - & - & - & - & \cite{zhang2020iterative}, \cite{sun2019infograph}, \cite{jin2020self}, \cite{wang2021graph} \\ \hline
BBBP \cite{martins2012bayesian,hu2020open} & Molecule Graphs & 2039 & 24.05 & 25.94 & - & 2 & \cite{hu2019strategies}, \cite{you2020graph}, \cite{zhang2020motif}, \cite{rong2020self}, \cite{xu2021selfsupervised} \\ \hline
Tox21 \cite{mayr2016deeptox,huang2016tox21challenge,hu2020open} & Molecule Graphs & 7831 & 18.51 & 25.94 & - & 12 & \cite{hu2019strategies}, \cite{you2020graph}, \cite{zhang2020motif}, \cite{rong2020self}, \cite{xu2021selfsupervised} \\ \hline
ToxCast \cite{richard2016toxcast,hu2020open} & Molecule Graphs & 8575 & 18.78 & 19.26 & - & 167 & \cite{hu2019strategies}, \cite{you2020graph}, \cite{zhang2020motif}, \cite{rong2020self}, \cite{xu2021selfsupervised} \\ \hline
ClinTox \cite{novick2013sweetlead} & Molecule Graphs & 1478 & 26.13 & 27.86 & - & 2 & \cite{hu2019strategies}, \cite{you2020graph}, \cite{zhang2020motif}, \cite{rong2020self}, \cite{xu2021selfsupervised} \\ \hline
MUV \cite{gardiner2011effectiveness} & Molecule Graphs & 93087 & 24.23 & 26.28 & - & 17 & \cite{hu2019strategies}, \cite{you2020graph}, \cite{xu2021selfsupervised} \\ \hline
HIV \cite{hu2020open} & Molecule Graphs & 41127 & 25.53 & 27.48 & - & 2 & \cite{hu2019strategies}, \cite{you2020graph}, \cite{zhang2020motif}, \cite{xu2021selfsupervised} \\ \hline
SIDER \cite{kuhn2016sider} & Molecule Graphs & 1427 & 33.64 & 35.36 & - & 27 & \cite{hu2019strategies}, \cite{you2020graph}, \cite{zhang2020motif}, \cite{rong2020self}, \cite{xu2021selfsupervised} \\ \hline
BACE \cite{subramanian2016computational} & Molecule Graphs & 1513 & 34.12 & 36.89 & - & 2 & \cite{hu2019strategies}, \cite{you2020graph}, \cite{zhang2020motif}, \cite{rong2020self}, \cite{xu2021selfsupervised} \\ \hline
PTC \cite{shervashidze2011weisfeiler} & Molecule Graphs & 344 & 14.29 & 14.69 & 19 & 2 & \tabincell{l}{\cite{zhang2020iterative}, \cite{hassani2020contrastive}, \cite{sun2019infograph}, \cite{wang2021graph}, \cite{gao2021topoter}, \cite{verma2020towards}, \\ \cite{hu2019strategies}, \cite{ren2021label}, \cite{sun2021sugar}} \\ \hline
NCI109 \cite{wale2008comparison,yanardag2015deep} & Molecule Graphs & 4127 & 29.68 & 32.13 & - & 2 & \cite{zeng2020contrastive}, \cite{sun2021sugar} \\ \hline
Mutagenicity \cite{riesen2008iam,kazius2005derivation} & Molecule Graphs & 4337 & 30.32 & 30.77 & - & 2 & \cite{zeng2020contrastive} \\ \hline

MNIST \cite{lecun1998gradient} & Others (Image) & - & 70000 & - & 784 & 10 & \cite{you2020graph} \\ \hline
CIFAR10 \cite{krizhevsky2009learning} & Others (Image) & - & 60000 & - & 1024 & 10 & \cite{you2020graph} \\ \hline
METR-LA \cite{jagadish2014big} & Others (Traffic) & 1 & 207 & 1515 & 2 & - & \cite{opolka2019spatio} \\ \hline
\tabincell{l}{Amazon- \\ Computers \cite{shchur2018pitfalls}} & Others (Purchase) & 1 & 13752 & 245861 & 767 & 10 & \tabincell{l}{\cite{zhu2020graph}, \cite{wan2020contrastive}, \cite{mavromatis2020graph}, \cite{hu2020gpt}, \cite{wang2020self}, \cite{thakoor2021bootstrapped}, \\ \cite{huang2021hop}, \cite{jing2021hdmi}, \cite{ma2021improving}, \cite{kefato2021self}, \cite{yu2020self}} \\ \hline
Amazon-Photo \cite{shchur2018pitfalls} & Others (Purchase) & 1 & 7650 & 119081 & 745 & 8 & \tabincell{l}{\cite{zhu2020graph}, \cite{jovanovic2021towards}, \cite{wan2020contrastive}, \cite{mavromatis2020graph}, \cite{hu2020gpt}, \cite{wang2020self}, \\ \cite{thakoor2021bootstrapped}, \cite{huang2021hop}, \cite{jing2021hdmi}, \cite{ma2021improving}, \cite{kefato2021self}, \cite{yu2020self}} \\ \hline
ogbn-products \cite{hu2020open} & Others (Purchase) & 1 & 2449029 & 61859140 & 100 & 47 & \cite{ma2021improving} \\ \hline

\end{tabular}
\end{table*}

\begin{table*}[!htbp]
\caption{A summary of open-source codes of the surveyed papers.}
\label{tabA4:github}
\centering
\begin{tabular}{ll}

\toprule
\multicolumn{1}{c}{Methods} & \multicolumn{1}{c}{Github} \\ 
\hline

Graph Completion \cite{you2020does} & \url{https:
//github.com/Shen-Lab/SS-GCNs} \\ \hline
Node Attribute Masking \cite{jin2020self} & \url{https://github.com/ChandlerBang/SelfTask-GNN} \\ \hline
Edge Attribute Masking \cite{hu2019strategies} & \url{http://snap.stanford.edu/gnn-pretrain} \\ \hline
Node Attribute Denoising \cite{manessi2020graph} & N.A. \\ \hline
\tabincell{l}{Adjacency Matrix \\ Reconstruction \cite{zhu2020self}} & N.A. \\ \hline
Graph Bert \cite{zhang2020graph} & \url{https://github.com/anonymous-sourcecode/Graph-Bert} \\ \hline
Pretrain-Recsys \cite{hao2021pre} & \url{https://github.com/jerryhao66/Pretrain-Recsys} \\ \hline
GPT-GNN \cite{hu2020gpt} & \url{https://github.com/acbull/GPT-GNN} \\ \hline

GraphCL \cite{you2020graph} & \url{https://github.com/Shen-Lab/GraphCL} \\ \hline
IGSD \cite{zhang2020iterative} & N.A. \\ \hline
DACL \cite{verma2020towards} & N.A. \\ \hline
LCC \cite{ren2021label} & \tabincell{l}{\url{https://github.com/YuxiangRen}} \\ \hline
CSSL \cite{zeng2020contrastive} & N.A. \\ \hline
GCC \cite{qiu2020gcc} & \url{https://github.com/THUDM/GCC} \\ \hline
GRACE \cite{zhu2020deep} & \url{https://github.com/CRIPAC-DIG/GRACE} \\ \hline
GCA \cite{zhu2020graph} & \url{https://github.com/CRIPAC-DIG/GCA} \\ \hline
GROC \cite{jovanovic2021towards} & N.A. \\ \hline
SEPT \cite{yu2021socially} & \url{https://github.com/Coder-Yu/QRec} \\ \hline
STDGI \cite{opolka2019spatio} & N.A. \\ \hline
GMI \cite{peng2020graph} & \url{https://github.com/zpeng27/GMI} \\ \hline
KS2L \cite{yu2020self} & N.A. \\ \hline
$\text{C}\text{G}^3$ \cite{wan2020contrastive} & N.A. \\ \hline
BGRL \cite{thakoor2021bootstrapped} & N.A. \\ \hline
SelfGNN \cite{kefato2021self} & \url{https://github.com/zekarias-tilahun/SelfGNN} \\ \hline
HeCo \cite{wang2021self} & \url{https://github.com/liun-online/HeCo} \\ \hline
PT-DGNN \cite{zhang2021pre} & \url{https://github.com/Mobzhang/PT-DGNN} \\ \hline
COAD \cite{chen2020coad} & \url{https://github.com/allanchen95/Expert-Linking} \\ \hline
Contrast-Reg \cite{ma2021improving} & N.A. \\ \hline

DGI \cite{velickovic2019deep} & \url{https://github.com/PetarV-/DGI} \\ \hline
HDMI \cite{jing2021hdmi} & N.A. \\ \hline
DMGI \cite{park2020unsupervised} & \url{https://github.com/pcy1302/DMGI} \\ \hline
MVGRL \cite{hassani2020contrastive} & \url{https://github.com/kavehhassani/mvgrl} \\ \hline
HDGI \cite{ren2019heterogeneous} & \url{https://github.com/YuxiangRen/Heterogeneous-Deep-Graph-Infomax} \\ \hline
Subg-Con \cite{jiao2020sub} & \url{https://github.com/yzjiao/Subg-Con} \\ \hline
Cotext Prediction \cite{hu2019strategies} & \url{http://snap.stanford.edu/gnn-pretrain} \\ \hline
GIC \cite{mavromatis2020graph} & \url{https://github.com/cmavro/Graph-InfoClust-GIC} \\ \hline
GraphLoG \cite{xu2021selfsupervised} & \url{https://openreview.net/forum?id=DAaaaqPv9-q} \\ \hline
MHCN \cite{yu2021self} & \url{https://github.com/Coder-Yu/RecQ} \\ \hline
EGI \cite{zhu2020transfer} & \url{https://openreview.net/forum?id=J_pvI6ap5Mn} \\ \hline
MICRO-Graph \cite{zhang2020motif} & \url{https://drive.google.com/file/d/1b751rpnV-SDmUJvKZZI-AvpfEa9eHxo9}  \\ \hline
InfoGraph \cite{sun2019infograph} & \url{https://github.com/fanyun-sun/InfoGraph} \\ \hline
SUGAR \cite{sun2021sugar}  & \url{https://github.com/RingBDStack/SUGAR} \\ \hline
BiGI \cite{cao2021bipartite} & \url{https://github.com/clhchtcjj/BiNE} \\ \hline
HTC \cite{wang2021graph} & N.A. \\ \hline

Node Property Prediction \cite{jin2020self} & \url{https://github.com/ChandlerBang/SelfTask-GNN} \\ \hline
\tabincell{l}{$\text{S}^2\text{GRL}$ \cite{peng2020self}} & N.A. \\ \hline
PairwiseDistance \cite{jin2020self} & \url{https://github.com/ChandlerBang/SelfTask-GNN} \\ \hline
PairwiseAttrSim \cite{jin2020self} & \url{https://github.com/ChandlerBang/SelfTask-GNN} \\ \hline
Distance2Cluster \cite{jin2020self} & \url{https://github.com/ChandlerBang/SelfTask-GNN} \\ \hline
EdgeMask \cite{jin2020self} & \url{https://github.com/ChandlerBang/SelfTask-GNN} \\ \hline
TopoTER \cite{gao2021topoter} & N.A. \\ \hline
Centrality Score Ranking \cite{hu2019pre} & N.A. \\ \hline
Meta-path Prediction \cite{hwang2020self} & \url{https://github.com/mlvlab/SELAR} \\ \hline
SLiCE \cite{wang2020self} & \url{https://github.com/pnnl/SLICE} \\ \hline
Distance2Labeled \cite{jin2020self} & \url{https://github.com/ChandlerBang/SelfTask-GNN} \\ \hline
ContextLabel \cite{jin2020self} & \url{https://github.com/ChandlerBang/SelfTask-GNN} \\ \hline
HCM \cite{huang2021hop} & N.A. \\ \hline

\tabincell{l}{Contextual Molecular \\ Property Prediction \cite{rong2020self}} & \url{https://github.com/tencent-ailab/grover} \\ \hline
Graph-level Motif Prediction \cite{rong2020self} & \url{https://github.com/tencent-ailab/grover} \\ \hline
Multi-stage Self-training \cite{li2018deeper} & \url{https://github.com/Davidham3/deeper_insights_into_GCNs}\\ \hline
Node Clustering \cite{you2020does} & \url{https:
//github.com/Shen-Lab/SS-GCNs} \\ \hline
Graph Partitioning \cite{you2020does} & \url{https:
//github.com/Shen-Lab/SS-GCNs} \\ \hline
CAGNN \cite{zhu2020cagnn} & N.A. \\ \hline
M3S \cite{sun2020multi} & \url{https://github.com/datake/M3S} \\ \hline
Cluster Preserving \cite{hu2019pre} & N.A. \\ \hline

\end{tabular}
\end{table*}

\begin{table*}[!htbp]
\caption{\textcolor{mark}{A summary of node classification accuracy (\%) of different methods on 8 commonly used datasets. When the code is not publicly available, or it is not impractical to run the released code from scratch, we replace the corresponding results with a dash ``-".}}
\label{tabA5:node}
\centering
\begin{tabular}{lcccccccc}

\toprule
& Cora & Citeseer & Pubmed & Wiki-CS & \tabincell{c}{Amazon-\\Computers} & Amazon-Photo & Coauthor-CS & \tabincell{c}{Coauthor-\\Physics} \\ 
\midrule

GRACE \cite{zhu2020deep} & 80.00 & 71.70 & 79.50 & 78.19 & 87.46 & 92.15 & 92.93 & 95.26 \\ \midrule
GMI \cite{peng2020graph} & 83.00 & 72.40 & 79.90 & 74.85 & 82.21 & 90.68 & - & - \\ \midrule
\tabincell{l}{Adjacency Matrix \\ Reconstruction \cite{zhu2020self}} & 83.80 & 72.95 & 81.23 & - & - & - & - & - \\ \midrule
Graph Bert \cite{zhang2020graph} & 84.30 & 71.20 & 79.30 & - & 87.57 & - & 92.99 & 95.41 \\ \midrule
$\text{C}\text{G}^3$ \cite{wan2020contrastive} & 83.40 & 73.60 & 80.20 & - & 88.06 & 92.21 & 92.30 & - \\ \midrule
$\text{S}^2\text{GRL}$ \cite{peng2020self} & 83.70 & 72.10 & 82.40 & 78.58 & 88.49 & - & 92.66 & 95.24 \\ \midrule
PairwiseDistance \cite{jin2020self} & 83.11 & 71.90 & 80.05 & - & - & - & - & - \\ \midrule
PairwiseAttrSim \cite{jin2020self} & 83.05 & 71.67 & 79.45 & - & - & - & - & - \\ \midrule
Contrast-Reg \cite{ma2021improving} & 82.65 & 72.98 & 80.10 & 77.13 & 84.93 & 91.09 & - & 94.38 \\ \midrule
Distance2Cluster \cite{jin2020self} & 83.55 & 71.44 & 79.88 & - & - & - & - & - \\ \midrule
TopoTER \cite{gao2021topoter} & 83.70 & 71.70 & 79.10 & - & - & - & 92.87 & 95.22 \\ \midrule
DGI \cite{velickovic2019deep} & 82.30 & 71.80 & 76.80 & 75.35 & 83.95 & 91.61 & 92.15 & 94.51 \\ \midrule
MVGRL \cite{hassani2020contrastive} & 82.90 & 72.60 & 79.40 & 77.52 & 87.52 & 91.74 & 92.11 & 95.33 \\ \midrule
Subg-Con \cite{jiao2020sub} & 83.50 & 73.20 & 81.00 & 78.69 & 88.65 & 92.42 & - & - \\ \midrule
Distance2Labeled \cite{jin2020self} & 83.39 & 71.64 & 79.51 & - & - & - & - & - \\ \midrule
ContextLabel \cite{jin2020self} & 82.76 & 72.59 & 82.31 & - & - & - & - & - \\ \midrule
GIC \cite{mavromatis2020graph} & 81.70 & 71.90 & 77.30 & - & 84.89 & 92.11 & 92.51 & 94.70 \\ \midrule
\tabincell{l}{Node Property Prediction \cite{jin2020self}} & 81.94 & 71.60 & 79.44 & - & - & - & - & - \\ \midrule
GCA \cite{zhu2020graph} & - & - & - & 78.35 & 88.94 & 92.53 & 93.10 & \textbf{95.75} \\ \midrule
M3S \cite{sun2020multi} & 81.60 & 71.94 & 79.28 & - & - & - & - & - \\ \midrule
KS2L \cite{yu2020self} & 84.60 & \textbf{74.20} & \textbf{83.80} & - & 86.80 & 92.40 & \textbf{93.30} & - \\ \midrule
BGRL \cite{thakoor2021bootstrapped} & - & - & - & \textbf{79.36} & \textbf{89.68} & 92.87 & 93.21 & 95.56 \\ \midrule
SelfGNN \cite{kefato2021self} & \textbf{85.30} & 72.30 & 82.70 & - & 88.80 & \textbf{93.80} & 92.90 & 95.50 \\ \bottomrule

\end{tabular}
\end{table*}
\begin{table*}[!htbp]
\caption{\textcolor{mark}{A summary of graph classification accuracy (\%) of different methods on 7 commonly used datasets. When the code is not publicly available, or it is not impractical to run the released code from scratch, we replace the corresponding results with a dash ``-".}}
\label{tabA6:graph}
\centering
\begin{tabular}{lccccccc}

\toprule
& MUTAG & PTC & NCI1 & RDT-B & RDT-M5K & IMDB-B & IMDB-M \\ 
\midrule

GraphCL \cite{you2020graph} & 86.80 & - & 77.87 & 89.53 & \textbf{55.99} & 71.14 & 51.83 \\ \midrule
IGSD \cite{zhang2020iterative} & 90.20 & 61.40 & 75.40 & - & - & 74.70 & 51.50 \\ \midrule
DACL \cite{verma2020towards} & 87.51 & 63.59 & - & 85.11 & 53.20 & 73.98 & 50.78 \\ \midrule
LCC \cite{ren2021label} & 90.50 & 65.90 & 82.90 & - & - & \textbf{76.10} & \textbf{52.40} \\ \midrule
CSSL \cite{zeng2020contrastive} & 88.95 & - & 80.09 & 84.21 & 53.76 & - & - \\ \midrule
GCC \cite{qiu2020gcc} & - & - & - & 87.80 & 53.00 & 75.60 & 50.90 \\ \midrule
MVGRL \cite{hassani2020contrastive} & 89.70 & 62.50 & - & 84.50 & 54.03 & 74.20 & 51.20 \\ \midrule
InfoGraph \cite{sun2019infograph} & 89.01 & 61.65 & - & 82.50 & 53.46 & 73.03 & 49.69 \\ \midrule
SUGAR \cite{sun2021sugar} & \textbf{96.74} & \textbf{77.53} & \textbf{84.39} & - & - & - & - \\ \midrule
HTC \cite{wang2021graph} & 91.80 & 65.50 & - & \textbf{91.10} & 55.70 & 73.30 & 50.60 \\ \midrule
TopoTER \cite{gao2021topoter} & 89.25 & 64.59 & - & 84.93 & 55.52 & 73.46 & 49.68 \\ \bottomrule

\end{tabular}
\end{table*}

\clearpage

\clearpage
\bibliographystyle{IEEEtran}
\bibliography{Bibliography-File}

\begin{thebibliography}{100}
\providecommand{\url}[1]{#1}
\csname url@samestyle\endcsname
\providecommand{\newblock}{\relax}
\providecommand{\bibinfo}[2]{#2}
\providecommand{\BIBentrySTDinterwordspacing}{\spaceskip=0pt\relax}
\providecommand{\BIBentryALTinterwordstretchfactor}{4}
\providecommand{\BIBentryALTinterwordspacing}{\spaceskip=\fontdimen2\font plus
\BIBentryALTinterwordstretchfactor\fontdimen3\font minus
  \fontdimen4\font\relax}
\providecommand{\BIBforeignlanguage}[2]{{%
\expandafter\ifx\csname l@#1\endcsname\relax
\typeout{** WARNING: IEEEtran.bst: No hyphenation pattern has been}%
\typeout{** loaded for the language `#1'. Using the pattern for}%
\typeout{** the default language instead.}%
\else
\language=\csname l@#1\endcsname
\fi
#2}}
\providecommand{\BIBdecl}{\relax}
\BIBdecl

\bibitem{liu2020self}
X.~Liu, F.~Zhang, Z.~Hou, Z.~Wang, L.~Mian, J.~Zhang, and J.~Tang,
  ``Self-supervised learning: Generative or contrastive,'' \emph{arXiv preprint
  arXiv:2006.08218}, vol.~1, no.~2, 2020.

\bibitem{he2020momentum}
K.~He, H.~Fan, Y.~Wu, S.~Xie, and R.~Girshick, ``Momentum contrast for
  unsupervised visual representation learning,'' in \emph{Proceedings of the
  IEEE/CVF Conference on Computer Vision and Pattern Recognition}, 2020, pp.
  9729--9738.

\bibitem{chen2020simple}
T.~Chen, S.~Kornblith, M.~Norouzi, and G.~Hinton, ``A simple framework for
  contrastive learning of visual representations,'' in \emph{International
  conference on machine learning}.\hskip 1em plus 0.5em minus 0.4em\relax PMLR,
  2020, pp. 1597--1607.

\bibitem{grill2020bootstrap}
J.-B. Grill, F.~Strub, F.~Altch{\'e}, C.~Tallec, P.~H. Richemond,
  E.~Buchatskaya, C.~Doersch, B.~A. Pires, Z.~D. Guo, M.~G. Azar \emph{et~al.},
  ``Bootstrap your own latent: A new approach to self-supervised learning,''
  \emph{arXiv preprint arXiv:2006.07733}, 2020.

\bibitem{devlin2018bert}
J.~Devlin, M.-W. Chang, K.~Lee, and K.~Toutanova, ``Bert: Pre-training of deep
  bidirectional transformers for language understanding,'' \emph{arXiv preprint
  arXiv:1810.04805}, 2018.

\bibitem{radford2019language}
A.~Radford, J.~Wu, R.~Child, D.~Luan, D.~Amodei, and I.~Sutskever, ``Language
  models are unsupervised multitask learners,'' \emph{OpenAI blog}, vol.~1,
  no.~8, p.~9, 2019.

\bibitem{lan2019albert}
Z.~Lan, M.~Chen, S.~Goodman, K.~Gimpel, P.~Sharma, and R.~Soricut, ``Albert: A
  lite bert for self-supervised learning of language representations,''
  \emph{arXiv preprint arXiv:1909.11942}, 2019.

\bibitem{hamilton2017inductive}
W.~L. Hamilton, R.~Ying, and J.~Leskovec, ``Inductive representation learning
  on large graphs,'' \emph{arXiv preprint arXiv:1706.02216}, 2017.

\bibitem{velickovic2019deep}
P.~Velickovic, W.~Fedus, W.~L. Hamilton, P.~Li{\`o}, Y.~Bengio, and R.~D.
  Hjelm, ``Deep graph infomax.'' in \emph{ICLR (Poster)}, 2019.

\bibitem{jin2020self}
W.~Jin, T.~Derr, H.~Liu, Y.~Wang, S.~Wang, Z.~Liu, and J.~Tang,
  ``Self-supervised learning on graphs: Deep insights and new direction,''
  \emph{arXiv preprint arXiv:2006.10141}, 2020.

\bibitem{hu2019strategies}
W.~Hu, B.~Liu, J.~Gomes, M.~Zitnik, P.~Liang, V.~Pande, and J.~Leskovec,
  ``Strategies for pre-training graph neural networks,'' \emph{arXiv preprint
  arXiv:1905.12265}, 2019.

\bibitem{you2020does}
Y.~You, T.~Chen, Z.~Wang, and Y.~Shen, ``When does self-supervision help graph
  convolutional networks?'' in \emph{International Conference on Machine
  Learning}.\hskip 1em plus 0.5em minus 0.4em\relax PMLR, 2020, pp.
  10\,871--10\,880.

\bibitem{manessi2020graph}
F.~Manessi and A.~Rozza, ``Graph-based neural network models with multiple
  self-supervised auxiliary tasks,'' \emph{arXiv preprint arXiv:2011.07267},
  2020.

\bibitem{zhu2020self}
Q.~Zhu, B.~Du, and P.~Yan, ``Self-supervised training of graph convolutional
  networks,'' \emph{arXiv preprint arXiv:2006.02380}, 2020.

\bibitem{zhang2020graph}
J.~Zhang, H.~Zhang, C.~Xia, and L.~Sun, ``Graph-bert: Only attention is needed
  for learning graph representations,'' \emph{arXiv preprint arXiv:2001.05140},
  2020.

\bibitem{peng2020graph}
Z.~Peng, W.~Huang, M.~Luo, Q.~Zheng, Y.~Rong, T.~Xu, and J.~Huang, ``Graph
  representation learning via graphical mutual information maximization,'' in
  \emph{Proceedings of The Web Conference 2020}, 2020, pp. 259--270.

\bibitem{hu2019pre}
Z.~Hu, C.~Fan, T.~Chen, K.-W. Chang, and Y.~Sun, ``Pre-training graph neural
  networks for generic structural feature extraction,'' \emph{arXiv preprint
  arXiv:1905.13728}, 2019.

\bibitem{hassani2020contrastive}
K.~Hassani and A.~H. Khasahmadi, ``Contrastive multi-view representation
  learning on graphs,'' in \emph{International Conference on Machine
  Learning}.\hskip 1em plus 0.5em minus 0.4em\relax PMLR, 2020, pp. 4116--4126.

\bibitem{jiao2020sub}
Y.~Jiao, Y.~Xiong, J.~Zhang, Y.~Zhang, T.~Zhang, and Y.~Zhu, ``Sub-graph
  contrast for scalable self-supervised graph representation learning,''
  \emph{arXiv preprint arXiv:2009.10273}, 2020.

\bibitem{xie2021self}
Y.~Xie, Z.~Xu, Z.~Wang, and S.~Ji, ``Self-supervised learning of graph neural
  networks: A unified review,'' \emph{arXiv preprint arXiv:2102.10757}, 2021.

\bibitem{liu2021graph}
Y.~Liu, S.~Pan, M.~Jin, C.~Zhou, F.~Xia, and P.~S. Yu, ``Graph self-supervised
  learning: A survey,'' \emph{arXiv preprint arXiv:2103.00111}, 2021.

\bibitem{kipf2016semi}
T.~N. Kipf and M.~Welling, ``Semi-supervised classification with graph
  convolutional networks,'' \emph{arXiv preprint arXiv:1609.02907}, 2016.

\bibitem{velivckovic2017graph}
P.~Veli{\v{c}}kovi{\'c}, G.~Cucurull, A.~Casanova, A.~Romero, P.~Lio, and
  Y.~Bengio, ``Graph attention networks,'' \emph{arXiv preprint
  arXiv:1710.10903}, 2017.

\bibitem{wu2020comprehensive}
Z.~Wu, S.~Pan, F.~Chen, G.~Long, C.~Zhang, and S.~Y. Philip, ``A comprehensive
  survey on graph neural networks,'' \emph{IEEE transactions on neural networks
  and learning systems}, 2020.

\bibitem{you2020graph}
Y.~You, T.~Chen, Y.~Sui, T.~Chen, Z.~Wang, and Y.~Shen, ``Graph contrastive
  learning with augmentations,'' \emph{Advances in Neural Information
  Processing Systems}, vol.~33, 2020.

\bibitem{zhu2020deep}
Y.~Zhu, Y.~Xu, F.~Yu, Q.~Liu, S.~Wu, and L.~Wang, ``Deep graph contrastive
  representation learning,'' \emph{arXiv preprint arXiv:2006.04131}, 2020.

\bibitem{thakoor2021bootstrapped}
S.~Thakoor, C.~Tallec, M.~G. Azar, R.~Munos, P.~Veli{\v{c}}kovi{\'c}, and
  M.~Valko, ``Bootstrapped representation learning on graphs,'' \emph{arXiv
  preprint arXiv:2102.06514}, 2021.

\bibitem{opolka2019spatio}
F.~L. Opolka, A.~Solomon, C.~Cangea, P.~Veli{\v{c}}kovi{\'c}, P.~Li{\`o}, and
  R.~D. Hjelm, ``Spatio-temporal deep graph infomax,'' \emph{arXiv preprint
  arXiv:1904.06316}, 2019.

\bibitem{ma2021improving}
K.~Ma, H.~Yang, H.~Yang, T.~Jin, P.~Chen, Y.~Chen, B.~F. Kamhoua, and J.~Cheng,
  ``Improving graph representation learning by contrastive regularization,''
  \emph{arXiv preprint arXiv:2101.11525}, 2021.

\bibitem{jing2021hdmi}
B.~Jing, C.~Park, and H.~Tong, ``Hdmi: High-order deep multiplex infomax,''
  \emph{arXiv preprint arXiv:2102.07810}, 2021.

\bibitem{ren2019heterogeneous}
Y.~Ren, B.~Liu, C.~Huang, P.~Dai, L.~Bo, and J.~Zhang, ``Heterogeneous deep
  graph infomax,'' \emph{arXiv preprint arXiv:1911.08538}, 2019.

\bibitem{hu2020gpt}
Z.~Hu, Y.~Dong, K.~Wang, K.-W. Chang, and Y.~Sun, ``Gpt-gnn: Generative
  pre-training of graph neural networks,'' in \emph{Proceedings of the 26th ACM
  SIGKDD International Conference on Knowledge Discovery \& Data Mining}, 2020,
  pp. 1857--1867.

\bibitem{zhang2020iterative}
H.~Zhang, S.~Lin, W.~Liu, P.~Zhou, J.~Tang, X.~Liang, and E.~P. Xing,
  ``Iterative graph self-distillation,'' \emph{arXiv preprint
  arXiv:2010.12609}, 2020.

\bibitem{zeng2020contrastive}
J.~Zeng and P.~Xie, ``Contrastive self-supervised learning for graph
  classification,'' \emph{arXiv preprint arXiv:2009.05923}, 2020.

\bibitem{kefato2021self}
Z.~T. Kefato and S.~Girdzijauskas, ``Self-supervised graph neural networks
  without explicit negative sampling,'' \emph{arXiv preprint arXiv:2103.14958},
  2021.

\bibitem{zhu2020transfer}
Q.~Zhu, Y.~Xu, H.~Wang, C.~Zhang, J.~Han, and C.~Yang, ``Transfer learning of
  graph neural networks with ego-graph information maximization,'' \emph{arXiv
  preprint arXiv:2009.05204}, 2020.

\bibitem{cao2021bipartite}
J.~Cao, X.~Lin, S.~Guo, L.~Liu, T.~Liu, and B.~Wang, ``Bipartite graph
  embedding via mutual information maximization,'' in \emph{Proceedings of the
  14th ACM International Conference on Web Search and Data Mining}, 2021, pp.
  635--643.

\bibitem{qiu2020gcc}
J.~Qiu, Q.~Chen, Y.~Dong, J.~Zhang, H.~Yang, M.~Ding, K.~Wang, and J.~Tang,
  ``Gcc: Graph contrastive coding for graph neural network pre-training,'' in
  \emph{Proceedings of the 26th ACM SIGKDD International Conference on
  Knowledge Discovery \& Data Mining}, 2020, pp. 1150--1160.

\bibitem{zhang2020motif}
S.~Zhang, Z.~Hu, A.~Subramonian, and Y.~Sun, ``Motif-driven contrastive
  learning of graph representations,'' \emph{arXiv preprint arXiv:2012.12533},
  2020.

\bibitem{zhu2020graph}
Y.~Zhu, Y.~Xu, F.~Yu, Q.~Liu, S.~Wu, and L.~Wang, ``Graph contrastive learning
  with adaptive augmentation,'' \emph{arXiv preprint arXiv:2010.14945}, 2020.

\bibitem{bonacich1987power}
P.~Bonacich, ``Power and centrality: A family of measures,'' \emph{American
  journal of sociology}, vol.~92, no.~5, pp. 1170--1182, 1987.

\bibitem{page1999pagerank}
L.~Page, S.~Brin, R.~Motwani, and T.~Winograd, ``The pagerank citation ranking:
  Bringing order to the web.'' Stanford InfoLab, Tech. Rep., 1999.

\bibitem{jovanovic2021towards}
N.~Jovanovi{\'c}, Z.~Meng, L.~Faber, and R.~Wattenhofer, ``Towards robust graph
  contrastive learning,'' \emph{arXiv preprint arXiv:2102.13085}, 2021.

\bibitem{kipf2016variational}
T.~N. Kipf and M.~Welling, ``Variational graph auto-encoders,'' \emph{arXiv
  preprint arXiv:1611.07308}, 2016.

\bibitem{ren2021label}
Y.~Ren, J.~Bai, and J.~Zhang, ``Label contrastive coding based graph neural
  network for graph classification,'' \emph{arXiv preprint arXiv:2101.05486},
  2021.

\bibitem{yu2021socially}
J.~Yu, H.~Yin, M.~Gao, X.~Xia, X.~Zhang, and N.~Q.~V. Hung, ``Socially-aware
  self-supervised tri-training for recommendation,'' \emph{arXiv preprint
  arXiv:2106.03569}, 2021.

\bibitem{wang2021self}
X.~Wang, N.~Liu, H.~Han, and C.~Shi, ``Self-supervised heterogeneous graph
  neural network with co-contrastive learning,'' \emph{arXiv preprint
  arXiv:2105.09111}, 2021.

\bibitem{mavromatis2020graph}
C.~Mavromatis and G.~Karypis, ``Graph infoclust: Leveraging cluster-level node
  information for unsupervised graph representation learning,'' \emph{arXiv
  preprint arXiv:2009.06946}, 2020.

\bibitem{sun2019infograph}
F.-Y. Sun, J.~Hoffmann, V.~Verma, and J.~Tang, ``Infograph: Unsupervised and
  semi-supervised graph-level representation learning via mutual information
  maximization,'' \emph{arXiv preprint arXiv:1908.01000}, 2019.

\bibitem{belghazi2018mutual}
M.~I. Belghazi, A.~Baratin, S.~Rajeshwar, S.~Ozair, Y.~Bengio, A.~Courville,
  and D.~Hjelm, ``Mutual information neural estimation,'' in
  \emph{International Conference on Machine Learning}.\hskip 1em plus 0.5em
  minus 0.4em\relax PMLR, 2018, pp. 531--540.

\bibitem{nowozin2016f}
S.~Nowozin, B.~Cseke, and R.~Tomioka, ``f-gan: Training generative neural
  samplers using variational divergence minimization,'' \emph{arXiv preprint
  arXiv:1606.00709}, 2016.

\bibitem{gutmann2010noise}
M.~Gutmann and A.~Hyv{\"a}rinen, ``Noise-contrastive estimation: A new
  estimation principle for unnormalized statistical models,'' in
  \emph{Proceedings of the Thirteenth International Conference on Artificial
  Intelligence and Statistics}.\hskip 1em plus 0.5em minus 0.4em\relax JMLR
  Workshop and Conference Proceedings, 2010, pp. 297--304.

\bibitem{sohn2016improved}
K.~Sohn, ``Improved deep metric learning with multi-class n-pair loss
  objective,'' in \emph{Proceedings of the 30th International Conference on
  Neural Information Processing Systems}, 2016, pp. 1857--1865.

\bibitem{tschannen2019mutual}
M.~Tschannen, J.~Djolonga, P.~K. Rubenstein, S.~Gelly, and M.~Lucic, ``On
  mutual information maximization for representation learning,'' \emph{arXiv
  preprint arXiv:1907.13625}, 2019.

\bibitem{schroff2015facenet}
F.~Schroff, D.~Kalenichenko, and J.~Philbin, ``Facenet: A unified embedding for
  face recognition and clustering,'' in \emph{Proceedings of the IEEE
  conference on computer vision and pattern recognition}, 2015, pp. 815--823.

\bibitem{chen2017beyond}
W.~Chen, X.~Chen, J.~Zhang, and K.~Huang, ``Beyond triplet loss: a deep
  quadruplet network for person re-identification,'' in \emph{Proceedings of
  the IEEE/CVF Conference on Computer Vision and Pattern Recognition}, 2017,
  pp. 403--412.

\bibitem{kemertas2020rankmi}
M.~Kemertas, L.~Pishdad, K.~G. Derpanis, and A.~Fazly, ``Rankmi: A mutual
  information maximizing ranking loss,'' in \emph{Proceedings of the IEEE/CVF
  Conference on Computer Vision and Pattern Recognition}, 2020, pp.
  14\,362--14\,371.

\bibitem{hinton2006reducing}
G.~E. Hinton and R.~R. Salakhutdinov, ``Reducing the dimensionality of data
  with neural networks,'' \emph{science}, vol. 313, no. 5786, pp. 504--507,
  2006.

\bibitem{peng2020self}
Z.~Peng, Y.~Dong, M.~Luo, X.-M. Wu, and Q.~Zheng, ``Self-supervised graph
  representation learning via global context prediction,'' \emph{arXiv preprint
  arXiv:2003.01604}, 2020.

\bibitem{Macqueen67somemethods}
J.~Macqueen, ``Some methods for classification and analysis of multivariate
  observations,'' in \emph{In 5-th Berkeley Symposium on Mathematical
  Statistics and Probability}, 1967, pp. 281--297.

\bibitem{gao2020clustering}
Z.~Gao, H.~Lin, S.~Li \emph{et~al.}, ``Clustering based on graph of density
  topology,'' \emph{arXiv preprint arXiv:2009.11612}, 2020.

\bibitem{wu2020deep}
L.~Wu, Z.~Liu, Z.~Zang, J.~Xia, S.~Li, S.~Li \emph{et~al.}, ``Deep clustering
  and representation learning that preserves geometric structures,''
  \emph{arXiv preprint arXiv:2009.09590}, 2020.

\bibitem{li2020consistent}
S.~Z. Li, L.~Wu, and Z.~Zang, ``Consistent representation learning for high
  dimensional data analysis,'' \emph{arXiv preprint arXiv:2012.00481}, 2020.

\bibitem{yang2019deep}
X.~Yang, C.~Deng, F.~Zheng, J.~Yan, and W.~Liu, ``Deep spectral clustering
  using dual autoencoder network,'' in \emph{Proceedings of the IEEE Conference
  on Computer Vision and Pattern Recognition}, 2019, pp. 4066--4075.

\bibitem{yang2016joint}
J.~Yang, D.~Parikh, and D.~Batra, ``Joint unsupervised learning of deep
  representations and image clusters,'' in \emph{Proceedings of the IEEE
  Conference on Computer Vision and Pattern Recognition}, 2016, pp. 5147--5156.

\bibitem{xie2016unsupervised}
J.~Xie, R.~Girshick, and A.~Farhadi, ``Unsupervised deep embedding for
  clustering analysis,'' in \emph{International conference on machine
  learning}, 2016, pp. 478--487.

\bibitem{mcconville2019n2d}
R.~McConville, R.~Santos-Rodriguez, R.~J. Piechocki, and I.~Craddock,
  ``N2d:(not too) deep clustering via clustering the local manifold of an
  autoencoded embedding,'' \emph{arXiv preprint arXiv:1908.05968}, 2019.

\bibitem{caron2018deep}
M.~Caron, P.~Bojanowski, A.~Joulin, and M.~Douze, ``Deep clustering for
  unsupervised learning of visual features,'' in \emph{Proceedings of the
  European Conference on Computer Vision (ECCV)}, 2018, pp. 132--149.

\bibitem{hwang2020self}
D.~Hwang, J.~Park, S.~Kwon, K.-M. Kim, J.-W. Ha, and H.~J. Kim,
  ``Self-supervised auxiliary learning with meta-paths for heterogeneous
  graphs,'' \emph{arXiv preprint arXiv:2007.08294}, 2020.

\bibitem{wang2020self}
P.~Wang, K.~Agarwal, C.~Ham, S.~Choudhury, and C.~K. Reddy, ``Self-supervised
  learning of contextual embeddings for link prediction in heterogeneous
  networks,'' \emph{arXiv preprint arXiv:2007.11192}, 2020.

\bibitem{li2018deeper}
Q.~Li, Z.~Han, and X.-M. Wu, ``Deeper insights into graph convolutional
  networks for semi-supervised learning,'' in \emph{Proceedings of the AAAI
  Conference on Artificial Intelligence}, vol.~32, no.~1, 2018.

\bibitem{blondel2008fast}
V.~D. Blondel, J.-L. Guillaume, R.~Lambiotte, and E.~Lefebvre, ``Fast unfolding
  of communities in large networks,'' \emph{Journal of statistical mechanics:
  theory and experiment}, vol. 2008, no.~10, p. P10008, 2008.

\bibitem{traag2019louvain}
V.~A. Traag, L.~Waltman, and N.~J. Van~Eck, ``From louvain to leiden:
  guaranteeing well-connected communities,'' \emph{Scientific reports}, vol.~9,
  no.~1, pp. 1--12, 2019.

\bibitem{zhu2020cagnn}
Y.~Zhu, Y.~Xu, F.~Yu, S.~Wu, and L.~Wang, ``Cagnn: Cluster-aware graph neural
  networks for unsupervised graph representation learning,'' \emph{arXiv
  preprint arXiv:2009.01674}, 2020.

\bibitem{sun2020multi}
K.~Sun, Z.~Lin, and Z.~Zhu, ``Multi-stage self-supervised learning for graph
  convolutional networks on graphs with few labeled nodes,'' in
  \emph{Proceedings of the AAAI Conference on Artificial Intelligence},
  vol.~34, no.~04, 2020, pp. 5892--5899.

\bibitem{rong2020self}
Y.~Rong, Y.~Bian, T.~Xu, W.~Xie, Y.~Wei, W.~Huang, and J.~Huang,
  ``Self-supervised graph transformer on large-scale molecular data,''
  \emph{Advances in Neural Information Processing Systems}, vol.~33, 2020.

\bibitem{yanardag2015deep}
P.~Yanardag and S.~Vishwanathan, ``Deep graph kernels,'' in \emph{Proceedings
  of the 21th ACM SIGKDD international conference on knowledge discovery and
  data mining}, 2015, pp. 1365--1374.

\bibitem{dobson2003distinguishing}
P.~D. Dobson and A.~J. Doig, ``Distinguishing enzyme structures from
  non-enzymes without alignments,'' \emph{Journal of molecular biology}, vol.
  330, no.~4, pp. 771--783, 2003.

\bibitem{zitnik2017predicting}
M.~Zitnik and J.~Leskovec, ``Predicting multicellular function through
  multi-layer tissue networks,'' \emph{Bioinformatics}, vol.~33, no.~14, pp.
  i190--i198, 2017.

\bibitem{park2020unsupervised}
C.~Park, D.~Kim, J.~Han, and H.~Yu, ``Unsupervised attributed multiplex network
  embedding,'' in \emph{Proceedings of the AAAI Conference on Artificial
  Intelligence}, vol.~34, no.~04, 2020, pp. 5371--5378.

\bibitem{zbontar2021barlow}
J.~Zbontar, L.~Jing, I.~Misra, Y.~LeCun, and S.~Deny, ``Barlow twins:
  Self-supervised learning via redundancy reduction,'' \emph{arXiv preprint
  arXiv:2103.03230}, 2021.

\bibitem{hao2021pre}
B.~Hao, J.~Zhang, H.~Yin, C.~Li, and H.~Chen, ``Pre-training graph neural
  networks for cold-start users and items representation,'' in
  \emph{Proceedings of the 14th ACM International Conference on Web Search and
  Data Mining}, 2021, pp. 265--273.

\bibitem{yu2020self}
L.~Yu, S.~Pei, C.~Zhang, L.~Ding, J.~Zhou, L.~Li, and X.~Zhang,
  ``Self-supervised smoothing graph neural networks,'' \emph{arXiv preprint
  arXiv:2009.00934}, 2020.

\bibitem{sun2021sugar}
Q.~Sun, H.~Peng, J.~Li, J.~Wu, Y.~Ning, P.~S. Yu, and L.~He, ``Sugar: Subgraph
  neural network with reinforcement pooling and self-supervised mutual
  information mechanism,'' \emph{arXiv preprint arXiv:2101.08170}, 2021.

\bibitem{chuang2020debiased}
C.-Y. Chuang, J.~Robinson, Y.-C. Lin, A.~Torralba, and S.~Jegelka, ``Debiased
  contrastive learning,'' \emph{Advances in Neural Information Processing
  Systems}, vol.~33, 2020.

\bibitem{kalantidis2020hard}
Y.~Kalantidis, M.~B. Sariyildiz, N.~Pion, P.~Weinzaepfel, and D.~Larlus, ``Hard
  negative mixing for contrastive learning,'' \emph{arXiv preprint
  arXiv:2010.01028}, 2020.

\bibitem{robinson2020contrastive}
J.~Robinson, C.-Y. Chuang, S.~Sra, and S.~Jegelka, ``Contrastive learning with
  hard negative samples,'' \emph{arXiv preprint arXiv:2010.04592}, 2020.

\bibitem{shang2019pre}
J.~Shang, T.~Ma, C.~Xiao, and J.~Sun, ``Pre-training of graph augmented
  transformers for medication recommendation,'' \emph{arXiv preprint
  arXiv:1906.00346}, 2019.

\bibitem{zhang2021pre}
J.~Zhang, K.~Chen, and Y.~Wang, ``Pre-training on dynamic graph neural
  networks,'' \emph{arXiv preprint arXiv:2102.12380}, 2021.

\bibitem{zhuang2020comprehensive}
F.~Zhuang, Z.~Qi, K.~Duan, D.~Xi, Y.~Zhu, H.~Zhu, H.~Xiong, and Q.~He, ``A
  comprehensive survey on transfer learning,'' \emph{Proceedings of the IEEE},
  vol. 109, no.~1, pp. 43--76, 2020.

\bibitem{verma2020towards}
V.~Verma, M.-T. Luong, K.~Kawaguchi, H.~Pham, and Q.~V. Le, ``Towards
  domain-agnostic contrastive learning,'' \emph{arXiv preprint
  arXiv:2011.04419}, 2020.

\bibitem{wan2020contrastive}
S.~Wan, S.~Pan, J.~Yang, and C.~Gong, ``Contrastive and generative graph
  convolutional networks for graph-based semi-supervised learning,''
  \emph{arXiv preprint arXiv:2009.07111}, 2020.

\bibitem{chen2020coad}
B.~Chen, J.~Zhang, X.~Zhang, X.~Tang, L.~Cai, H.~Chen, C.~Li, P.~Zhang, and
  J.~Tang, ``Coad: Contrastive pre-training with adversarial fine-tuning for
  zero-shot expert linking,'' \emph{arXiv preprint arXiv:2012.11336}, 2020.

\bibitem{kipf2019contrastive}
T.~Kipf, E.~van~der Pol, and M.~Welling, ``Contrastive learning of structured
  world models,'' \emph{arXiv preprint arXiv:1911.12247}, 2019.

\bibitem{cheng2021drug}
S.~Cheng, L.~Zhang, B.~Jin, Q.~Zhang, and X.~Lu, ``Drug target prediction using
  graph representation learning via substructures contrast,'' 2021.

\bibitem{xu2021selfsupervised}
\BIBentryALTinterwordspacing
M.~Xu, H.~Wang, B.~Ni, H.~Guo, and J.~Tang, ``Self-supervised graph-level
  representation learning with local and global structure,'' 2021. [Online].
  Available: \url{https://openreview.net/forum?id=DAaaaqPv9-q}
\BIBentrySTDinterwordspacing

\bibitem{xu1993rival}
L.~Xu, A.~Krzyzak, and E.~Oja, ``Rival penalized competitive learning for
  clustering analysis, rbf net, and curve detection,'' \emph{IEEE Transactions
  on Neural networks}, vol.~4, no.~4, pp. 636--649, 1993.

\bibitem{yu2021self}
J.~Yu, H.~Yin, J.~Li, Q.~Wang, N.~Q.~V. Hung, and X.~Zhang, ``Self-supervised
  multi-channel hypergraph convolutional network for social recommendation,''
  \emph{arXiv preprint arXiv:2101.06448}, 2021.

\bibitem{wang2021graph}
C.~Wang and Z.~Liu, ``Graph representation learning by ensemble aggregating
  subgraphs via mutual information maximization,'' \emph{arXiv preprint
  arXiv:2103.13125}, 2021.

\bibitem{fatemi2021slaps}
B.~Fatemi, L.~E. Asri, and S.~M. Kazemi, ``Slaps: Self-supervision improves
  structure learning for graph neural networks,'' \emph{arXiv preprint
  arXiv:2102.05034}, 2021.

\bibitem{gao2021topoter}
\BIBentryALTinterwordspacing
X.~Gao, W.~Hu, and G.-J. Qi, ``Topo{\{}ter{\}}: Unsupervised learning of
  topology transformation equivariant representations,'' 2021. [Online].
  Available: \url{https://openreview.net/forum?id=9az9VKjOx00}
\BIBentrySTDinterwordspacing

\bibitem{zhu2003semi}
X.~Zhu, Z.~Ghahramani, and J.~D. Lafferty, ``Semi-supervised learning using
  gaussian fields and harmonic functions,'' in \emph{Proceedings of the 20th
  International conference on Machine learning (ICML-03)}, 2003, pp. 912--919.

\bibitem{sen2008collective}
P.~Sen, G.~Namata, M.~Bilgic, L.~Getoor, B.~Galligher, and T.~Eliassi-Rad,
  ``Collective classification in network data,'' \emph{AI magazine}, vol.~29,
  no.~3, pp. 93--93, 2008.

\bibitem{sehanobish2020self}
A.~Sehanobish, N.~G. Ravindra, and D.~van Dijk, ``Self-supervised edge features
  for improved graph neural network training,'' \emph{arXiv preprint
  arXiv:2007.04777}, 2020.

\bibitem{yasunaga2020graph}
M.~Yasunaga and P.~Liang, ``Graph-based, self-supervised program repair from
  diagnostic feedback,'' in \emph{International Conference on Machine
  Learning}.\hskip 1em plus 0.5em minus 0.4em\relax PMLR, 2020, pp.
  10\,799--10\,808.

\bibitem{huang2021hop}
T.~Huang, Y.~Pei, V.~Menkovski, and M.~Pechenizkiy, ``Hop-count based
  self-supervised anomaly detection on attributed networks,'' \emph{arXiv
  preprint arXiv:2104.07917}, 2021.

\bibitem{giles1998citeseer}
C.~L. Giles, K.~D. Bollacker, and S.~Lawrence, ``Citeseer: An automatic
  citation indexing system,'' in \emph{Proceedings of the third ACM conference
  on Digital libraries}, 1998, pp. 89--98.

\bibitem{mccallum2000automating}
A.~K. McCallum, K.~Nigam, J.~Rennie, and K.~Seymore, ``Automating the
  construction of internet portals with machine learning,'' \emph{Information
  Retrieval}, vol.~3, no.~2, pp. 127--163, 2000.

\bibitem{mernyei2020wiki}
P.~Mernyei and C.~Cangea, ``Wiki-cs: A wikipedia-based benchmark for graph
  neural networks,'' \emph{arXiv preprint arXiv:2007.02901}, 2020.

\bibitem{shchur2018pitfalls}
O.~Shchur, M.~Mumme, A.~Bojchevski, and S.~G{\"u}nnemann, ``Pitfalls of graph
  neural network evaluation,'' \emph{arXiv preprint arXiv:1811.05868}, 2018.

\bibitem{hu2020open}
W.~Hu, M.~Fey, M.~Zitnik, Y.~Dong, H.~Ren, B.~Liu, M.~Catasta, and J.~Leskovec,
  ``Open graph benchmark: Datasets for machine learning on graphs,''
  \emph{arXiv preprint arXiv:2005.00687}, 2020.

\bibitem{li2015unsupervised}
J.~Li, X.~Hu, J.~Tang, and H.~Liu, ``Unsupervised streaming feature selection
  in social media,'' in \emph{Proceedings of the 24th ACM International on
  Conference on Information and Knowledge Management}, 2015, pp. 1041--1050.

\bibitem{harper2015movielens}
F.~M. Harper and J.~A. Konstan, ``The movielens datasets: History and
  context,'' \emph{Acm transactions on interactive intelligent systems (tiis)},
  vol.~5, no.~4, pp. 1--19, 2015.

\bibitem{shervashidze2011weisfeiler}
N.~Shervashidze, P.~Schweitzer, E.~J. Van~Leeuwen, K.~Mehlhorn, and K.~M.
  Borgwardt, ``Weisfeiler-lehman graph kernels.'' \emph{Journal of Machine
  Learning Research}, vol.~12, no.~9, 2011.

\bibitem{borgwardt2005protein}
K.~M. Borgwardt, C.~S. Ong, S.~Sch{\"o}nauer, S.~Vishwanathan, A.~J. Smola, and
  H.-P. Kriegel, ``Protein function prediction via graph kernels,''
  \emph{Bioinformatics}, vol.~21, no. suppl\_1, pp. i47--i56, 2005.

\bibitem{wale2008comparison}
N.~Wale, I.~A. Watson, and G.~Karypis, ``Comparison of descriptor spaces for
  chemical compound retrieval and classification,'' \emph{Knowledge and
  Information Systems}, vol.~14, no.~3, pp. 347--375, 2008.

\bibitem{debnath1991structure}
A.~K. Debnath, R.~L. Lopez~de Compadre, G.~Debnath, A.~J. Shusterman, and
  C.~Hansch, ``Structure-activity relationship of mutagenic aromatic and
  heteroaromatic nitro compounds. correlation with molecular orbital energies
  and hydrophobicity,'' \emph{Journal of medicinal chemistry}, vol.~34, no.~2,
  pp. 786--797, 1991.

\bibitem{kriege2012subgraph}
N.~Kriege and P.~Mutzel, ``Subgraph matching kernels for attributed graphs,''
  \emph{arXiv preprint arXiv:1206.6483}, 2012.

\bibitem{ramakrishnan2014quantum}
R.~Ramakrishnan, P.~O. Dral, M.~Rupp, and O.~A. Von~Lilienfeld, ``Quantum
  chemistry structures and properties of 134 kilo molecules,'' \emph{Scientific
  data}, vol.~1, no.~1, pp. 1--7, 2014.

\bibitem{martins2012bayesian}
I.~F. Martins, A.~L. Teixeira, L.~Pinheiro, and A.~O. Falcao, ``A bayesian
  approach to in silico blood-brain barrier penetration modeling,''
  \emph{Journal of chemical information and modeling}, vol.~52, no.~6, pp.
  1686--1697, 2012.

\bibitem{mayr2016deeptox}
A.~Mayr, G.~Klambauer, T.~Unterthiner, and S.~Hochreiter, ``Deeptox: toxicity
  prediction using deep learning,'' \emph{Frontiers in Environmental Science},
  vol.~3, p.~80, 2016.

\bibitem{huang2016tox21challenge}
R.~Huang, M.~Xia, D.-T. Nguyen, T.~Zhao, S.~Sakamuru, J.~Zhao, S.~A. Shahane,
  A.~Rossoshek, and A.~Simeonov, ``Tox21challenge to build predictive models of
  nuclear receptor and stress response pathways as mediated by exposure to
  environmental chemicals and drugs,'' \emph{Frontiers in Environmental
  Science}, vol.~3, p.~85, 2016.

\bibitem{richard2016toxcast}
A.~M. Richard, R.~S. Judson, K.~A. Houck, C.~M. Grulke, P.~Volarath,
  I.~Thillainadarajah, C.~Yang, J.~Rathman, M.~T. Martin, J.~F. Wambaugh
  \emph{et~al.}, ``Toxcast chemical landscape: paving the road to 21st century
  toxicology,'' \emph{Chemical research in toxicology}, vol.~29, no.~8, pp.
  1225--1251, 2016.

\bibitem{novick2013sweetlead}
P.~A. Novick, O.~F. Ortiz, J.~Poelman, A.~Y. Abdulhay, and V.~S. Pande,
  ``Sweetlead: an in silico database of approved drugs, regulated chemicals,
  and herbal isolates for computer-aided drug discovery,'' \emph{PloS one},
  vol.~8, no.~11, p. e79568, 2013.

\bibitem{gardiner2011effectiveness}
E.~J. Gardiner, J.~D. Holliday, C.~O’Dowd, and P.~Willett, ``Effectiveness of
  2d fingerprints for scaffold hopping,'' \emph{Future medicinal chemistry},
  vol.~3, no.~4, pp. 405--414, 2011.

\bibitem{kuhn2016sider}
M.~Kuhn, I.~Letunic, L.~J. Jensen, and P.~Bork, ``The sider database of drugs
  and side effects,'' \emph{Nucleic acids research}, vol.~44, no.~D1, pp.
  D1075--D1079, 2016.

\bibitem{subramanian2016computational}
G.~Subramanian, B.~Ramsundar, V.~Pande, and R.~A. Denny, ``Computational
  modeling of $\beta$-secretase 1 (bace-1) inhibitors using ligand based
  approaches,'' \emph{Journal of chemical information and modeling}, vol.~56,
  no.~10, pp. 1936--1949, 2016.

\bibitem{riesen2008iam}
K.~Riesen and H.~Bunke, ``Iam graph database repository for graph based pattern
  recognition and machine learning,'' in \emph{Joint IAPR International
  Workshops on Statistical Techniques in Pattern Recognition (SPR) and
  Structural and Syntactic Pattern Recognition (SSPR)}.\hskip 1em plus 0.5em
  minus 0.4em\relax Springer, 2008, pp. 287--297.

\bibitem{kazius2005derivation}
J.~Kazius, R.~McGuire, and R.~Bursi, ``Derivation and validation of
  toxicophores for mutagenicity prediction,'' \emph{Journal of medicinal
  chemistry}, vol.~48, no.~1, pp. 312--320, 2005.

\bibitem{lecun1998gradient}
Y.~LeCun, L.~Bottou, Y.~Bengio, and P.~Haffner, ``Gradient-based learning
  applied to document recognition,'' \emph{Proceedings of the IEEE}, vol.~86,
  no.~11, pp. 2278--2324, 1998.

\bibitem{krizhevsky2009learning}
A.~Krizhevsky, G.~Hinton \emph{et~al.}, ``Learning multiple layers of features
  from tiny images,'' \emph{Master's thesis, Department of Computer Science,
  University of Toronto}, 2009.

\bibitem{jagadish2014big}
H.~V. Jagadish, J.~Gehrke, A.~Labrinidis, Y.~Papakonstantinou, J.~M. Patel,
  R.~Ramakrishnan, and C.~Shahabi, ``Big data and its technical challenges,''
  \emph{Communications of the ACM}, vol.~57, no.~7, pp. 86--94, 2014.

\end{thebibliography}

\end{document}